\documentclass[acmsmall,screen]{acmart} %

\RequirePackage{snapshot}

\usepackage{placeins}
\usepackage{algorithm}
\usepackage{algpseudocode}
\usepackage{listings}
\usepackage[normalem]{ulem}

\usepackage{tabularx,longtable}
\usepackage{tablefootnote}
\usepackage{color,colortbl}
\usepackage{multicol}
\usepackage{multirow}
\usepackage{siunitx}
\usepackage{subcaption}
\usepackage[capitalise,noabbrev]{cleveref}%

\newcolumntype{C}{>{\centering\arraybackslash}X}
\newcolumntype{R}{>{\raggedleft\arraybackslash}X}
\newcolumntype{L}{>{\raggedright\arraybackslash}X}

\graphicspath{%
  {figures/}%
  {figures/bizarre}%
  {figures/maxlen100}%
  {figures/model_python_to_java_metric_noof}%
  {figures/sdxp}%
  {figures/ylc}%
  {figures/zuck}%
}

\def\e{\epsilon}
\def\R{{\mathbb{R}}}

\newcommand{\twocaptions}[2][]{~\hfil #1\hfil #2 \hfil~\\\ignorespaces}

\definecolor{Gray}{gray}{0.9}
\definecolor{Red}{rgb}{1.0,0.4,0.4}
\definecolor{RED}{rgb}{1.0,0.4,0.4}

\newcommand{\songoku}{0}  %

\newcommand{\ie}{i.e.,}
\newcommand{\eg}{e.g.,}
\newcommand{\hp}{$\aleph$-parameter}
\newcommand{\HP}{$\aleph$P}
\newcommand{\al}{$\aleph$-loss}

\newif\ifshowchanges
\showchangesfalse%
\newcommand{\change}[1]{%
  \ifshowchanges
    {\begingroup\color{red}#1\endgroup}%
    \else
    {#1}%
  \fi
}

\newif\ifshowrevisions
\showrevisionsfalse
\newcommand{\revision}[1]{%
  \ifshowrevisions
    {\begingroup\color{magenta}#1\endgroup}%
  \else
    {#1}%
  \fi
}

\lstdefinestyle{pythonstyle}{
    language=Python,
    basicstyle=\ttfamily\scriptsize,
    keywordstyle=\color{black}\bfseries,
    stringstyle=\color{red},
    commentstyle=\color{green!50!black}\itshape,
    breaklines=true,
    frame=none, %
    showstringspaces=false,
}

\begin{document}

\title{Evolutionary Retrofitting}

\author{Mathurin Videau}
\orcid{0009-0002-3672-808X}
\affiliation{\institution{Meta AI}
  \city{Paris}
  \country{France}}
\affiliation{\institution{TAU, INRIA and LISN (CNRS \& Univ. Paris-Saclay)}
  \city{Orsay}
  \country{France}}
\email{mvideau@meta.com}
\author{Mariia Zameshina}
\orcid{0009-0005-4599-2749}
\affiliation{\institution{Univ Gustave Eiffel, CNRS, LIGM}
\city{Champs-sur-Marne}
\country{France}
}
\affiliation{\institution{Meta AI} \city{Paris} \country{France}}
\email{mzameshina@gmail.com}
\author{Alessandro Leite}
\orcid{0000-0002-3071-8019}
\affiliation{\institution{INSA Rouen Normandy, University of Rouen Normandy, LITIS UR 4108}
\city{Rouen}
\country{France}}
\email{aleite@insa-rouen.fr}
\author{Laurent Najman}
\orcid{0009-0004-7294-7081}
\affiliation{\institution{Univ Gustave Eiffel, CNRS, LIGM}
\city{Champs-sur-Marne}
\country{France}
}
\affiliation{\institution{Department of Mathematics - Khalifa University}
\city{Abu Dhabi}
\country{UAE}
}
\email{laurent.najman@esiee.fr}
\author{Marc Schoenauer}
\orcid{0000-0003-1450-6830}
\affiliation{\institution{TAU, INRIA and LISN (CNRS \& Univ. Paris-Saclay)}
  \city{Orsay}
  \country{France}}
\email{marc.schoenauer@inria.fr}
\author{Olivier Teytaud} 
\orcid{0000-0001-5570-5209}
\affiliation{\institution{Thales - CortAIx-Labs}
\city{Palaiseau}
\country{France}
}

\begin{abstract} 
 AfterLearnER (After Learning Evolutionary Retrofitting) consists in applying evolutionary optimization to refine fully trained machine learning models by optimizing a set of carefully chosen parameters or hyperparameters of the model, with respect to some actual, exact, and hence possibly non-differentiable error signal, performed on a subset of the standard validation set. 
 The efficiency of AfterLearnER is demonstrated by tackling non-differentiable signals such as threshold-based criteria in depth sensing, the word error rate in speech re-synthesis, the number of kills per life at Doom, computational accuracy or BLEU in code translation,  image quality in 3D generative adversarial networks~(GANs), \revision{and} user feedback in image generation via Latent Diffusion Models~(LDM). \change{This retrofitting can be done after training, or} dynamically at inference time by taking into account the user \change{feedback}. The advantages of AfterLearnER are its versatility, the possibility to use non-differentiable feedback, including human evaluations \change{(\ie{} no gradient is needed)}, the limited overfitting supported by a theoretical study, and its anytime behavior.
Last but not least, AfterLearnER requires only a \change{small} amount of feedback, i.e., a few dozen to a few hundred scalars, compared to the tens of thousands needed in most related published works. 
\end{abstract}


\settopmatter{printacmref=false}
\setcopyright{none}
\renewcommand\footnotetextcopyrightpermission[1]{}
\pagestyle{plain}

\maketitle

\section{Introduction}

{\em Retrofitting is the addition of new technology or features to older systems} \citep{douglas:06,dawson:07,dixon:13}. Retrofitting is routinely used in industry, e.g., in the building sector, to adapt old buildings to new needs or new regulations.
When it comes to machine learning models, the term is mainly used in natural language processing~(NLP) since the seminal work of \citet{retrofitting2015}, who modified the representation of the word vector to take into account some semantic knowledge. This is done in a {\em post hoc} way, \ie{} without retraining the whole network\change{, though nevertheless requiring a differentiable auxiliary loss function (more in \cref{sec:retrofittingNLP})}.

Other examples of such retrofitting (though not using that term) are used for transfer learning when only some of the last layers of a deep neural network that had been trained for a given task are fine-tuned for another one~\cite{oquab2023dinov2} \change{(more in \cref{sec:fineTuning})}. However, even though \change{they do} not involve the whole set of weights, such re-trainings use gradient descent and\change{,} therefore\change{,} only work with differentiable losses. 

On the opposite, gradient-free (aka black-box) optimization offers two main advantages: it does not require the computation of the gradient, thus decreasing the computational cost \change{and memory requirements}; and it can handle non-differentiable  \change{functions} that are out of reach of gradient-based methods. Indeed, there are many situations where the actual goal of the ML model is better expressed as a non-differentiable metric, and the differentiable \revision{metric} used for training with stochastic gradient descent~(SGD) approaches is only a proxy of such \change{an} actual goal. Examples of such \change{situations} will be given in~\cref{sec:experiments}, like e.g., the number of kills in \change{the} Doom video game~(\Cref{doomxp}).

However, black-box optimizers generally do not scale very well \change{with the number of variables}, and cannot be used to optimize a large set of parameters (e.g., the weights) of current deep models -- with some significant exceptions like, for instance, \change{OpenAI Evolution Strategies \citep{salimans2017evolution}. }

With this in mind, this work proposes AfterLearnER, for {\em After Learning Evolutionary Retrofitting}, which optimizes \change{in a post-hoc way} a small set of parameters and/or hyperparameters of a fully trained model, \change{referred to in the following as the {\bf \hp{}s}\footnote{\change{We use the singular (\hp{}) to denote the full set of parameters or hyperparameters optimized by AfterLearnER.}}}. AfterLearnER \change{is agnostic with respect to the learning algorithm used to train the model}. \change{Furthermore, it can optimize any specific loss function, here termed the {\bf \al{}}, that can be different from the one used during the model training.} Notably, AfterLearnER can handle non-differentiable \al{}es, allowing for better alignment with the actual goals of the learning task without the need to run any gradient backpropagation cycle. \change{A detailed presentation of AfterLearnER is given in~\cref{sec:afterlearner}}.

AfterLearnER clearly relates to hyperparameter optimization~(HPO) \cite{HPO-Hutter2019}, transfer learning \cite{transferLearningSurvey2021}, and fine-tuning~\cite{fineTuningSurvey2023} approaches. Furthermore, because AfterLearnER \change{operates} after \revision{the} standard training \revision{process}, it semantically relates to a number of test-time adaptation~(TTA) approaches~\cite{efficientTTA2022,contrastiveTTA2022}. The \change{crucial} difference from \change{most of} these works is that AfterLearnER can handle non-differentiable \al{}es. AfterLearnER can also be used to handle distributional data shift (like TTA approaches\change{~\cite{efficientTTA2022}}) or to adapt the pre-trained model to new tasks, like transfer learning \cite{oquab2023dinov2}, though it is initially conceived to improve the pre-trained model for the same task, the $\aleph$-loss providing a different point of view on that task. These issues \revision{are} discussed in more detail in~\cref{sota} below and illustrated by the experiments presented in~\cref{sec:experiments}.

\revision{
\paragraph{\textbf{Contributions.}} The main contributions of this work are as follows:

\begin{itemize}
    \item We propose \textbf{AfterLearnER} (After Learning Evolutionary Retrofitting), a novel gradient-free optimization framework that enables the post-hoc tuning of a small set of parameters or hyperparameters (\hp{}s) of fully trained machine learning models, using arbitrary target objectives, including non-differentiable loss functions ($\aleph$-losses), without requiring gradient computation or backpropagation.
    \item In contrast to conventional \textbf{fine-tuning}, \textbf{transfer learning}, and \textbf{test-time adaptation} approaches, which rely on differentiable surrogate losses and retraining cycles, AfterLearnER operates entirely in a black-box optimization setting and can directly optimize the actual performance metrics of interest, even when they are non-differentiable or hard to model.
    \item We show that AfterLearnER is \textbf{training} and \textbf{model-agnostic}, and can be seamlessly applied across a wide range of application domains and modalities. We report experimental results in both offline and online settings, including depth estimation, speech re-synthesis, reinforcement learning for video games, code translation, 3D generative modeling, and text-to-image diffusion.
    \item We provide extensive experimental evidence that AfterLearnER can significantly improve the alignment of a model with real-world target metrics, even when those metrics are not accessible via gradient-based optimization, thus broadening the scope of practical retrofitting beyond what is achievable with conventional fine-tuning, transfer learning, or test-time adaptation approaches.
\end{itemize}
}

The paper is organized as follows. \Cref{sota} surveys the related works: \Cref{sec:diff} (resp. \Cref{non-diff-rewards}) discusses differentiable (resp. non-differentiable) losses, and \Cref{tools} introduces the black-box optimization algorithms used in AfterLearnER.
\Cref{sec:methods} presents AfterLearnER in detail, following some kind of ``user guide'' format. \Cref{sec:rationale} discusses a priori the advantages of the proposed methodology. \Cref{sec:settings} presents our experimental setup and \Cref{sec:experiments} presents the experimental results obtained across various applications by AfterLearnER in offline mode for Depth sensing~(\Cref{midas}), speech re-synthesis~(\Cref{spre}), reinforcement learning for Doom video-game (\Cref{doomxp}), and code translation (\Cref{ct}); and by AfterLearnER in online mode for \ifthenelse{\songoku=0}{interactive GANs (\Cref{egan}),}{} 3D-GANs (\Cref{eg3dcase}), and text-to-image (\Cref{sd,sec:onlinesd}). Finally, \cref{discussion} globally discusses the results, and~\cref{sec:conclusion} concludes the paper. %

\section{\change{Related work}}
\label{sec:background}

\label{sota}

AfterLearnER is related to a number of previous works, with some significant differences that are summarized in \cref{bigcompa} along different axes and discussed in turn in this \change{section}. \textcolor{black}{We are aware that many other works could have been cited here. However, we did our best to cover the most prominent lines of related work, even if we did not cite all relevant papers in each subsection below \change{to avoid bloating}.}

\change{We first survey approaches optimizing a differentiable loss before moving on to those that handle non-differentiable losses. And because the latter is the context of AfterLearnER, \cref{tools} then briefly describes black-box optimization tools, focusing on the ones used in the present work.}

\subsection
{\change{Differentiable loss functions}}\label{sec:diff}%

\change{The works below require a differentiable loss function -- a critical difference with AfterLearnER.}

\begin{table}[tb]\centering\scriptsize
  \caption{\label{bigcompa}Comparison between different works related to AfterLearnER. The columns from left to right are: Whether their target task is the same as the one used during pre-training; Whether any additional knowledge is needed for the optimization; What proportion of the model parameters are modified; Whether some gradient is needed or not; The number of full backprop optimization that \change{is} performed (if any) after the pre-training; The corresponding typical references (there are dozens of other works, and we cannot pretend to be exhaustive).}.
\begin{tabular}{ccccccc}
\toprule
\multirow{2}{*}{Framework} & Final  & Additional  & Reoptimized & \multirow{2}{*}{Gradient needed} & Additional  & \multirow{2}{*}{References} \\
         &  task  & knowledge   & parameters  &              & Full Backprop & \\
\midrule
Hyperparameter tuning &   Same    &         --        & Small part   & No (outer loop) &  Many    & <1>\\
\rowcolor{Gray} Fine-tuning           & Different & Smaller dataset  & Depends & Yes   & A few & <2> \\
Transfer learning     & Different & Distinct dataset & Entire net & Yes & A few & <3><7> \\
\rowcolor{Gray}  Test-Time Training   &   Same    & -- & Large part   &  Yes  &  A few  & <4> \\ 
Test-Time Adaptation   &   Same    & -- & Small part   &  Yes  &  None  & <5> \\ 
\rowcolor{Gray} Retrofitting          &   Same    & High-level info & Small part   &  Depends       &  None    & <6> \\
           &      &  & Small part & No &   & \\
 \multirow{-2}{*}{AfterLearnER}& \multirow{-2}{*}{Same} & \multirow{-2}{*}{Coarse grain info} & or latent vars & (possibly interactive) & \multirow{-2}{*}{None} & \multirow{-2}{*}{}\\
\bottomrule
\end{tabular}
\begin{minipage}{\textwidth}
\begin{multicols}{2}
\begin{itemize}
\item[<1>] \citet{HPO-Hutter2019}
\item[<2>] \citet{fineTuningSurvey2023}
\item[<3>] \citet{transferLearningSurvey2021}
\item[<4>] \citet{TTTseminalICML2020}
\item[<5>] \citet{TENT2021}
\item[<6>] \citet{faruqui-etal-2015-retrofitting}
\item[<7>] \citet{dennis}
\end{itemize}
\end{multicols}
\end{minipage}
\end{table}

\subsubsection{Fine-tuning and transfer learning}\label{sec:fineTuning}

Fine-tuning \change{amounts to optimizing some weights (generally a few layers) of a trained neural network to improve its accuracy for some more specific task than the one it was trained on.} Fine-tuning has now become a routine process in deep learning~\cite{fineTuningSurvey2023}. It is used\change{,} for instance\change{,} to specialize general-purpose \revision{Large Language Models}~(LLMs) for downstream tasks~\cite{ouyang2022training}, including supervised learning and Reinforcement Learning from Human Feedback (see \cref{sec:rlhf} below). Still\change{,} in the context of LLMs, it can be used to ``improve'' the outcomes of the model following the programmer's preferences~\cite{RightWingGPT23}. In such situations, the whole model is fine-tuned using standard gradient-based optimization of the training loss itself, and even though only a fraction of the iterations is needed, this still has a significant cost.

Another use of fine-tuning relates to transfer learning~\cite{transferLearningSurvey2021}: a model that has been trained for a specific task can be fine-tuned to perform a different but similar task, freezing most of the model, \change{and} optimizing only a fraction of it (e.g., one or a few last layers). 
Under some mild hypotheses, it has been demonstrated~\cite{small1,small2} and later proved~\cite{small3} that normalization layers are indeed good candidates for such fine-tuning. This approach is sometimes referred to as Feature Extraction: the frozen layers de facto become feature \change{extractors}. 
Still, the number of \change{unfrozen weights (the \hp{} here)} can be rather high (e.g., more than 100,000 in~\cite{small1}). Furthermore, here again, a few iterations of the full backprop optimization are needed:
The optimization is thus costly, both in \revision{computation} and in memory. 

\change{To mitigate the cost of fine-tuning, Parameter Efficient Fine-Tuning (PEFT) strategies have been developed. A common PEFT approach involves adding a few new blocks to the model, initialized as identity functions while freezing all existing weights, as demonstrated by \citet{houlsby2019parameter}. These newly introduced parameters allow the model to specialize for new tasks. Another PEFT approach injects trainable low-rank decomposition matrices into each layer of the model~\cite
{hu2021lora,liu2024dora}. These matrices adapt the model to the new task, making PEFT a practical alternative to full model fine-tuning.}

As its name says, however, the goal of transfer learning is to handle tasks that are not the ones for which the model was pre-trained, or to modify the behavior of the model, whereas the \textbf{main goal of AfterLearnER is to improve the outputs of the model on the very same task for which it was initially trained, possibly using a non-differentiable \al{} that better describes its desired behavior}. Also, the \hp{} that AfterLearnER tunes is generally much smaller than that of the above works, even when considering that most weights are frozen.

\subsubsection{Test-Time Training / Adaptation}

\textcolor{black}{Whereas fine-tuning and transfer learning pertain to supervised learning in that they require labeled examples of the new task, there are many real-world situations where such labels are not available. A large body of work has been devoted to such contexts, in particular\change{,} to handle the case of distribution shift between training and testing times, pertaining to {\em unsupervised domain adaptation}. A survey and discussion of several of such approaches is given in~\cite{efficientTTA2022}.  
}

\change{A first line of research assigns some labels to the unlabeled test data and performs some {\em test time training}. Such labels can be self-supervision labels, e.g., the number of \ang{90} rotations in images  \cite{TTTseminalICML2020} and the upper layers of the pre-trained model are gradually modified by standard backprop; or one can use some pseudo-labels obtained from a copy of the pre-trained model that is gradually updated \cite{contrastiveTTA2022}. However, in both cases, the size of the \hp{} is rather large (a large fraction of the original model), and the \al{} must be some usual supervised training loss, hence differentiable.
}

\change{On the other hand,
{\em test-time adaptation~(TTA)}, as proposed by \citet{TENT2021}, and further refined by, e.g., \citet{efficientTTA2022,stableTTA2023}, performs some generic entropy minimization, and only optimizes some affine transformation parameters of the normalization layers of the original deep network, in order to fully preserve the pre-trained model. \change{The} TTA \hp{} is hence very small, a feature shared with AfterLearnER. 
However, the entropy minimization is gradient-based, and the \al{} is limited to the chosen entropy \textemdash{} or possibly another differentiable loss \cite{conjugatePseudoLabelsTTA2022}.
}

\textcolor{black}{The main advantage of AfterLearnER compared to test-time adaptation remains the possible use of non-differentiable \al{}, together with, in most cases, the \change{small amount of data needed and the} small compute needed (see \cref{sec:experiments}).
}

\subsubsection{Retrofitting in NLP}
\label{sec:retrofittingNLP}
Retrofitting in Natural Language Processing enhances word vector representations by refining them with relational information from semantic lexicons. This post-\change{hoc} technique adjusts pre-trained word vectors so that semantically related words are closer in the \change{representation} vector space.

\citet{faruqui-etal-2015-retrofitting} introduced a graph-based retrofitting method that refines the word vectors using belief propagation on a graph constructed from the lexical relationships, encouraging linked words to have similar vector representation\change{s}. This process retains the original distributional properties while integrating relational information. ConceptNet \cite{speer2017conceptnet}, a multilingual knowledge graph, links words and sentences through labeled edges and improves the understanding of word meanings. Combined retrofitting with word embeddings forms a hybrid semantic space called ConceptNet Numberbatch, which provides a richer understanding of language.

In summary, retrofitting for NLP leverages external knowledge to refine word representations, enhancing their quality and interpretability for various NLP tasks. \textcolor{black}{Both AfterLearnER and the above works aim to take new knowledge into account to improve the task the model was pre-trained on by modifying a few of its parameters \change{(the \hp{}s)}. However, AfterLearnER can modify any type of \hp{} and is not limited to modifying the latent representation of the model. Furthermore, AfterLearnER uses far less new knowledge than the full lexicons required by NLP retrofitting approaches~\cite{Chiu2019}.}

\subsection{\change{Non-differentiable loss functions}}
\label{non-diff-rewards}

Stochastic gradient descent and its many variants perform extremely well for many machine learning tasks, but can only handle differentiable \change{loss functions}. 
\change{Non-differentiable criteria} used in depth estimation \change{(\cref{midas})},
simulators as in first-person shooters \change{(\cref{doomxp})}, or criteria based on compilers as in code translation \change{(\cref{ct})}, or human satisfaction \change{as in image generation (\cref{eg3dcase,sd,egan})} are beyond gradient-based approaches.

\subsubsection{Reinforcement Learning from Human Feedback}\label{sec:rlhf}
A particular case of non-differentiable \change{loss function is the one built from} human feedback. Reinforcement learning from human feedback~(RLHF) has recently gained importance in image generation~\cite{lee2023aligning} and large language models~(LLMs). However, most low-budget RLHF papers (working with dozens or hundreds of human feedback samples) are in the field of robotics~\cite{rlhfbot2}. The trend in LLMs is more in the dozens of thousands, or even the millions~\cite{abra}, though limited human feedback is sometimes sufficient~\cite{kim2023} and real-time user-specific preferences can sometimes be taken into account~\cite{mathuringecco}.
Low-budget RLHF can be based on inverse reinforcement learning~(IRL) i.e., simulating human preferences from preference signals in a reward model~\cite{inverserl}.
To reduce the need for abundant ground truth data, it is important to focus on small but impactful weights of the model as \hp{}. As already mentioned \change{in}~\cref{sec:fineTuning}, \citet{small1,small2,small3} present such possible small impactful parts, \change{for instance,} the normalization layers, which intersect entirely the information flow from input to output.

\subsubsection{\change{Indirectly using gradients}}
\change{On the other hand, several workarounds have been proposed to handle non-differentiable losses.}
\change{The principle of reward shaping, in various contexts such as non-differentiability or delayed reward, is to modify the loss function, to make it easier to optimize. Using approximations or modifications of the real loss is, however, risky, as pointed out by~\cite{rh}: the solution to the reformulated problem might not be a solution for the original problem. }
In terms of optimization algorithms, \citet{nd1} \change{propose} a method specific to rank-based criteria. \citet{nd2} and \citet{nd3} propose to approximate non-differentiable \change{loss} functions by differentiable ones. \citet{nd4} focuses on criteria defined on the confusion matrix. \citet{nd5} focus on a random subspace of a huge parameter space. \citet{nd6} apply the Reinforce~\cite{Williams_92_SimpleStatisticalGradientFollowing} algorithm. \change{
Also, in~\cite{TENT2021,stableTTA2023,small6}, differentiable criteria are designed as proxies to the actual target \change{loss} functions, adding one level of approximation to the goal, and hence possibly leading to less accurate solutions. %
\revision{On the other hand}, AfterLearnER \change{explicitly handles the non-differentiable \al{},} based on generic black-box optimization (\cref{tools}).}

\subsubsection{Hyperparameter Tuning} \label{sec:hyperparameter}
Hyperparameter optimization~(HPO) has become routine in the ML world in general (see e.g., \citet{HPO-Hutter2019} for a survey), and in deep learning in particular. The \hp{}, \change{in such context,} is made of the usual hyperparameters of the training pipeline and can include architecture parameters (HPO is then called neural architecture search~(NAS)). The optimization is then a two-tiered process, as shown in~\cref{afterlearner}-top, in which the inner loop is a full backprop optimization, and the outer loop is a black-box optimization (e.g., Bayesian~\cite{BO4NAS2018} or Evolutionary~\cite{denserGPEM2019}). But the process is extremely costly because the \al{} is then the performance of the fully re-trained network. As such, we \change{do} not consider it as an alternative to AfterLearnER, \change{nor do we compare AfterLearnER with HPO approaches as they do not tackle the same problems.}

\subsection{Tools for black-box optimization}\label{tools}

Black-box optimization (BBO) refers to optimization methods that only need, from the \change{loss} function $f$ \change{(aka the objective function in the optimization world)}, the value of $f\left(x\right)$ \change{for any given} $x$. In particular, no gradient is used, and no knowledge of the internal structure of $f$ is necessary. \change{Many approaches have been proposed in the literature, from Mathematical Programming (deterministic methods that offer guarantees but sometimes poorly scale up \revision{(unless the problem to solve has some good mathematical properties)} to Metaheuristics~(\ie{} stochastic approaches that often experimentally outperform Mathematical Programming methods but hardly find the exact solution), and many intermediate algorithms. }\change{Among the most popular metaheuristics are Evolutionary Algorithms~\cite{Schwefel81,goldberg:gabook89,HEC97,EibenSmithBook2015}, and Iterated Local Search~\cite{StutzleILS2018}, including Simulated Annealing~\cite{sa1983}, Tabu Search~\cite{Glover:TabuSearch97}, Differential Evolution~(DE)~\cite{de}, and Particle Swarm Optimization~(PSO)~\cite{pso}.}

\Cref{bbo} \change{in \cref{askandtell}} outlines the functioning of the optimization algorithms used in this study, within a parallel computing context, using the well-known {\it ask-and-tell} interface:
the algorithm $o$ is defined by its methods $initialize$, $tell$ (informs $o$ of the value of $f(x)$), $ask$ (returns next value $x$ to evaluate), and $recommend$ (returns its best guess for the optimum). 
There are many libraries for black-box optimization. AfterLearnER uses Nevergrad~\cite{nevergrad} that implements, among others, Evolutionary Algorithms~\cite{EibenSmithBook2015}, Mathematical Programming~\cite{cobyla,powell}, and handcrafted combinations~\cite{iclrbb}. 

In terms of black-box optimization tool\change{s}, AfterLearnER can \change{perform several internal runs} using different optimizers, as developed in~\cref{mainalg}. By default, it uses a single algorithm, namely the optimization wizard~(\Cref{sec:ngopt}) NGOpt~\cite{NGOpt}, which automatically selects a relevant algorithm based on the available computational budget, the problem dimension, and the types of variables to optimize. 
\change{Other algorithms are used in some cases, for the sake of comparison: $(1+1)$ refers to the $(1+1)$-evolution strategy with one-fifth rule~\cite{rechenberg73}, Discrete$(1+1)$ is a classical evolutionary algorithm with mutation rate $1/d$ in dimension $d$ (see discussions in \cite{lengler,relengler}),
DiagonalCMA (the diagonal version of Covariance Matrix Adaptation Evolution Strategies)~\cite{diagcma,CMA}, Differential Evolution~(DE)~\cite{de}, 
Particle Swarm Optimization~(PSO)~\cite{pso}, Lengler which extends the Discrete$(1+1)$ to a decreasing mutation rate~\cite{relengler}; TBPSA refers to Test-Based Population Size Adaptation as in \cite{vasilfoga}, Powell~\cite{powell}, SQP (\change{Sequential Quadratic Programming}~\cite{artelyssqp}), GeneticDE which combines several variants of DE, Cobyla~\cite{cobyla}.}
They are chosen according to the recommendations described in the Nevergrad documentation, as detailed later \change{(\Cref{subsec:doom:results} or~\cref{ct})}. 
\change{We sometimes use the baseline Zero for validation: this is a naive method just recommending the center of the search space, we should, therefore, always do better than this.}
Some algorithms used in our experiments also use prefixes as follows.
``Optim'' means ``Optimistic''~ \cite{munos_bandits_2014-1}, Disc. refers to Discrete~\cite{nevergrad}, ``Recomb'' refers to recombinations by crossover~\cite{holland}. ``Port'' refers to Portfolio, which is the name used in \change{Nevergrad} for the choice of mutation rate in~\cite{danglehre}, ``Split'' refers to an automatic decomposition in blocks of variables\cite{nevergrad}, ``Prog'' refers to progressive optimization inspired by~\cite{wam}, ``Noisy'' refers to a combination with bandit algorithms related to \cite{igel,cauwetnoise,repeatde}.
We use $k<50$ runs (see $k$ in the table) and pick up the one with the best validation \change{\al}, as in~\cref{mainalg}.
The names of all methods are those in Nevergrad~\cite{nevergrad}.

\section{\change{The {AfterLearnER} Algorithm}} %
\label{setup}\label{sec:methods}

\begin{figure}[!t]
    \centering
    \includegraphics[width=.7\textwidth]{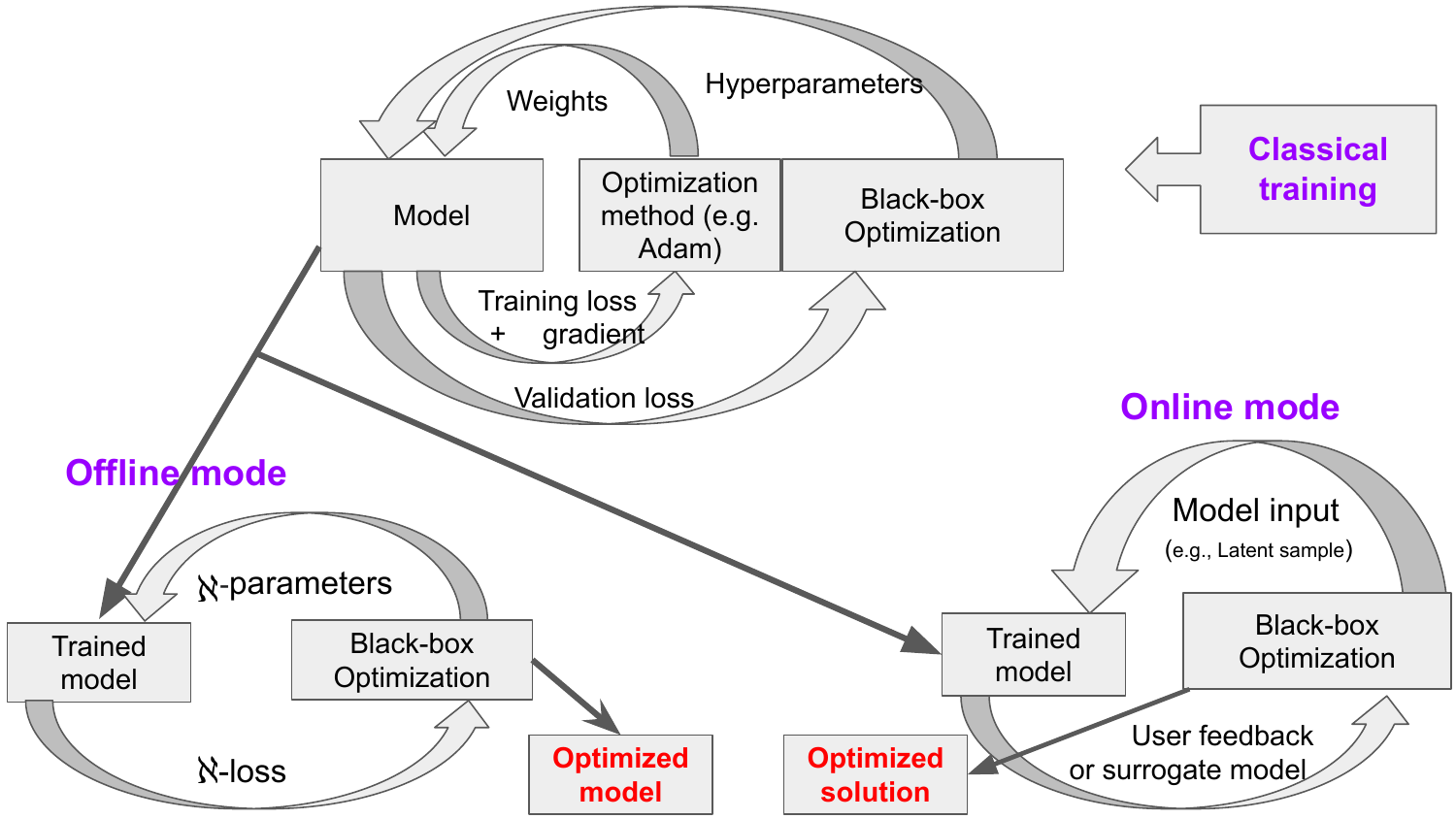}
    \ifthenelse{\songoku=1}{
    \caption{AfterLearnER vs Classical ML. {\bf Top}: Standard gradient-based training (e.g., backpropagation) and hyperparameter tuning (the outer loop). {\bf Bottom}: The two modes of AfterLearnER. {\bf Left}: In the {\em offline mode}, retrofitting of some parameters (termed \hp{}, see text) of the trained model, once and for all before test time, as in \cref{sec:un,sec:deux,sec:sept,sec:huit}, and the output of AfterLearnER is an optimized model. {\bf Right}: In the {\em online mode}, the \hp{} can also include some model input in the latent space, the objective can then be dynamic (the loss, the user feedback, or a surrogate model), as in~\cref{sec:quatre,sec:cinq,sec:six}. The output of AfterLearnER is then an improved output for the given input. \label{afterlearner}}}{
    \caption{AfterLearnER vs. Classical ML. {\bf Top}: Standard gradient-based training (e.g., backpropagation) and hyperparameter tuning (the outer loop). {\bf Bottom}: The two modes of AfterLearnER. {\bf Left}: In the {\em offline mode}, retrofitting of some parameters (termed \hp{}, see text) of the trained model, once and for all before test time, as in \cref{sec:un,sec:deux,sec:sept,sec:huit}, and the output of AfterLearnER is an optimized model. {\bf Right}: In the {\em online mode}, the \hp{} can also include some model input in the latent space, the objective can then be dynamic (the loss, the user feedback, or a surrogate model), as in~\cref{sec:trois,sec:quatre,sec:cinq,sec:six}. The output of AfterLearneER is then an improved output for the given input. \label{afterlearner}}}
\end{figure}

\subsection{\change{Overview}}\label{sec:afterlearner}

The overall view of AfterLearER is given in~\cref{afterlearner}. AfterLearnER can operate before test/inference time (bottom left) or at test/inference time (bottom right) \change{-- more details below}. In both cases, the user should proceed as follows: 
\begin{itemize} 
\item Select the \hp{}, a small subset of the hyperparameters and/or model parameters to be fine-tuned. Several examples of such \hp{} are given in the experimental \cref{sec:experiments,sec:onlineexperiments} \change{and} detailed in~\cref{overview}.
\item  Choose a focused, reliable, possibly non-differentiable \al{} function (e.g., from human feedback) as the objective of the retrofitting. In particular, it can (and \change{is}) be different from the loss used for training the model\change{, which} is required to be differentiable. 

\item Find approximately \change{the} optimal values of the \hp{} using a black-box optimization algorithm to minimize the \al, using only the results of a few inferences \change{on a validation set}, in particular without ever retraining the whole model.
\end{itemize}

AfterLearnER \change{always} operates \change{on the fully trained model, i.e.,} {\bf after the classical backprop training}. As said, there are two modes in AfterLearnER. 
In {\em offline mode} (\Cref{afterlearner}-bottom left, examples in \cref{midas,spre,setup,doomxp,ct}), AfterLearnER optimizes the \hp{} w.r.t. the $\aleph$-loss once and for all, using some low-volume coarse-grain validation data. In particular, it only needs a few inferences of the trained model (and in particular\change{,} no gradient backpropagation).

In the {\em online mode} (\Cref{afterlearner}-bottom right and~
\ifthenelse{\songoku=0}{\cref{egan,eg3dcase,sd,sec:onlinesd}}
{\cref{eg3dcase,sd,sec:onlinesd}}), AfterLearnER optimizes the model answer to each test example during inference, and the \hp{} can typically contain some latent variables, input to the model. This amounts to online retraining based on some small aggregated feedback (see~\cref{sec:SAAF}).

\change{
Note that these two modes are not very different. In both cases, \change{for optimal performance} on an application, we optimize it using the actual satisfaction criterion. When there is one actionable user feedback (e.g., for image generation, see the experiments in \cref{sec:onlineexperiments}), it is used to construct the \al: this is the {\bf online mode}. On the other hand, when playing Doom (\cref{doomxp}) or estimating depth (\cref{midas}), getting user feedback on a significant part of the results would take a significant time. AfterLearnER is then run in {\bf offline mode}, based on data previously gathered. \change{The main difference is that the \al{}, and hence the criterion for success (see also \cref{sec:alephloss}), can be computed in the offline mode, whereas it will be based on human ratings, and hence, rather slowly, in the online mode.}
In both cases, the implementation follows~\cref{mainalg} detailed below, which is, codewise, implemented in a few lines as shown in~\cref{tools}; for a deep learning model stored in GPU memory, we need two more lines for reading/writing directly the weights as shown in~\cref{apcode}.
}

\subsection{\change{The Algorithm}}\label{sec:algorithm}

The AfterLearnER algorithm and its default \change{hyperparameter} settings are presented in~\cref{mainalg}. 
Given an \al{} %
to be minimized on some validation set $\mathcal V$ (running the trained model at hand on the data points in $\mathcal V$), AfterLearnER needs an integer $k$ (the number of \change{times the embedded optimizer(s) will be launched within one run of AfterLearnER, hereafter called number of {\bf internal runs}}), a non-empty list of black-box optimization algorithms, and a list of possible budgets. It then loops k independent times (\Cref{alg:loop}), launching (Line 4) one randomly chosen black-box algorithm in the list (Line 2) to minimize the \al{} on $\mathcal V$, \change{until exhausting} one of the available budgets (Line 3). The result of the run is stored (Line 5), and the best result of the $k$ \change{internal} runs is returned by AfterLearnER (Line 7).

By default, the list of black-box algorithms is made of a single algorithm\change{;} the wizard $NGOpt$ \cite{NGOpt}, presented in more detail in~\cref{sec:ngopt}. As a wizard, it can work in all dimensions/domains/budgets. The budget and the number of runs depend on the expected computational cost and the available overall compute budget. See \cref{mainparam} for some examples of AfterLearnER settings used in the present work. 

\begin{algorithm}
\begin{algorithmic}[1]
	\Require \al{} $\ell$ %
 \Require $\mathcal V $ validation set
	\Require List of black-box optimization algorithms $O$ \Comment{By default $O=\{NGOpt\}$} 
	\Require List of budgets $B$ \Comment{No default value } 
	\Require Number of independent \change{internal} runs $k$ \Comment{By default 1} 
	\For{$i \in \{1,\dots,k\}$}\label{alg:loop}
	    \State{Randomly draw $o_i\in O$} \Comment{only if $\#O>1$}
	    \State{Randomly draw $b_i\in B$}
	    \State{Run $o_i$ to minimize $\ell$ on $\mathcal V$ with budget $b_i$}
            \State{\hp{} $p_i=o_i.recommend()$} \Comment{The result of \change{internal run $i$}}
	\EndFor
	\State{\Return \hp{} $p = ArgMin_{i\in[1,k]} \ell(p_i)$} \Comment {\change{Best result in $k$ internal runs}}
\end{algorithmic}
	\caption{\label{mainalg}The AfterLearnER algorithm. }
\end{algorithm}

\subsection{\change{Rationale for AfterLearnER}}\label{sec:mRethods}\label{sec:rationale}

\change{The main advantage of AfterLearnER is that it works on the real end-user objective function, without requiring an adaptation to gradient-based optimization.}
This section discusses the other potential benefits of using retrofitting beyond the obvious generic advantages of any gradient-free optimization: in settings in which the gradient cannot be computed or does not make any sense\change{;} \eg{} discrete metrics or discrete domains. %

\subsubsection{Computational cost} \label{sec:ccae}
Even when the gradient exists and is easy to compute, it might require a large computational cost, and re-optimizing billions of weights for each value of the hyperparameters is time-consuming. AfterLearnER has a low computational cost: no gradient must be computed during AfterLearnER, therefore saving the memory and computation costs of the backward pass. Examples of budget are given in \Cref{mainparam}.

\subsubsection{\change{Theoretical Analysis of Overfitting Risk}}
\label{maths}
Whereas gradient-based optimization does not lead to any bound on the generalization error, such bounds can be obtained when using black-box optimization, as detailed now. These bounds suggest that the risk of overfitting of AfterLearnER is low. 

Indeed, %
\change{consider $\delta_\epsilon$} the risk of an $\epsilon$-divergence between the empirical risk and the risk in generalization for a given \hp{}:
\[
\delta_\epsilon = P(|(expected\ loss)\ -\ (empirical\ loss\ on\ dataset)| > \epsilon)
\]

If we use the validation error for picking up the best \hp{}, in a list of $N$ \hp{}s (eg{} this list is created by random sampling as in random search, or by low-discrepancy search), the risk of deviation $\e$ for that \hp{} is at most \change{$N\cdot \delta_\epsilon$} (Bonferroni correction~\cite{bonferroni}) instead of \change{$\delta_\epsilon$}: 
\[
\change{P(\exists \mbox{ \hp{}\ } c, |(expected\ loss)_c\ -\ (empirical\ loss\ on\ dataset)_c| > \epsilon) \leq N\delta_\epsilon}
\]
\indent{}If the $N$ \hp s are not defined a priori but built by optimization, with $\lambda$ \hp{}s per iteration and selection of the best at each iteration, then, \change{assuming the optimization algorithm is deterministic, } 
\change{the number of possible internal states of the optimization algorithm
after the choice of one \hp{} among $\lambda$ possible choices is $\lambda$. 
Hence the list of $M$ possible internal states after $n$ iterations has length $\lambda^n$
}~\cite{teytaudfournier}. 
\change{The total risk is therefore} at most $\lambda^n\cdot \revision{\delta_\epsilon}$ where $n$ is the number of iterations.

\change{This proof is valid for deterministic optimization algorithms, with internal states and final output depending only on the selection of one \hp{} at each of the $n$ iterations. 
In the deterministic case, the random part only depends on the evaluation \change{of} a dataset, which is possibly randomized, and on the sampling process providing the dataset. $\delta_\epsilon$ refers to a probability for this probability universe.
For a randomized optimization algorithm, the risk $\delta_{\epsilon,\omega}$ of a deviation $\epsilon$ also depends on $\omega$, the random seed of the optimization algorithm. Then, we get an overall risk 
\begin{equation}
\delta_\epsilon={\mathbb E}_\omega \delta_{\epsilon,\omega}.\label{chain}
\end{equation}
\indent{}However, the deterministic reasoning above also holds for a stochastic optimization algorithm for any fixed random seed $\omega$, and therefore~\cref{chain} makes it also valid for a stochastic optimization algorithm.
}

And if the total number $N$ of evaluations is \change{fixed at} $N$, then $n = N/\lambda$ and the total risk is at most $\lambda^{N/\lambda}\cdot \delta_\epsilon$. %

Furthermore, \change{if AfterLearnER performs $k>1$ independent optimization runs (see internal runs in~\cref{mainalg}), and if the budget \revision{$B$} is divided by $k$}, then the Bonferroni correction leads to a risk \change{bounded by}
\begin{eqnarray}
\underbrace{k}_{\text{Bonferroni correction for $k$ runs}}\cdot \underbrace{\lambda^{(B/(\lambda\cdot k))}}_{\text{Branching factor per run}} \cdot \underbrace{\delta}_{\text{Risk for a single model}}\label{eqov}
\end{eqnarray} 
for the resulting parametrization. 
\revision{These equations, interestingly, lead to upper bounds less than 1 in many cases of interest (\eg{} Hoeffding bound for upper bounding $\delta_\epsilon$, and a dataset size 8000, and $\epsilon=0.05$, and $\lambda=2$ and budget $B=100$, lead to a bound $<0.01$.)}
According to these bounds, increasing the number $k$ of independent runs for a fixed total budget decreases the risk of overfitting. \change{This is particularly visible in~\cref{multirunsresynt}, for the Speech Resynthesis problem, for which overfitting is critical.} Also, according to this bound, increasing the parallelism $\lambda$ decreases the risk of overfitting. 
\revision{A drawback, however, is that increasing the number of runs makes the results closer to a random search and, therefore, possibly less precise.
}

\subsubsection{Versatility, convenience}\label{sec:vc}

Black-box optimization is a very versatile \change{approach}, allowing the user to easily explore different objective functions. For Doom~(\Cref{doomxp}), for fair comparisons, we report results on the same losses \change{as} the ones used by the authors of previous works~\cite{doomai}. However, a significant benefit of AfterLearnER is its ability to adapt a model to a different loss, the \al, without the need to manually craft \change{reward} shaping or to use an approximate proxy. 

We note that the approach can be applied for the optimization of a complex system, including several deep learning models combined with other mathematical programming tools or signal processing methods, even if some of these blocks are non\change{-}differentiable, as illustrated by the experiments presented in~\cref{sec:experiments}.%

\subsubsection{Small Aggregated Anytime Feedback} \label{sec:SAAF}

In summary, the potential benefit of AfterLearnER is that it only needs feedback \change{with the following properties:}

\begin{itemize}
    \item \textbf{Small: }A few dozen scalars is enough, as will be demonstrated in the experiments (see \cref{overview}).
    \item \textbf{Aggregated: } A feedback of one scalar per model is sufficient, instead of one scalar per data point or one scalar and a gradient for each mini-batch. This is visible in our applications when training once and for all a model between training and testing, in \ifthenelse{\songoku=0}{\cref{sec:un,sec:deux,sec:trois,sec:sept,sec:huit}}{\cref{sec:un,sec:deux,sec:sept,sec:huit}}.
    \item \textbf{Anytime: }AfterLearner can be stopped when desired, given the real-time user feedback~\cite{anytime}.
\end{itemize}
\change{The purpose of the rest of the paper is to validate experimentally AfterLearnER on eight diverse use cases that highlight these properties.}

\section{\change{Experimental settings}}\label{sec:settings}
AfterLearnER will now be validated on eight use cases whose main characteristics \change{(including the \hp{}s)} are given in \cref{overview}, while the corresponding AfterLearnER \change{hyperparameters} used in their respective experiments are given in~\cref{mainparam}. 

\subsection{AfterLearnER User Guide}
The choice of these test beds and their setups \change{has} been guided by the following principles, \change{which} can be viewed as a kind of User Guide for AfterLearnER.
\begin{enumerate}
   \item \textbf{Optimize a small but impactful part: } \citet{small1,small2,small3} have demonstrated how a small part of the parameters of an ML model \change{(i.e., a small number of weights)} can have a big impact on its performance. Typically, for a huge network or a combination of modules, one should look for the smallest set of weights that intersect all paths from inputs to outputs. 
 Furthermore, it is important to keep the dimension of the optimization problem small: black-box optimization algorithms do not scale up very well in general \textemdash{} opposite to gradient-based optimization algorithms.
 The \change{second column of \cref{overview} describes the \hp{}s of all use cases, while the third column shows that,  except for Doom, the volume of feedbacks}  is small compared to~\eg{}~\change{\citet{rlhf,hitlrl}}.
 	\item \textbf{Don't fear non-differentiable \al es: } This is the main advantage of AfterLearnER: \citet{small1,small2,small3} optimize a small set of parameters (as we do), but with differentiable criteria. There are many methods for differentiable optimization in machine learning, and gradient-free algorithms cannot outperform these methods unless the gradient is \change{un}reliable. We note that \citet{efficientTTA2022,stableTTA2023,small6} propose to create (possibly without any human feedback or ground truth, thanks to unsupervised entropy minimization) a differentiable criterion based on entropy. However, AfterLearnER focus\change{es} on cases in which a non-differentiable \change{criterion} is worth a direct optimization.
\end{enumerate}
\ifthenelse{\songoku=1}{
}{
\begin{table*}
   \centering
   \caption{
   The size of the  \hp{}s optimized by AfterLearnER varies between 1 and 16384. All the feedbacks in this table are not differentiable. For comparison, a classical ML training with $e$ epochs on a dataset of cardinal $S$ and a gradient of size $p$ has a feedback of size $(\R\cdot \R^p)^{e\cdot S}$.\label{overview}
}
   \footnotesize{%
         \begin{tabularx}{\linewidth}{p{.25\textwidth}XXX}
		\toprule
		\multicolumn{1}{c}{\textbf{Problem}} & \multicolumn{1}{c}{\textbf{\hp{}}} & \multicolumn{1}{c}{\textbf{Feedback volume}} & \textbf{Type of feedback}\\
		\midrule		
		\multirow{2}{*}{Depth sensing} & $6$ input normalization parame\-ters & 300 feedbacks by the large model ($\R^{300}$) & Depth estimate by the large model\\
	\rowcolor{Gray}	\multirow{2}{*}{Speech resynthesis} & $2$ real values (formerly hard-coded) & 40 automatically computed feedbacks for TextLess %
    & \multirow{2}{*}{Word Error Rate}\\  
	\multirow{2}{*}{Doom AI} & Rescaling of the output layer ($35$ weights) & Kills per life in VizDoom, in $\R^{60000}$ & \multirow{2}{*}{VizDoom simulation}\\
      \rowcolor{Gray}   \multirow{2}{*}{Code Translation} & Rescaling between encoder \& decoder ($1024$ weights)& BLEU + comp. accuracy on valid set, in $\R^{1000}$ & Automatically computed score\\
       \bottomrule
Facial Composites (offline fairness, \cref{fairalg}) & \multirow{2}{*}{Single random seed} & 5 user clicks among 30 screens %
& Human preference for the initialization \\
\rowcolor{Gray}		\multirow{2}{*}{EG3D-cats, global modifs} &  Latent variable $z$ (problem dependent) & Image quality: 200 user inputs, one per image%
& \multirow{2}{*}{Surrogate model} \\
		\multirow{2}{*}{EG3D-cats, local modifs} &  Latent variable $z$ (problem dependent) & Image quality: 200 user inputs, one per image%
& \multirow{2}{*}{Surrogate model} \\
        
\rowcolor{Gray} Latent Diffusion Model, 200 images%
    & Latent variable $z$, \hfill in~$\R^{4\cdot 64\cdot 64}$ & Image quality: 200 user inputs, one per image   & \multirow{2}{*}{Human rating + surrogate } \\
 	       Latent Diffusion Model, \hfill 15~images
    & Latent variable $z$, \hfill in~$\R^{4\cdot 64\cdot 64}$ & \multirow{2}{*}{Variable, a few clicks}   & \multirow{2}{*}{Human rating} \\
		\bottomrule
	\end{tabularx}
 }
\end{table*}

}
\begin{table*}
    \centering
    \caption{Typical AfterLearnER parametrization for the test cases of \cref{sec:experiments} and \cref{sec:onlineexperiments} (as several runs are considered, the budgets are not all the same). \change{The specific case of LDM-15 uses an additional} ``Voronoi crossover'' described in \cref{sec:onlinesd}. \\{\footnotesize \change{\(^{\star}\) 10000 for EG3D is an upper bound: The actual number is frequently just a few dozen}. \\\(^{\star\star}\) for this application to LDM, 200 is the number of human feedbacks used in the offline mode for training the surrogate model, which is then used online at each generation.}
    }\label{mainparam}
    \small{
        \begin{tabular}{rllc}
		\toprule
		\multicolumn{1}{c}{\textbf{Problem}} & \multicolumn{1}{c}{{\bf List of Optimizers $O$}} & \multicolumn{1}{c}{{\bf List of Budgets $B$}} \\
		\midrule
		Depth sensing & $\{NGOpt\}$                        & $\{50\}$               \\ %
		\rowcolor{Gray}Speech resynthesis & $\{NGOpt\}$                   & $\{2, 10, 20, 40, 80\}$ \\
		Doom AI           & $\{ OptimDisc(1+1)\}$  & $\{300, 600, 800, 1000, 1200\}$ \\ %
		\rowcolor{Gray}Code translation& $\{DiagonalCMA\}$ & $\{1000\}$ \\ %
		\ifthenelse{\songoku=1}{Image generation}{Facial Composites} Fairness & $\{RandomSearch\}$   & $\{30\}$               \\ %
		\rowcolor{Gray} EG3D-cats, global modifs & $\{ RandomSearch\}$ & 10000* \\ %
		EG3D-cats, local modifs & $\{ NGOpt \}$       & 10000* \\ %
 \rowcolor{Gray} Latent Diffusion Models \change{- 200}  & Lengler (surrogate model) & 200 images** \\
   Latent Diffusion Models \change{- 15}& Lengler (with crossover) & \change{a few batches ($\approx15$ images)} \\

		\bottomrule
	\end{tabular}
    }	
\end{table*}
\ifthenelse{\songoku=0}{
\change{\subsection{Experimental Protocol}}

\change{
\subsubsection{Baselines}
\label{sec:baselines}
Because AfterLearnER starts from a fully trained model, this model is the baseline to which AfterLearner should be compared.
We have used, as far as possible, the implementations that the authors have provided on the internet (\eg{} on GitHub). \change{Each run is} detailed in turn in the corresponding sections.
}
}

\subsubsection{Comparison Metrics} 
\label{comparison-metrics}
\change{There are two fundamental metrics that can be used to compare AfterLearnER with the baselines:  (a) the {\bf training loss} that was used to train the model that AfterLearnER tries to improve, and (b) the {\bf \al{}} used by AfterLearnER during its optimization process. %
One may point out that both metrics are challenging to compare.
First, AfterLearnER could not rapidly obtain positive results without benefiting from the excellent initial point provided by the baseline. Second, standard training and AfterLearnER do not have the same objectives: the \al{} is supposedly closer to the actual goal of the user than the training loss, which is either a differentiable proxy for the \al{} \change{(in the offline mode)}, or lacks user input \change{(in the online mode)}. 
However, because user satisfaction is our ultimate goal, we will use the latter and compare AfterLearnER with the corresponding baseline using the \al{}. 
} %

\subsubsection{Comparing \al es}\label{sec:alephloss}
\change{
For each use case, the performances of \change{the model optimized by }AfterLearnER (in terms of \al{}, as argued above) are evaluated against the baselines on several different scenarios, or {\em contexts} that will be described in the respective sections. These contexts can involve different hyperparameters of AfterLearnER (such as the number of internal runs $k$, the different budgets for the different internal runs B, or the possible choices of the embedded optimizer O, see \cref{mainalg}), or different situations depending on the problem at hand. For each context, \change{a small number} of independent runs of AfterLearnER are performed on a validation set, and the average \al{} (with std. dev.) on a separate test set are reported, as one can see below.} %

\subsubsection{P-values}
\label{sec:pvalue}
\revision{
As stated in~\cref{mainalg}, one run of AfterLearnER launches $k$ internal runs with budgets $b_1,\dots,b_k$, and thus its total budget is $N=\sum_{i=1}^k b_i$. Thus, despite the low cost of AfterLearnER compared to a full backpropagation training procedure, such repeated experiments are still costly, and in some (though not all) of our experimental contexts, a rather small number of independent runs of AfterLearner itself is performed, \textbf{making comparative statistically significant tests impossible for some unique contexts}. 
Nevertheless, when considering several contexts simultaneously, it becomes possible to test the statistical relevance of our findings by computing the p-values through a binomial test (Fisher's exact test), with the null hypothesis being ``the model improved by AfterLearnER does not outperform the baseline on a given context with probability $\geq \frac12$''. We refer to this measure simply by ``{\bf p-value}'' in the following, in the absence of any ambiguity.
}

\subsection{Outline of experiments}
\change{Next two sections present the experimental validation of AfterLearnER on eight test cases.
\cref{sec:experiments} details four experiments that pertain to the \textbf{offline mode}, and directly use \cref{mainalg}: Depth Sensing (\cref{midas}), Speech Resynthesis (\cref{spre}), Doom (\cref{doomxp})\change{,} and Code Translation (\cref{ct}).}

\change{\cref{sec:onlineexperiments} in turn describes the  \ifthenelse{\songoku=0}{four}{three} experiments that the \textbf{online mode} of AfterLearnER, which is then tightly embedded in the inference process: \ifthenelse{\songoku=0}{Facial Composites (\cref{egan}),}{} 3D-GANs with EG3D (\cref{eg3dcase}), Latent Diffusion Models (LDM) with a surrogate model learnt on human feedback (\cref{sd}), and Interactive LDM 
 (\cref{sec:onlinesd}).\\
}

\section{Offline Mode Experiments} \label{sec:experiments}

\subsection{Depth sensing} \label{midas}\label{sec:un}

\subsubsection{Context}
MiDaS~\cite{midas} is a famous deep network for monocular depth estimation from a single image. 
This neural network exists in three trained variants: large, hybrid\change{,} and small. The large one, known for its precision \cite{birkl2023midas}, is considered here as the ground truth. 

Following \cite{midas}, ImageNet images are first resized to $144\times144$ pixels. Then\change{,} AfterLearnER (\cref{mainalg}) uses a set $\mathcal V$ of 300 images randomly drawn from \change{the} ImageNet validation set (of size 50,000). With $b=50$, the feedback used by AfterLearnER is made of the 50 scalars obtained in applying 50 variants of the model  (corresponding to different \hp{}s iteratively provided by AfterLearnER) to the images in $\mathcal{V}$, for each of the $k$ \change{internal} runs. Furthermore, a set ${\mathcal V}'$ of 
500 images randomly drawn from \change{the} ImageNet test set \change{are} used to assess AfterLearnER performance. 

\subsubsection{\hp{}}
AfterLearnER tunes the 6 normalization parameters (multiplication and bias for each of 3 channels) for optimizing the \al{} on $\mathcal V$. 

\subsubsection{\al: Estimating depth} \label{sec:alossMidas}
\citet{depthcriteria} point out that there exist several depth estimation criteria created for different applications: a criterion \change{that} is good for local differences (relative depth) might be plain wrong for global profiles.
Therefore, a diversity of criteria has been proposed, including, among others, the absolute relative error~(AbsRel), the weighted human disagreement rate~ \cite{Monocular_relative_depth_perception_with_web_stereo_data_supervision}, the scale-invariant root-mean-square error~\cite{frozen}, or the frequency of obtaining a mismatch greater than a given threshold~\cite{midas}. These different losses are used in different areas~\cite{deathmap,deathmap2,deathmap3}.

Following \citep{midas}, we focus here on { the frequency of failing}, by a factor greater than $1.25^i$, with $i=1, 2$, or $3$ (further referred to as Threshold $i$).
These criteria are not continuous and are flat almost everywhere\change{,} so that their gradient is essentially zero.
The baselines corresponding to the different Thresholds are computed using the code available at~\url{github.com/isl-org/MiDaS}. 

\subsubsection{Results}

AfterLearnER (\cref{mainalg}) is launched here with values 
$O=\{NGOpt\}$, $B=\{b\}$ (by default $50$), for different values of $k$. %

\begin{figure}[!ht]
\includegraphics[width=.32\textwidth]{{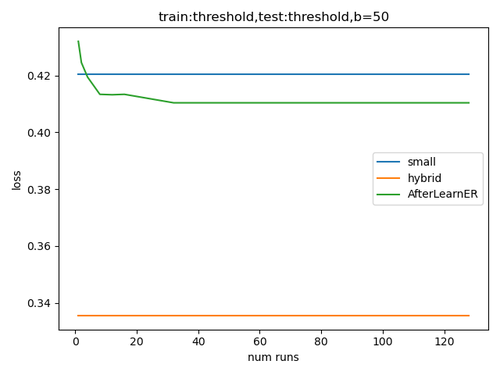}}
\includegraphics[width=.32\textwidth]{{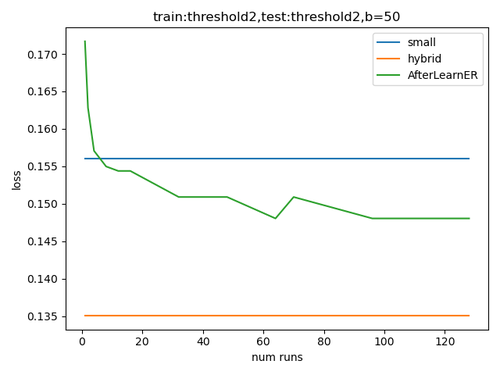}}
\includegraphics[width=.32\textwidth]{{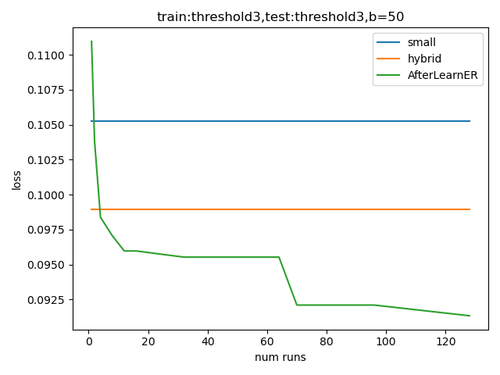}}
    \caption{\label{xpmidas} Depth Sensing: \change{Average} \change{(over 60 runs)} \al es on the test set \change{$\mathcal{V}'$} of the \textcolor{blue}{small} and \textcolor{orange}{hybrid} models \change{(the baselines)}, and of \textcolor{green}{the model optimized by AfterLearnER} on Thresholds 1 (left), 2 (middle) or 3 (right), for different values of  $k$ \change{(x-axis)}, with $b=50$. 
    }
\end{figure}

\cref{xpmidas} displays the \al{} on the test set $\mathcal{V}'$ versus the number of \change{internal }runs $k$, each one with a budget of 50 (other results are presented in  \Cref{appmidas}). %
These results demonstrate an enhancement in the \al{} optimized using the validation set, consistently improving the performance of the Small baseline and sometimes outperforming the performance of the Hybrid model, even with $k<10$.  \Cref{appmidas} also presents evidence of effective transferability: specifically, training on Threshold~3 (i.e., with ratio $1.25^3$) not only improve\change{s the} outcomes in the test set for that metric but also yields positive results for other metrics discussed in~\cref{sec:alossMidas} above. %

\subsubsection{Discussion}
Whereas AfterLearnER needs no more than 50 scalars of feedback in order to improve the Small model, a larger budget allows it to sometimes outperform the Hybrid model with just 300 scalars (e.g., $k=6$ and $b=50$ in~\cref{xpmidas}). And though we used in the present case the large MiDaS model as a ground truth for the sake of reproducibility, if the feedback were to be asked to human raters, it would remain a reasonable effort for small $k$s. 

Furthermore, we observe that results are robust and are effectively transferred to different loss functions (\Cref{appmidas}).

\subsection{Speech resynthesis: a few dozen \change{of} feedbacks for improving sound quality} \label{spre}\label{sec:deux}
TextLessLib~\cite{resynt} is a deep learning library for speech synthesis without text representations. It includes a demo of speech resynthesis~\cite{resynt2} from low bit rates.%

\subsubsection{Context}
The discrete resynthesis operation can be viewed as a form of lossy compression for speech. Approaches such as that of \citet{resynt2} directly train in the waveform using an autoencoder, with adversarial and reconstruction loss. Here, we \change{directly} optimize the Word Error Rate (WER)~\cite{resynt}, which is not differentiable. We use data and code from \cite{resynt2}, splitting the training data into 75\% for the~\hp{} optimization and 25\% as a post-optimization test set. The results can, therefore, not be compared numerically to the results in the original work as we use additional information. Nonetheless, the results show that a few dozen scalar feedbacks are enough for a significant improvement.

\subsubsection{\hp{}}
AfterLearnER handles here two parameters of the Tacotron/Vocoder model \cite{tacotron}: the sigma in WaveGlow/inference~\cite{waveglow} and the denoiser strength. These two parameters are hard\change{-}coded in the original code.

\subsubsection{\al: Word-Error-Rate} This measure evaluates how accurately the audio can be transcribed after compression, assessing how well the textless compression preserves speech-specific characteristics and clarity. WER is computed through the wav2vec 2.0-based~\cite{wav2vec2} and Automatic Speech Recognition~(ASR) system: the code compares the output with the ground-truth transcription. The WER is also used to evaluate the retrofitted model on the test set.

\subsubsection{Results: robust low  budget quality improvement}
The results of different parameterizations of AfterLearnER are presented in~\cref{multirunsresynt}. 
One can observe that picking up the best outcome from multiple shorter \change{internal }runs (\ie{} \change{a large $k$} with a small budget $b$ per run) yields the most favorable results. In certain instances, the \change{internal} number of runs $k$ is even so large, and the budget per run so limited that the approach  resembles a random search (\eg{} a $9.9\%$ improvement observed with the best of 8 runs each having a budget of 2). These experiments demonstrate that AfterLearnER can significantly improve test results at a low \change{\al{}} within a few evaluations on a small validation set. 

\subsubsection{Discussion} \Cref{multirunsresynt} shows that for this use case with minimal sample size, the utilization of a high number of runs \change{(and a smaller $b$)} consistently enhances performance in generalization, aligned with the theoretical discussions presented in~\cref{maths}: such results experimentally demonstrate the reduction of the risk of overfitting by using a large number of runs and a small budget per run.
\begin{table*}
    \centering
    \caption{Word Error Rate (\al{}) on the test set in textless Speech Resynthesis for different numbers of internal runs $k$ and different budgets $b$ per internal run. 
    The baseline WER is 7.84. 
     For each  \change{pair $(k,b)$}, several independent runs are run \change{(at least 6 per row -- std dev in the last colum): 100\% of contexts} lead to an improvement on average. %
\change{The p-value (\cref{sec:pvalue}) over the 16 independent \change{contexts}, under the null assumption that the probability of improvement is not greater than .5, is significant (p-val$\leq1.6e-5$).} \change{
Consistently with Section \ref{maths}, we observe low overfitting and excellent performance for $k$ large.
}
}
     \label{multirunsresynt}
    \small{
    \begin{tabular}{ccccc}
      \toprule
         \textbf{Internal} & \textbf{Budget}   & \textbf{WER}  & \textbf{Average}  & \textbf{Std. dev.} \\
         \textbf{runs} $k$& \textbf{$b$}   &  $(\downarrow)$ & \textbf{ improv.} (\%, $\uparrow$) & \textbf{WER} \\
       \midrule
        \rowcolor{Gray}
        &  2 &     7.488 &  4.511 &  0.335\\
        \rowcolor{Gray}
        &  10 &     7.558 &  3.620 &  0.202\\
        \rowcolor{Gray}
        &  40 &     7.736 &  1.361 &  0.235\\
        \rowcolor{Gray}
        \multirow{-4}{*}{3} &  80 &     7.542 &  3.828 &  0.298\\
        &  2 &     7.306 &  6.837 &  0.387\\
        &  10 &     7.501 &  4.355 &  0.147\\
        &  40 &     7.751 &  1.163 &  0.217\\
        \multirow{-4}{*}{4} &  80 &     7.378 &  5.924 &  0.201\\
        \rowcolor{Gray}
        &  2 &     7.222 &  7.908 &  0.397\\
        \rowcolor{Gray}
        &  10 &     7.442 &  5.104 &  0.130\\
        \rowcolor{Gray}
       \multirow{-3}{*}{5} &  40 &     7.803 &  0.502 &  0.149\\
       
        &  2 &     7.273 &  7.259 &  0.417\\
       \multirow{-3}{*}{6}   &  10 &     7.433 &  5.216 &  0.104\\
        \rowcolor{Gray} &  2 &     7.097 &  9.506 &  0.361\\
        \rowcolor{Gray}
        \multirow{-2}{*}{7} &  10 &     7.398 &  5.661 &  0.076\\
       8 &  2 &     7.067 &  {\bf{9.880}} &  0.344\\
    \bottomrule
  \end{tabular}
}
\end{table*}

\subsection{Doom AI: Retrofitting for Deep Reinforcement Learning} \label{doomxp}\label{sec:sept}

\subsubsection{Context}
We explore the application of AfterLearnER in the context of the game ``Doom'', in the framework defined by \citet{doomai}. 
Our study focuses on two distinct setups: ``Normal mode'', where the agent combats 10 bots, and ``Terminator mode'', involving a confrontation with 20 bots. As the base trained network, we use the agent defined in~ \cite{doomai}, which is trained in \textit{Normal mode}.

\subsubsection{\hp{}}
We add a flattened action head of length 35 in the final layer of the baseline network, where the output $o$ \change{(of dimension 35)} is adjusted to $o\cdot \exp\left(r\right)$: These 35 scalars are the \hp{} to optimize. 

\subsubsection{\al: Kill/Death ratio} Since the main objective in Doom is to kill as many bots as possible while minimizing \change{your own} deaths, we optimize the non-differentiable and long-term objective of the kill/death ratio.
However, this reinforcement signal is very noisy: when $k>1$, it might happen that the comparison between the different signals on the last iteration of each \change{internal }run, is a wrong indicator of which of the $k$ runs is the best one. So, instead of the last value of this kill/death ratio, we use a moving average of the last $8$ values as the \al{} returned to the black-box optimizer. %

\subsubsection{Results: AfterlearnER kills monsters in Doom}\label{subsec:doom:results}
A clear improvement can be seen in~\cref{arnoldtab}.
In that Table, the suffix 10 (resp. 100) refers to dividing the scale by 10, \ie{} we use $\exp\left(0.1 r\right)$ (resp. $\exp\left(0.01 r\right)$) instead of $\exp\left(r\right)$. The suffix w20 means that we use $P=20$ parallel workers inside each of the $k$ \change{internal} runs, as opposed to a single worker: this reduces the \change{wall-clock computational cost}, but also the number of available feedbacks, as shown in~\cref{bbo} \change{in Appendix B}. 
\change{Note that these experiments are run in preemptable mode on the cluster, allowing for interruptions, hence the uncontrolled number of internal runs $k$ in \cref{arnoldtab} (last column). We observe, nonetheless, positive results in almost all cases}. %

\change
{Here, we use a single black-box optimization method $o$ in $O$ for each experiment, but we repeat the experiment with distinct $o$ for the sake of comparisons}.
We refer to~\cref{tools} for more details about the different algorithms mentioned in \cref{arnoldtab}.

\subsubsection{Discussion}
The terminator mode is a priori easier to optimize: the baseline agent was originally optimized for normal mode, a priori leaving a more significant gap of potential improvement by AfterLearnER for this transfer learning. 

In terminator mode, the best results are obtained by methods based on DiscreteOnePlusOne. Therefore, we only used this optimization algorithm for the normal mode, and \change{we can observe an improvement} over the baseline in all cases. The validation score for TBPSA is the lowest: this is because TBPSA~\cite{vasilfoga} explores widely and tries to learn from samples far away from the current optimum (as recommended in~\cite{mlis}). Its performance in the test is nonetheless among the top 4.

\begin{table*}[!ht]
    \centering
    \caption{\label{arnoldtab} Results on Doom. The red marker indicates the only detrimental run of AfterLearnER, which reduced test performance. 
\change{For each of the 16 lines, we run AfterLearnER several times \change{(at least 3 runs per row)}. The p-value over these 16 lines (\cref{sec:pvalue}) is less than $3.10^{-4}$.} 
}%
    \footnotesize{%
    
    \begin{tabularx}{\linewidth}{crCCCc}
		\toprule
& & {\bf Total} & & {\bf Average} & {\bf Number of}\\
& {\bf Optimizer} & {\bf budget} & \multirow{-2}{*}{{\bf Validation}} & {\bf test} & \change{\bf internal} \\
& & $\sum_{i\leq k} b_i$ & \multirow{-2}{*}{{\bf score ($\uparrow$)}} & {\bf score ($\uparrow$)} & {\bf runs} $k$ \\
		\midrule
		& \textbf{Baseline} &\multicolumn{2}{c}{ } &  3.61 & \\
  \cmidrule{3-5}
& DiagonalCMAw20&9348&2.447&3.642&11\\
& NoisyDisc(1+1)w20&6864&3.590&3.616&15\\
 \rowcolor{Red}
& Noisy(1+1)w20&10880&3.319&3.589&17\\
& OptimDisc(1+1)10&4500&3.700&\textbf{3.719}&11\\
& OptimDisc(1+1)w20&5980&3.712&3.682&13\\
& OptimNoisy(1+1)w20&6764&3.651& 3.685&14\\
& ProgNoisyw20&8028&3.114&3.615&18\\
& RecombPortOptimNoisyDisc(1+1)10&5400&3.686&3.658&15\\
& RecombPortOptimNoisyDisc(1+1)w20&5696&3.630&3.639&17\\
& SplitNoisyw20&13144&3.234&3.646&18\\
\multirow{-13}{*}{Terminator mode} & TBPSAw20&5964&2.404&3.678&15\\
\cmidrule{2-6}

&		\textbf{Baseline} &\multicolumn{2}{c}{ } &  3.31 & \\
	\cmidrule{3-5}
& OptimDisc(1+1)100&51500&3.382&3.356&87\\
& OptimDisc(1+1)10&31200&3.419&3.399&50\\
& OptimDisc(1+1)w20&6644&3.380&3.430&20\\
& RecombPortOptimNoisyDisc(1+1)100&46900&3.453&\textbf{3.447}&75\\
\multirow{-7}{*}{Normal mode} & RecombPortOptimNoisyDisc(1+1)10&19950&3.346&3.329&36\\
\bottomrule
  \end{tabularx}
 }
\end{table*}

\subsection{Code translation} \label{ct}\label{sec:huit}

\subsubsection{Context} Transcompilers perform code translation, i.e., convert source code from one programming language to another while maintaining the same level of abstraction. Unlike traditional compilers, which translate high-level code to low-level languages like assembly, transcompilers focus on translating between languages of similar complexity. This process is useful for updating obsolete codebases or integrating code written in different languages. \citet{unsuptran} leverage unsupervised machine translation techniques to develop a neural transcompiler. Trained on source code from open-source GitHub projects, their transformer model can accurately translate functions between C++, Java, and Python. This method uses only monolingual source code, requires no specific language expertise, and can be generalized to other programming languages. The present section uses this neural transcompiler as a baseline, and applies AfterLearnER before test/inference time (\cref{mainalg}) to improve its performance.

\subsubsection{\hp{}: linear rescaling}
A linear rescaling layer (1024 weights) is added between the encoder and the decoder, and these 1024 scalars are the \hp{} to be optimized. 

\subsubsection{\al: a diversity of criteria}
Many criteria can be used for code translation: computational accuracy, which checks that both programs' outputs are equal for the same input, code speed, code size, BLEU~\cite{bleu}, which checks the correctness and readability of the code by measuring the similarity of n-grams between generated code and reference code, adjusting for length. Recent papers~\cite{bugct} pointed out that overfitting or other issues can arise, making the joint improvement of robust and diversified criteria relevant, in particular on smaller but curated data sets. In this work, we use either the {\bf computational accuracy} (frequency of obtaining a code which satisfies the requirements) or the {\bf BLEU}~\citep{bleu}\footnote{From the abstract of \citep{bleu}: BLEU is {\em a method of automatic machine translation evaluation that is quick, inexpensive, and language-independent, that correlates highly with human evaluation, and that has little marginal cost per run.}}. None of these \al{}es is differentiable.

 \begin{figure*}[t]
   \centering\centering
    \includegraphics[width=.45\textwidth]{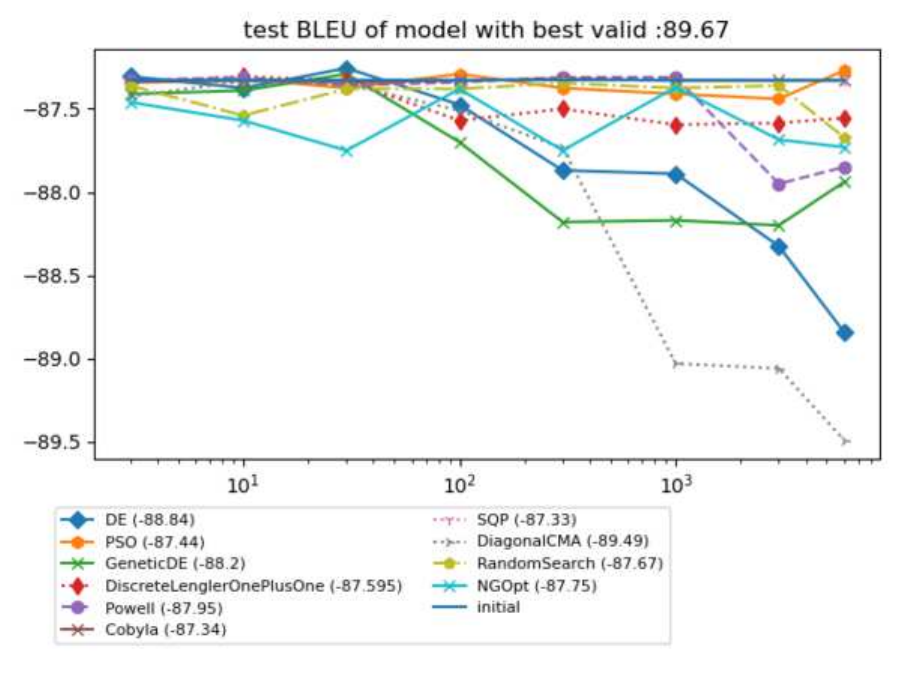}
    \includegraphics[width=.45\textwidth]{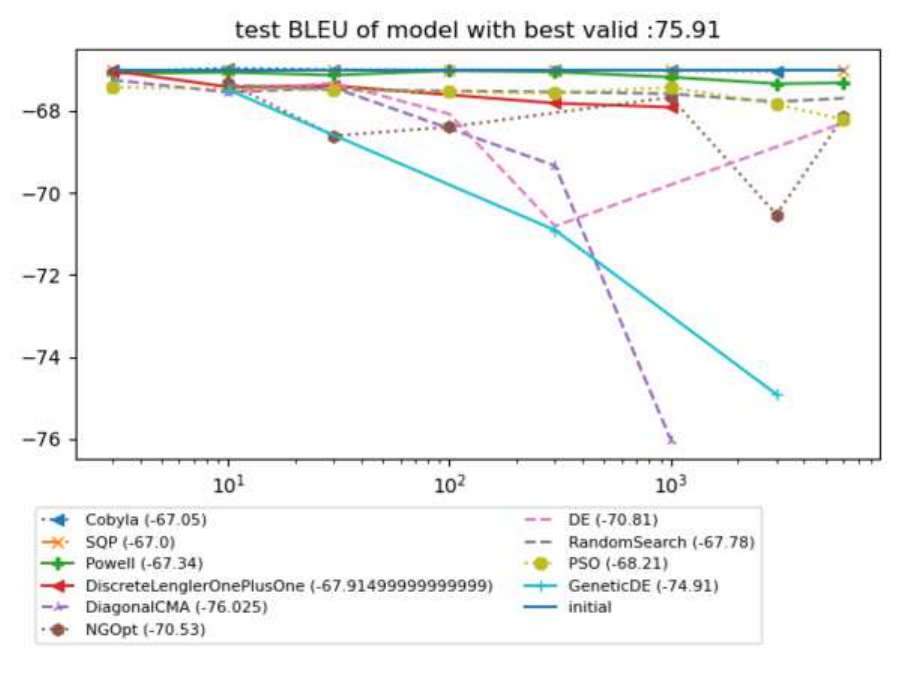}\\
 	\twocaptions[BLEU: C++ $\to$ Java]{BLEU: Python $\to$ C++}
 \includegraphics[width=.45\textwidth]{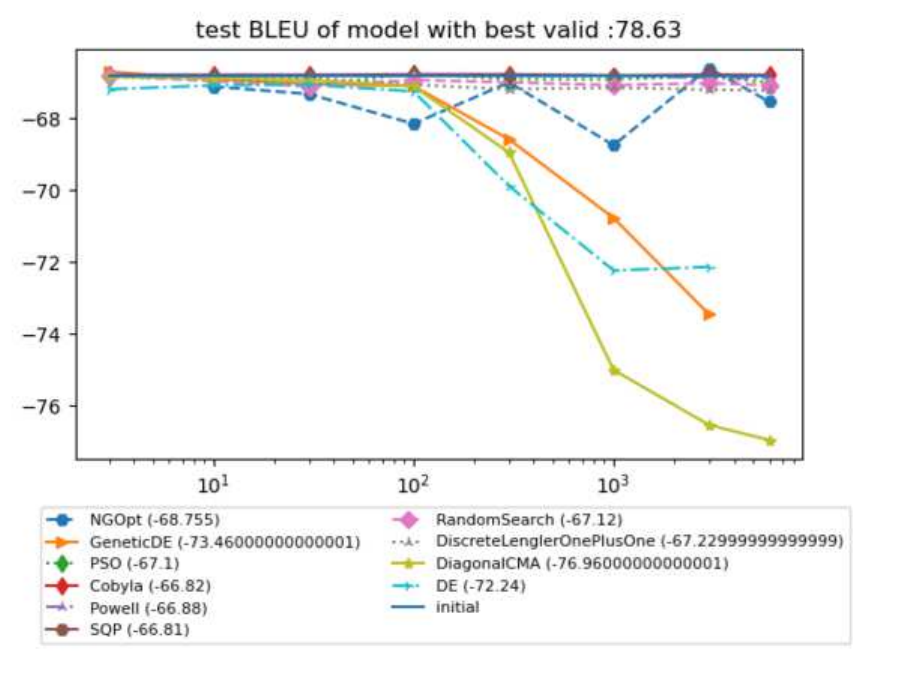}
 \includegraphics[width=.45\textwidth]{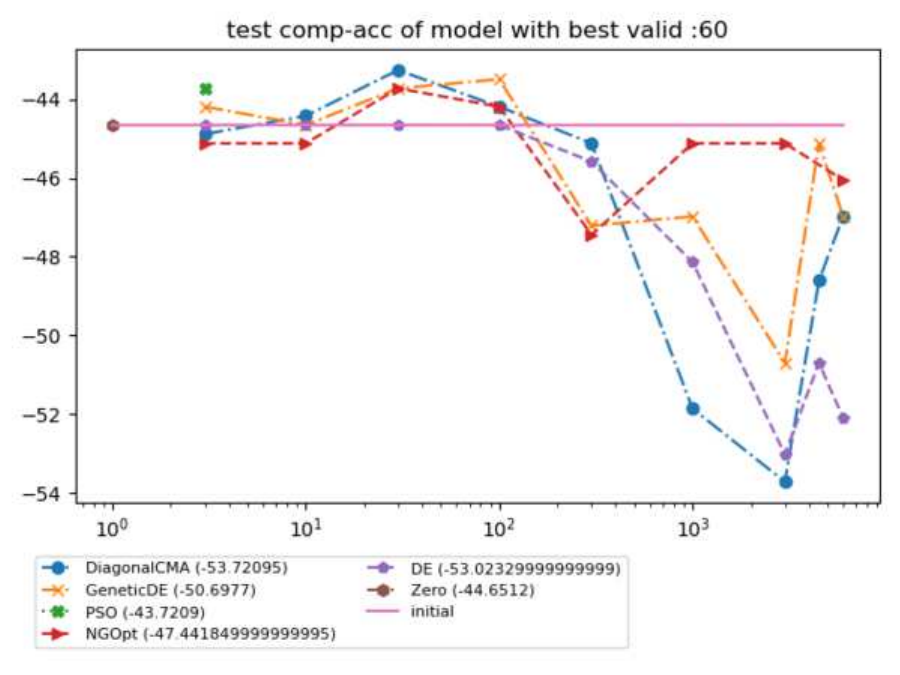}\\
 	\twocaptions[BLEU: Python $\to$ Java]{Comp. accuracy: Python $\to$ Java}
	 \caption{\label{figcodegen}  Results of AfterLearnER on several problems from~ \cite{unsuptran} with BLEU or Accuracy \al{}s (the lower the better). Each curve is the average over 3 runs of AfterLearner \change{with $k=1$ (i.e., one internal run) and the set $O$ limited to a single optimization method.} The horizontal line is the baseline, i.e., without any retrofitting. 
     \change{Note that if we remove Zero (which is a baseline run for validation, which does not modify the baseline) and SQP (which does not make sense for this dimensionality), AfterLearner is successful in 31 independent cases out of 32,  with p-value $\leq 1e-8$ (\cref{sec:pvalue})}.}
\end{figure*}

\subsubsection{Results: AfterLearnER can code} 
\Cref{figcodegen} compares the solutions obtained using AfterLearnER (with different black-box optimizers $o$) to the baseline solution using 
a feedback limited to $b$ ground truth scalar numbers ($b$ on the $x$-axis). $O$ is restricted to a singleton $\{o\}$ (see $o$ in the legend) and a single run ($k=1$).

The results obtained with AfterLearER improve over the baseline\revision{,} whatever the optimizer used, though some optimizer\revision{s} perform better than others. Empirically, {\em DiagonalCMA} is often the best choice in this context.

\subsubsection{Discussion} 
These experiments have been performed with the distribution of test cases in the original code from~\citet{unsuptran}, \ie{} \change{code size limit at 100 tokens}: we investigated the case of a limit 512, but further work is needed on the dataset as there are redundancies between valid\change{ation} and test in that context in particular when using computational accuracy, which aggregates distinct but functionally similar codes.

\section{\change{Online Mode Experiments}} \label{sec:onlineexperiments}

\ifthenelse{\songoku=0}{
\subsection{\change%
{Evolutionary GAN: Facial Composites with Fairness}
}\label{egan}\label{sec:trois}

\subsubsection{Context: Improving diversity in the latent space of Evolutionary Interactive GANs.} %
Generative adversarial networks~(GANs)~\cite{gan,pytorchganzoo,goodcgan,pgan} are now a widely used generative model. Interactive GANs~\cite{tog,evogan} propose images to the user and repeatedly modify their proposals based on \change{the} user's feedback. 
We first present such an interactive GAN, namely the one from \citep{evogan}, presented in an instance of AfterLearnER \change{in online mode}, namely ~\cref{algevolgan}.

Given that the human feedback is based on clicks, the black-box optimizer $o$ must be a rank-based algorithm, i.e., it only uses comparisons between fitnesses, regardless of actual values. Likewise, the default value of $o$ is Lengler's method~\cite{lengler} because it is the recommended variant of DiscreteOnePlusOne (see~\cite{nevergrad}).
Despite the effectiveness of DiscreteOnePlusOne for taking into account user feedback on the fly, unfairness \change{due to the} lack of diversity remains an issue due to the so-called mode collapse. Thus, it is frequently challenging to get some image classes~\cite{fairgan}.

Therefore, we propose~\cref{fairalg}, a variant of AfterLearnER for choosing the seed parameter $s$ in~\cref{algevolgan} so that it generates a more diverse initial population.
\cref{algevolgan} can be seen as having a seed $s$ as \hp{}, that we optimize with a simplified AfterLearnER~(\cref{fairalg}) as follows:
\begin{itemize}
  \item As a proxy of~\cref{algevolgan}, we use the user feedback on the initial screen generated by~\cref{algevolgan} for a specific random seed $i$.
  \item We optimize this random seed $i$ by enumerating $30$ possibilities and requesting user feedback: what are the 5 most diverse seeds? 
  \item Then, the algorithm uses one of the 5 preferred seeds (randomly drawn).
\end{itemize}

\change{The motivation for randomly using one of the 5 preferred seeds is to ensure that the code is still randomized and that repeated runs can have some diversity even in the first iteration. Note that \change{later} iterations are anyway randomized.}

\subsubsection{\hp{}: The random seed} Before running~\cref{algevolgan}, we optimize the ``random'' seed that produces the most diverse initial images according to the user feedback, as described in~\cref{fairalg}.

\subsubsection{\al: Interactive measure of diversity} To evaluate the efficiency of~\cref{algevolgan}, which now incorporates seed optimization from~\cref{fairalg}, various facial categories were targeted, and the \al{} measures the user's satisfaction. We report \change{the number of iterations over } 25-image screens required to achieve satisfactory results in the user's eye for each category. Here, we measure diversity considering up to four categories: Female Asian, Female Blonde, Male Black, and Male Blond~(see~\cref{examplesfaces}).

\subsubsection{Results: fairness improvement on various categories}
Preliminary runs show that it is frequently hard to reach\change{, as far as we can tell from very long runs,} some categories with the original version of the algorithm (\Cref{algevolgan}) when the initial population is not diverse enough. Hence our first experiment will check that several such categories can be reached with our optimized ``random'' seed \change{obtained by~\cref{fairalg}}. 
On average, the algorithm achieves the desired outcomes within $3.0 \pm 0.2$ screen iterations over the four categories. We conclude that this limitation was primarily due to the initial set of faces on the presented screen belonging to the same category, thereby creating a high hurdle for diverging from that specific cluster.

\begin{algorithm}
	\begin{algorithmic}[1]
		\Require{A black-box rank based optimizer $o$} \Comment{by default  DiscreteLenglerOnePlusOne}
        \Require {$\lambda$, number of images per screen}
        \Require{$\mu<\lambda$ number of clicks (selected images) per screen}
        \Require {a GAN converting latent variables $(c_i)_{1\leq i \leq \lambda}$ to images}
		\State {If $s$ is not specified, $s \leftarrow$ random()}\label{randomSeedInit} 
		\State{From $s$, generate first $\lambda$ parametrizations $c_1,\dots,c_\lambda$}
        \label{latentGeneration}		
        \While {user not satisfied}
			\State{Display $GAN(c_1),\dots,GAN(c_\lambda)$ to the user} \label{imageGeneration}%
			\State{user clicks on $\mu$ images  $(i_1,\ldots,i_\mu)\in \{1,\dots,\lambda\}^\mu$ }
			\State{Report to $o$ that the fitnesses of $c_{i_1},\ldots,c_{i\mu}$ are greater than those of $c_j$ for $j\not \in\{i_1,\dots,i_\mu\}$}
			\State{$o$ gives back next $\lambda$ parametrizations $c_1,\dots,c_\lambda$} %
        \EndWhile
	\end{algorithmic}
	\caption{\label{algevolgan} Base algorithm from~\citep{evogan} for generating images as desired by users. \change{Users} only click on their favorite image, on each screen.}%
\end{algorithm}
\begin{algorithm}
	\begin{algorithmic}[1]
		\Require{$\lambda$, number of images per screen.}
        \Require {a GAN converting latent variables $(c_i)_{1\leq i \leq \lambda}$ to images}
		\State{Randomly choose $s_1,\dots,s_{30}$ random seeds}
		\For{$1\leq i \leq 30$}
       \Comment{Surrogate of \cref{algevolgan} with random seed $i$}
		    \State{GAN generates $\lambda$ images using seed $i$} \Comment {similarly to~\cref{algevolgan}, \cref{latentGeneration}}.%
		    \State Display this screen to the user
		\EndFor
		\State{ask the user for the 5 most diverse screens $i_1,\dots,i_5$} \Comment{Human Feedback}
		\State{\cref{randomSeedInit} of \cref{algevolgan} becomes "$s\leftarrow$ random-choice$([i_1,\dots,i_5])$" }%
	\end{algorithmic}
	\caption{\label{fairalg}AfterLearnER for seed optimization: after running this algorithm, the 
 random seed (\cref{randomSeedInit} of~\cref{algevolgan}) becomes ``Initialize $o$ with a random choice of $s$ in an optimized subset of 5 \change{seeds}''.}
\end{algorithm}

\subsubsection{Discussion}
This experiment is a combination of~\citep{evogan} and a straightforward application of AfterLearnER to optimize its ``random'' seed: in spite of its simplicity, this approach solves a real diversity problem and is a reminder that taking care of a seed can have a big impact (as also reported in~\cite{seedbridge}, which applied seed optimization for winning a world computer championship of bridge). 
For more technical novelty, \cref{eg3dcase} extends this work to 3D GANs, adds surrogate models, and considers local convergence (see the ``evolutionary'' version).
\Cref{sd} extends it to latent diffusion models, 
and \cref{sec:onlinesd} adds Voronoi crossovers.%

} %

\subsection{EG3D-cats: local and global optimization of a surrogate model}
\label{eg3dcase}\label{sec:quatre}

\subsubsection{Context: Offline optimization of the latent variable thanks to a surrogate model} \label{sec:surrogateGAN}
EG3D-cats~\cite{eg3d} is a pioneering work in 3D GANs, with impressive results. In the present Section, we improve EG3D-cats by learning in the latent space: instead of randomly drawing a latent variable $z$, we learn a surrogate model of image quality and use it to optimize the input $z$ to the GAN, in order to \change{avoid} $z$'s leading to low quality. 

\paragraph{Building the surrogate model}
We proceed as follows:
\begin{itemize}
	\item Generate 200 images corresponding to 200 randomly chosen latent variables $z$.
	\item Get binary user feedback: good or bad, for each image.
	\item Train a decision tree (we use the scikit-learn implementation~\cite{sklearn}) to approximate the probability, for a given latent variable $z$, that it generates a bad image: $tree(z)\approx P\left(bad|z\right)$.
\end{itemize}

\subsubsection{\hp{}:} The latent variable $z$ of EG3D-cats\change{: dimension 512}. 

\subsubsection{\al: The surrogate model} 
\begin{figure}[t]\centering
	\includegraphics[width=.48\textwidth]{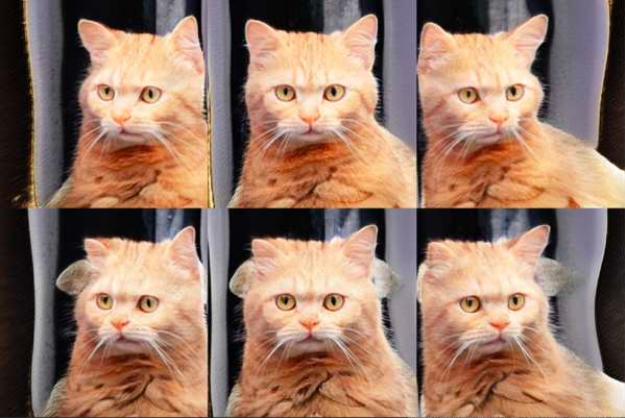}
 	\includegraphics[width=.447\textwidth]{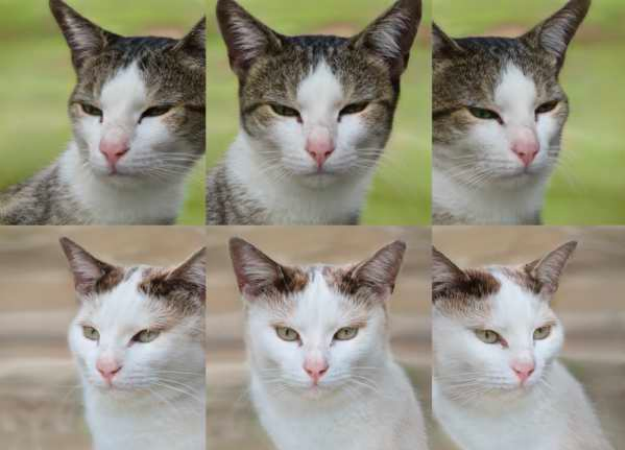}
	\caption{\label{evogan1} Example of evolution: unmodified cat (bottom) generated by EG3D-cats and its evolutionary counterpart (top). Left: the ears have been significantly improved, while the neck is improved on the right.}
\end{figure}

\begin{figure}[t]\centering
	\includegraphics[width=.48\textwidth]{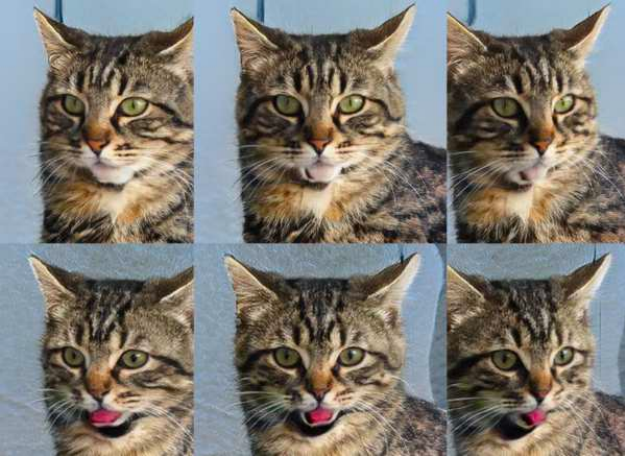}
	\includegraphics[width=.45\textwidth]{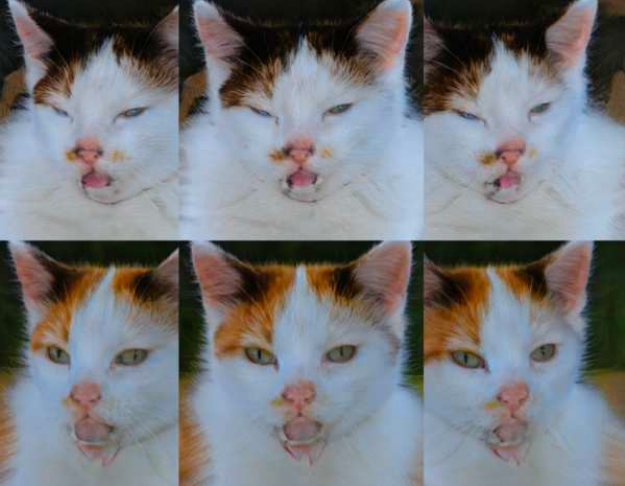}
	\caption{\label{evogan2} Example of evolution: unmodified cat (bottom) generated by EG3D-cats and its evolutionary counterpart (top). In both cases, the mouth has been significantly modified, though it is still not perfect. }
\end{figure}

We propose two variants of the algorithm, both using the trained decision tree described above to optimize the \hp{} $z$:

\begin{itemize}
	\item The {\em random search variant} applies random search on $tree$: Randomly draw $z$ until $tree(z) =0$, i.e., the trained decision tree predicts that the image will be ok.%
	\item The {\em evolutionary version} (searching for local modifications) applies NGOpt on the objective function $l:x\mapsto tree(z_0+\epsilon \cdot x)$ for some randomly drawn $z_0$. For a small $\epsilon$ (\eg{} $1\mathrm{e}{-2}$), the final value is close to the starting point $z_0$, thus preserving the diversity of the original GAN: this is the main advantage of the evolutionary version compared to the random search. \change{Note that the evolutionary version works in $\R^n$ and has, therefore, no theoretical guarantee of staying close to $z_0$: however, it is initialized with normal random variables, and therefore the small $\epsilon$ implies that it searches in the neighborhood of $z_0$ first.}
\end{itemize}
\change{In both cases, we stop at a budget of 10000.}
To evaluate the approach, we use human raters to assess the improvement between the initial $z$ and the optimized $z$. The humans (not working in the computer science field) are presented with a cat generated by EG3D-cats and a cat generated by AfterLearnER: they choose the cat image they prefer or click on ``no preference''. The success rates are evaluated among cases in which they do have a preference. For the evolutionary variant, raters have to choose one of both cat images, but only images for which the initial $z_0$ was not satisfactory (i.e. $P(bad|z_0) >0$) are presented: these statistical comparisons are based on paired data, i.e., $z_0$ vs $z_0+\epsilon \cdot x^\star$ with $x^\star$ recommended by the optimization run. \Cref{evogan1,evogan2} present some qualitative results before and after the evolution done by AfterLearnER.

\subsubsection{Results: offline improvement of the model} %
The results\change{, with 4 human raters,} are presented in~\cref{tabeg3d}. In all setups, we observe an improvement when using AfterLearnER compared to vanilla EG3D (i.e., all numbers are larger than $50\%$). \ifthenelse{\songoku=0}{The most significant improvement is seen in the enhancement of faces, with 90\% of cases showing \change{a} modification.}{} Of course, the different random seeds used for computing these statistics are distinct from those used for creating the training set of 200 images.

\ifthenelse{\songoku=0}{
\begin{table*}
   \centering
   \caption{Results of the AfterLearnER improvement of the EG3D-cats algorithm with human raters, using a double-blind interface with randomized left/right positioning of compared images: the percentage shows how frequently our method outperformed the vanilla LDM according to human raters.\label{tabeg3d}}
   \footnotesize{%
  \begin{tabularx}{\linewidth}{rclX}
    \toprule
      \multicolumn{1}{c}{\textbf{Initial model}} & \textbf{Proposed method} & \textbf{Frequency of~improvement} $(\uparrow)$ & \textbf{Remark} \\
    \midrule
 afhqcats512-128      & Random search  &  65\% (160 images) & Cats   \\ %
      \rowcolor{Gray}
 	afhqcats512-128      &    & 71\% (193 images) & Cats \\ %

      \rowcolor{Gray}
 ffhqrebalanced512-64 & \multirow{-2}{*}{Evolution}     & 90\% (180 images) & Faces \\
    \bottomrule
  \end{tabularx}}
\end{table*}}{
\begin{table*}
   \centering
   \caption{Results of the evolutionary version of the EG3D-cats algorithm, \change{as estimated by} human raters, using a double-blind interface with randomized left/right positioning of compared images: the percentage shows how frequently our method outperformed the vanilla LDM according to human raters.\label{tabeg3d}}
   \footnotesize{%
  \begin{tabularx}{\linewidth}{rclX}
    \toprule
      \multicolumn{1}{c}{\textbf{Initial model}} & \textbf{Proposed method} & \textbf{Frequency of~improvement} $(\uparrow)$ & \textbf{Remark} \\
    \midrule
 afhqcats512-128      & Random search  &  65\% (160 images) & Cats   \\ 
      \rowcolor{Gray}
 	afhqcats512-128      & Evolution   & 71\% (193 images) & Cats \\ 
    \bottomrule
  \end{tabularx}}
\end{table*}}


\subsubsection{Discussion} We demonstrated that it is possible to learn a good approximation of the set of $z$'s that lead to errors using a decision tree at a reasonable cost: labelling 200 examples is feasible for a motivated human.

{
\subsection{LDM: %
learning a surrogate model on 200 human feedbacks
\label{sec:cinq}}%
\label{sd}

\subsubsection{Context: Using feedback to steer the model output towards human preferences.}

Latent Diffusion Models (LDM)~\cite{stablediffusion} recently provided impressive results in text-to-image generation: a prompt conditions the output. 
This section is based on~\cite{mathuringecco}, and applies the same techniques on latent models as in~\cref{eg3dcase}.%

\paragraph{Building the surrogate model}
\begin{itemize}
	\item Generate 200 images on the prompt ``a photograph of an astronaut riding a horse''.
	\item Get a binary user feedback: good or bad, with a special emphasis on the quality of crops.
	\item Train a classifier to approximate the probability, for a given latent variable $z$, that it generates a bad image: $tree(z) \simeq P\left(bad|z\right)$. %
\end{itemize}

\change{Depending on the experiment, the classifier is }a decision tree or a neural network, and we use the scikit-learn implementation in both cases~\cite{sklearn}\footnote
 {We use \lstinline[language=Python,basicstyle=\ttfamily,breaklines=true]{MLPClassifier(solver='lbfgs', alpha=1e-5, hidden_layer_sizes=(5, 2), random_state=1)}
 ~and  \\\lstinline[language=Python,basicstyle=\ttfamily,breaklines=true]{tree.DecisionTreeClassifier()}. 
}.

\subsubsection{\hp{}} The latent variables of LDM.

\subsubsection{\al: The Surrogate Model}\label{metrics:satisfaction_rate} 

As in \cref{sec:surrogateGAN}, we use the trained decision tree as AfterLearnER \al{} to determine an optimal latent value $z$ that minimizes $tree(z) \approx P\left(bad|z\right)$. The optimization is stopped as soon as it reaches some $z$ satisfying $tree\left(z\right)=0$.

To assess the effectiveness of this approach, we rely on human raters. Specifically, raters are asked: “Is the crop of this image of top quality and does it include the entire astronaut and horse?” We refer to this evaluation metric as the 'satisfaction rate'.

\subsubsection{Results}
\change{ The training is done on images labeled by us, and the performance is measured on other generated images with/without AfterLearner, with different seeds. The human raters, in randomized order among the 4 tested methods, work in double-blind. The prompt is ``A photograph of an astronaut riding a horse''.}
The results are presented in~\cref{figsd}. We observe progress in all contexts: the frequency of poorly cropped images dramatically decreases.

\begin{table*}
  \centering
  \caption{ Results of AfterLearnER on LDM. Between parenthesis the number of rated images. The overall comparison LDM vs. AfterLearnER+LDM is statistically significant, with p-value 0.04 for Fisher's exact test. The 'satisfaction rate' is described in \cref{metrics:satisfaction_rate}.
\change{The p-value (\cref{sec:pvalue}) for the difference between LDM (2.9\% on 486 runs) and the average case (5.7\% on 382 runs) is p-val$=0.02$.}
}\label{figsd}
  \begin{tabular}{rccc}
   \toprule
    \multicolumn{1}{c}{\textbf{Experimental setup}} & \textbf{Optimizer} & \textbf{Loss}  & \textbf{Satisfaction rate} $(\uparrow)$ \\
   \midrule
     LDM & & &  2.9\% (486) \\
     \cmidrule{2-4}
		\rowcolor{Gray} LDM+AfterLearnER & Random search   & Decision tree & 5.1 \% (98) \\
		LDM+AfterLearnER & Lengler         & Decision tree & 5.6 \% (180) \\
	\rowcolor{Gray} 	LDM+AfterLearnER & Discrete$(1+1)$ & Neural net    & 6.7 \% (104) \\
    \cmidrule{2-4}
      Average LDM+AfterLearnER &   & &   5.7 \% (382)\\
    \bottomrule
   \end{tabular}
\end{table*}

\subsubsection{Discussion: }
This success looks surprising: how can we learn something with 200 examples, whereas the latent space has \change{dimension} $4\cdot64\cdot64 = 16384$? 
Nevertheless, the results presented in the next Section will confirm these results, switching to an online context, with an even lower budget.

\subsection{LDM with online human feedback, learning from 15 images}\label{sec:six}
\label{sec:onlinesd}

\subsubsection{Context: }In order to double-check the results above (successful inference from a few data points in a huge latent space), we perform the following experiment:
\begin{itemize}
   \item 15 images are generated with the same text input.
   \item The users select their five favorite images.
   \item Learning a ``local'' surrogate model on the fly: these \change{(at most 5)} favorite images are separated from the other 10, using scikit-learn neural network.
   \item Similarly to both previous sections, we minimize the probability of generating bad images (according to the local surrogate model) using Nevergrad (Lengler algorithm~\cite{lengler}), with an initial point created by crossover \change{based on the favorite images as parents} \change{(as proposed in~\cite{mathuringecco} and as detailed below)}.
\end{itemize}
\label{sec:co}

\paragraph{Application of the crossover operator \change{for 2 parents}} The vanilla optimization by Nevergrad is here improved by creating a starting point for the optimization run built using a specific crossover operator, so-called Voronoi crossover~\cite{hamda:02,voronoi1,voronoi2,voronoi3,mathuringecco}, in order to provide a better merging between different individuals than by geometric averages: the Voronoi split between $z_1$ and $z_2$ for getting a child corresponding to a crossover between images with latent variables $z_1$ and $z_2$ (\Cref{co}). These Voronoi combinations are used as starting points for the optimization by AfterLearnER. %
\begin{algorithm}[t]\centering
    \begin{algorithmic}
        \Require{Latent variables $z_1$ and $z_2$, coordinates $p_1\in I_1=LDM(z_1)$ and $p_2\in I_2=LDM(z_2)$.}
        \State{Create a new latent variable $z$ of same shape as $z_1$ and $z_2$.}
        \For{each $p$ spatial coordinate in $z$.}
          \If {$||p_1-p|| < ||p_2-p||$}
            \State{$z(p)=z_1(p)$}
          \Else
            \State{$z(p)=z_2(p)$}
          \EndIf
        \EndFor
    \end{algorithmic}
	\caption{Pseudo-code for the crossover of images with latent variables $z_1$ and $z_2$, at points $p_1$ and $p_2$. Here, the user has selected the point $p_1$ in the image generated from the latent variable $z_1$ and the point $p_2$ in the image generated from the latent variable $z_2$, out of 15 images. Latent variables are tensors with shape $64\times 64 \times 4$.}\label{co}
\end{algorithm}

\textcolor{black}{
\paragraph{Application of the crossover operator for more than 2 clicks by users}
Two modifications of \cref{co} are needed. First, we might have more than 2 images chosen by the user, so we extend~\cref{co} to more than two images. Second, we need to create more than one offspring, so that randomization is necessary. 
For these two extensions, we follow \cite{mathuringecco}: let us consider $k$ clicks, with positions $p_1,\dots,p_k$, in images built on latent variables $z_{i_1},\dots,z_{i_k}$.
Each offspring image $I_{Voronoi}=G(z_{Voronoi})$ is built by constructing a latent variable $z_{Voronoi}$ as in Figures~\ref{contractionfactor}-\ref{vor2}.
\begin{eqnarray}
r& & \mbox{ is randomly drawn uniformly in }[1,2] \label{contractionfactor}\\
z_{Voronoi}(x)&=&z_{i_j}(x)\mbox{ if } \ \forall u\neq j, ||x-p_j|| < \frac1r ||x-p_u||\label{vor1}\\
z_{Voronoi}(x)&\sim& \mathcal{N}(0,\,1) \mbox{for each channel otherwise} \label{vor2}
\end{eqnarray}}

}{}
\subsubsection{\al} To evaluate this approach, we use a specific prompt and quality \change{criterion} (see below). We generate a batch of images with the vanilla LDM, and a second batch, using AfterLearnER, after selection of the 5 best by the user. 

We then compare the number of images that fit the given \change{criterion} in the initial batch (no user feedback) and in subsequent batches (i.e., with user feedback).

\subsubsection{Results} 
\begin{figure*}[t]\centering
\includegraphics[width=.65\textwidth]{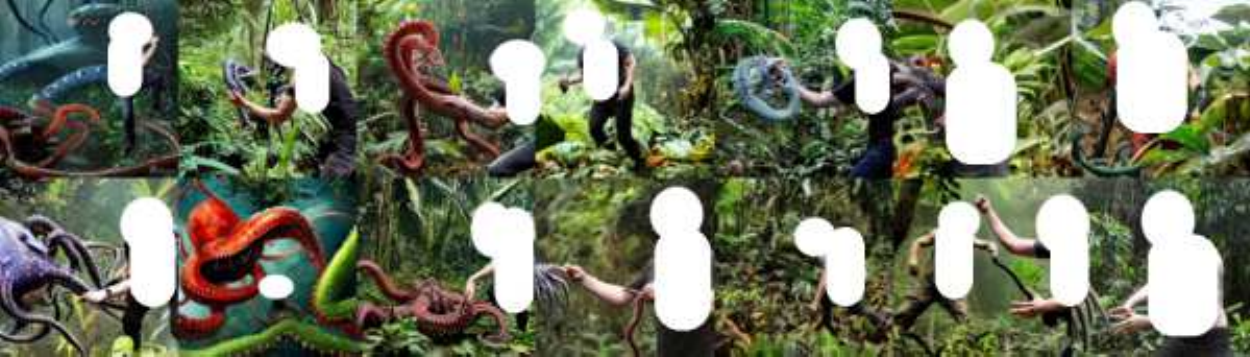}\\
First batch of 14 images. Tentacles: in 11 images. XY is the only human: in 12 images.\\
\includegraphics[width=.65\textwidth]{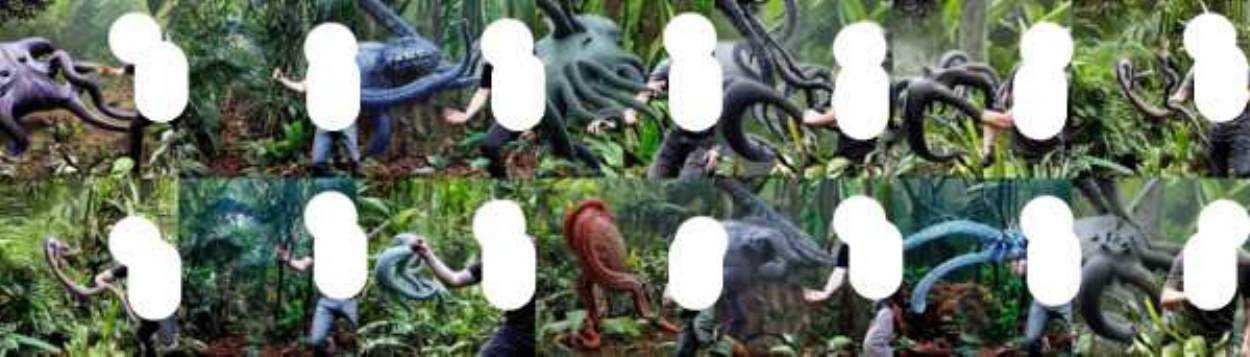}\\
Second batch of 14 images. Tentacles in 12 images. XY is the only human: in all images.\\
\includegraphics[width=.65\textwidth]{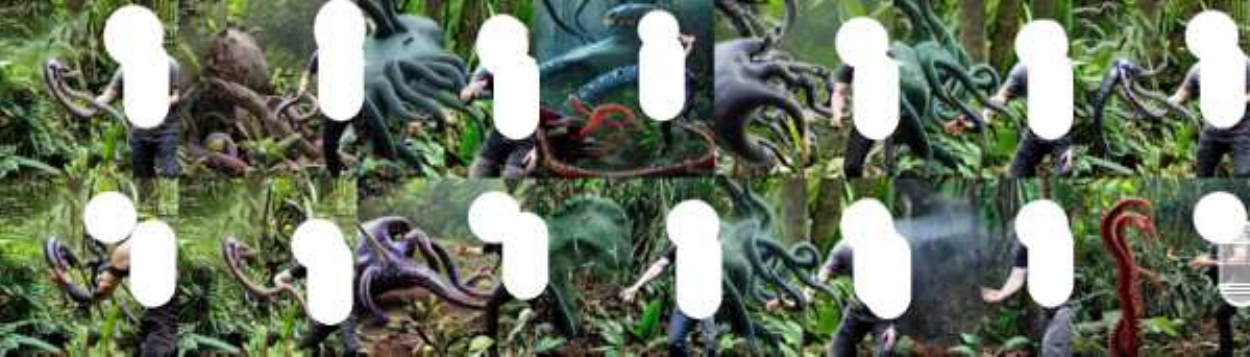}\\
Third batch of 14 images. Tentacles: in 13 images. XY is the only human: in all images.\\
\caption{\label{zuck}XXX YYY (anonymized) fighting tentacles in the jungle. The original LDM is already good here, but we observe an increased frequency of XXX YYY and an increased frequency of tentacles in the new batches. A more challenging task is proposed in~\cref{ylc}.}
\end{figure*}

\begin{figure*}[t]\centering
\includegraphics[width=.9\textwidth]{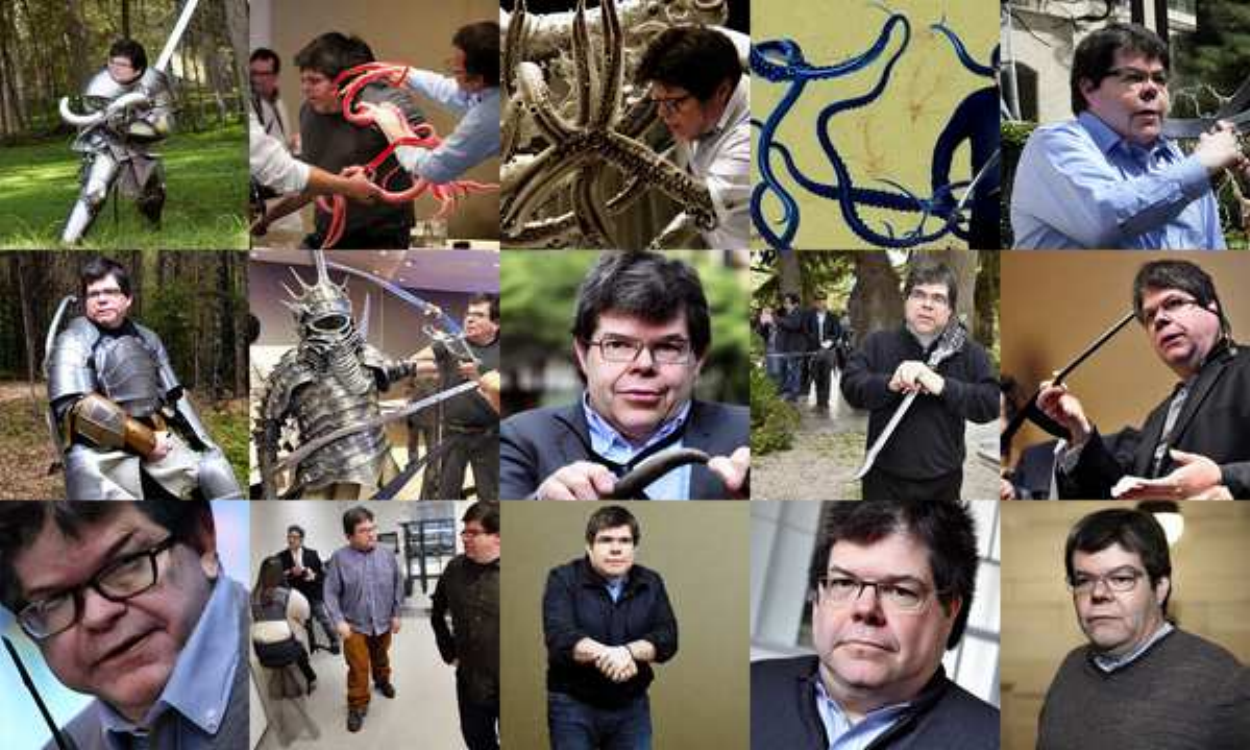}\\
Initial generations of 15 images, by the vanilla LDM code. Tentacles in 4 or 5 images, over 15: tentacles are difficult to obtain when a scientist is in the picture.\\
\includegraphics[width=.9\textwidth]{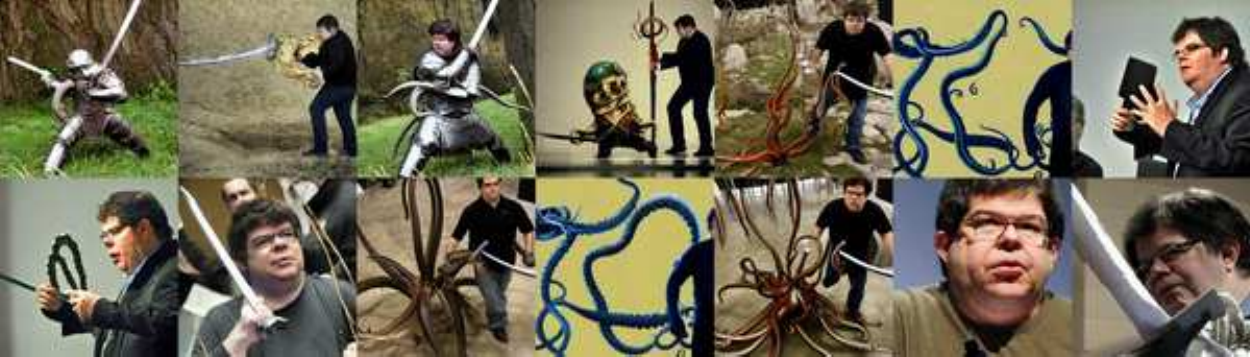}\\
After selection of the 5 preferred images out of the 15 above, 14 new generations by AfterLearnER. Tentacles in 10/14 images.\\
\caption{\label{ylc}A scientist fighting tentacles in the jungle. The frequency of tentacles has increased a lot. All images have significant differences in terms of position. We observe again that 15 images evaluated by the user are enough for a meaningful impact.}
\end{figure*}

\begin{figure*}[t]
  \centering
  \includegraphics[width=.9\textwidth]{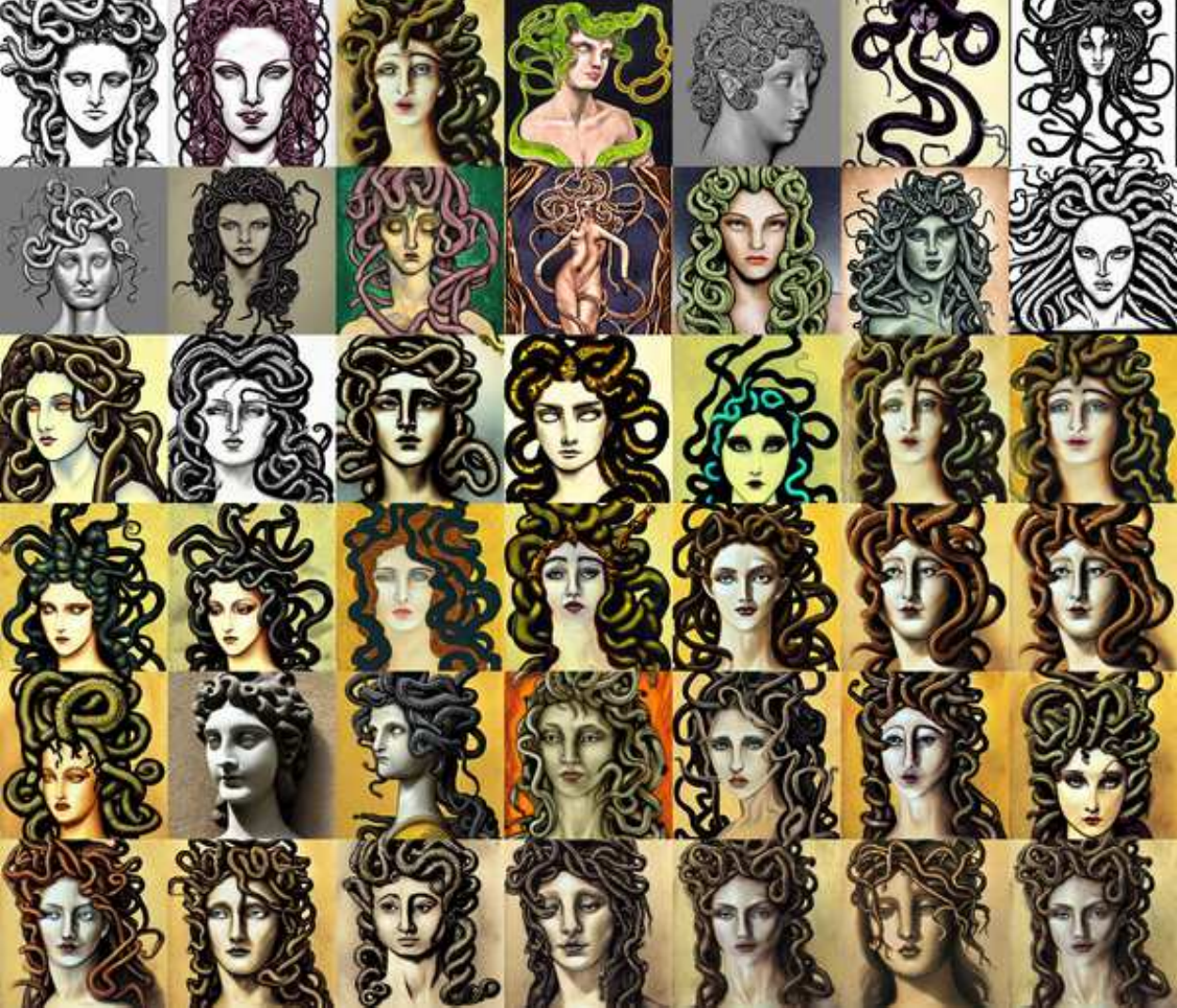}
   \caption{Medusa. Top two rows: first batch, \ie{} equivalent to default LDM. Rows 3 and 4: second generation, after clicking on the most realistic image of the first batch. Rows 5 and 6: generation 3, after clicking on the two most realistic images of generation 2. Colors become better, and sus-sternal dimples appear. We observe that a user feedback about 2 generations of images are enough for improving outputs.}\label{medusa123}
\end{figure*}

AfterLearnER was tested on three different setups:
\begin{itemize}
    \item First setup: the prompt is ``A close-up photographic portrait of a young woman with colored hair.''. We consider images 15 to 28, in 2 cases: first case, the user selects red hair; and second case, the user selects blue hair. Observation: the next generation by AfterLearnER + LDM have dramatically greater proportions of red hair (resp. blue hair). \ifthenelse{\songoku=0}{Results are presented in~\cref{xpsd}.}{}
    \item Second context: ``fighting Cthulhu in the jungle''.
    For the second setup (\Cref{zuck}, \Cref{warrior}), we present more sophisticated experiments with celebrities fighting Cthulhu. We observe a moderate improvement, on this easy setting of very famous people: some people, like X Y (anonymized celebrity), are easier to generate in arbitrary contexts, so that the classical LDM performs well. The prompt becomes much harder for a scientist like Yann LeCun: \change{in this harder setting, }the improvement becomes much greater \change{(\cref{ylc})}.%
    \item Third setup,
    we now check that AfterLearnER can also have an impact on realism. We just ask for ``Medusa'' and feed AfterLearnER with the realism opinion of the user. The second batch uses preferences mentioned by the user in the first batch, then the third batch uses preferences mentioned by the user in the second batch, and so on. Each batch improves over the previous one.
    Within 3 batches of 14 images, we get way more realistic portraits of Medusa~(\Cref{medusa123}).
\end{itemize}
Results are aggregated in~\cref{summarytable}. They are always positive, with a clear gap in 3 out of 4 contexts.
\subsubsection{Discussion}In each setup, we increase the proportion of images that fit specific criteria. This approach is particularly effective for generating images that are difficult to produce by default, such as ``Yann LeCun fighting tentacles in the Jungle.'' In these challenging scenarios, the text-to-image model struggles to combine elements like Yann LeCun, a jungle, and tentacles. However, by selecting the right images, we can guide the model to produce outputs that better match user preferences, showing the effectiveness of the proposed approach.

\begin{table}
  \centering
  \caption{\label{summarytable}Overview of our low-budget image generation experiments. FB refers to user feedback (clicks on the preferred images inside a screen presenting a batch of images). For~ \cref{medusa123} the quality is not measured by quantified metrics and the reader can compare the top 2 rows to the bottom 2 rows there.}

\begin{tabular}{c|c|c|c}
\toprule
Figure          &          & Context                      & Observation \\
\midrule
            & Top      & 2nd batch, after FB      & Mostly red hair \\
\ifthenelse{\songoku=0}{Fig.}{}      &          & on red hair in batch 1       &                \\
\ifthenelse{\songoku=0}{\ref{xpsd}}{}                & Bottom   & 2nd batch, after FB      & All blue hair \\
                &          & on blue hair in batch 1      &               \\
\cmidrule{1-4}
        & Top     & Initial batch                   & Success 11/14 \\
Fig.  & Middle  & Batch 2 after FB on batch 1 & Success 12/14 \\
\ref{zuck}            & Botom   & Batch 3 after FB on         & Success 13/14 \\
            &         & batches 1 and 2                 &  \\
\cmidrule{1-4}
Fig.       & Top     & Initial batch               & Success $\simeq 30\%$ \\
\ref{ylc}  & Bottom  & After FB on the 1st batch & Success $\simeq 71\%$ \\
\cmidrule{1-4}
Fig.             & Top    & Initial batch               & Low quality \\
\ref{medusa123}  & Middle & After FB on 1st batch       &   \\
                 & Bottom & After FB on batches 1 and 2 & High quality    \\
\bottomrule
\end{tabular}
\end{table}
\ifthenelse{\songoku=0}{
\begin{figure*}[t]\centering
\includegraphics[width=.49\textwidth]{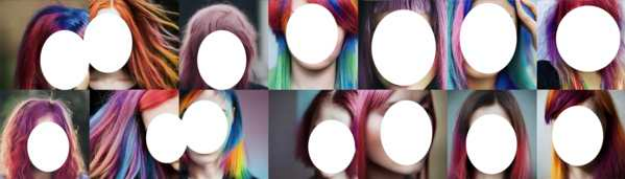}\\
The user selects red hair. Images 15 to 28, after a warm up of 14 images.\\
\includegraphics[width=.49\textwidth]{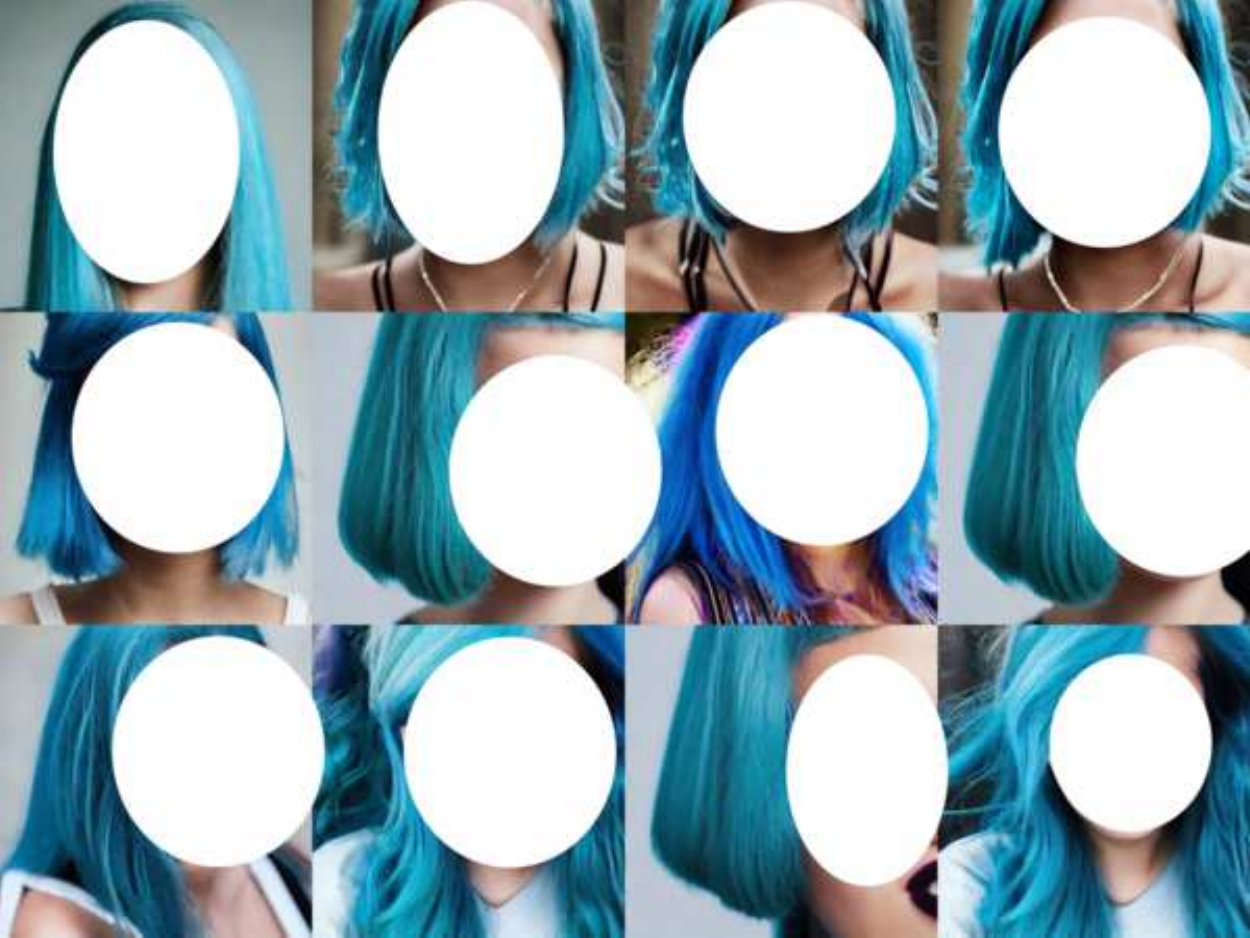}\\
The user selects blue hair. Images 15 to 26, after a warm up of 14 images.\\
\caption{\label{xpsd}LDM with prompt ``A close up photographic portrait of a young woman with colored hair''. We present generated images, after a warm up over 14 images. The only difference between top and bottom is that the user clicks differently during the first batch. We observe variations in the necklaces and eyes when the rest of the image looks similar. In spite of the very low budget (the human user provides feedback on 14 images only) there is a clear increase in the target hair color.}%
\end{figure*}
}{}
\ifthenelse{\songoku=0}{
\begin{figure}[t]\centering
\includegraphics[width=.32\textwidth]{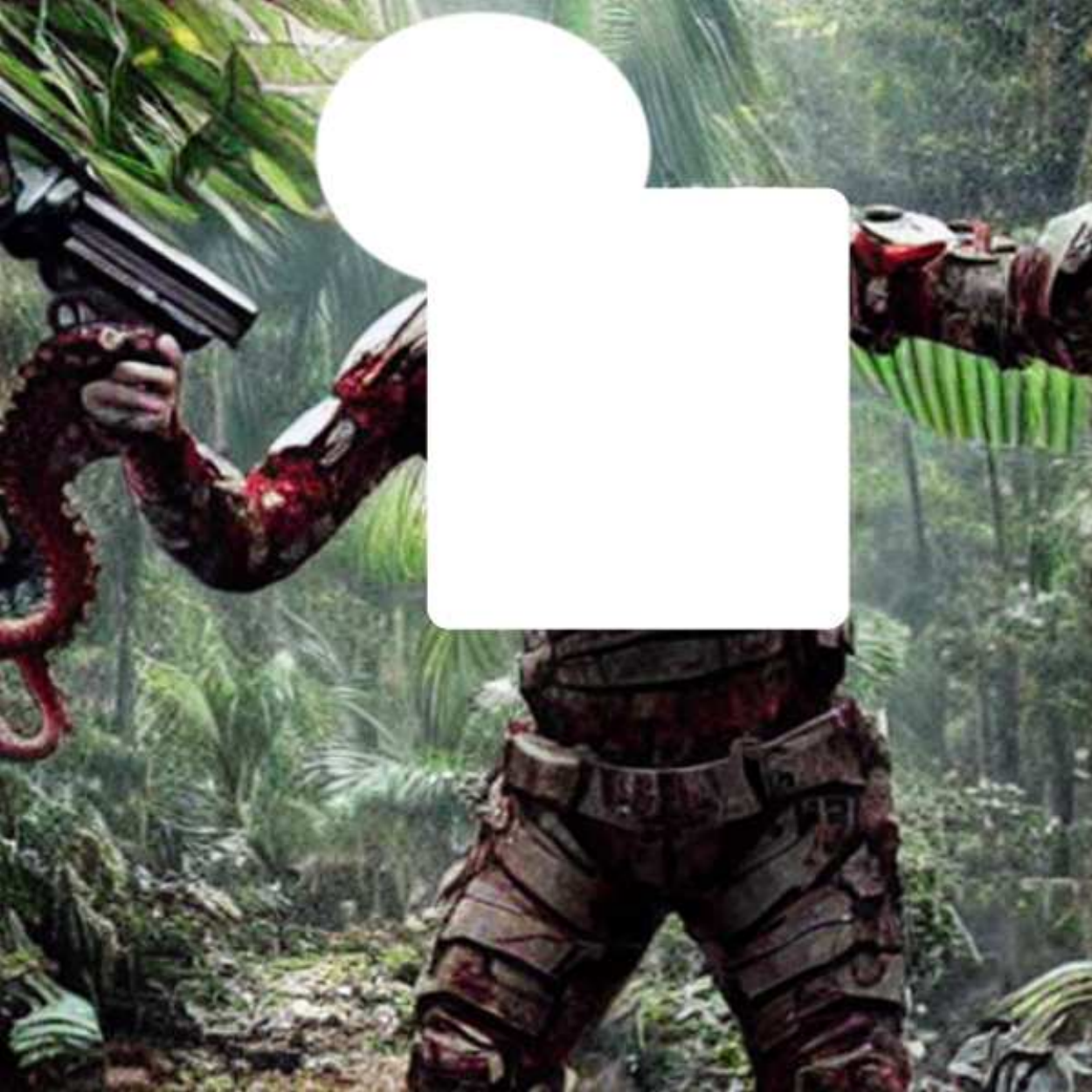}
\includegraphics[width=.32\textwidth]{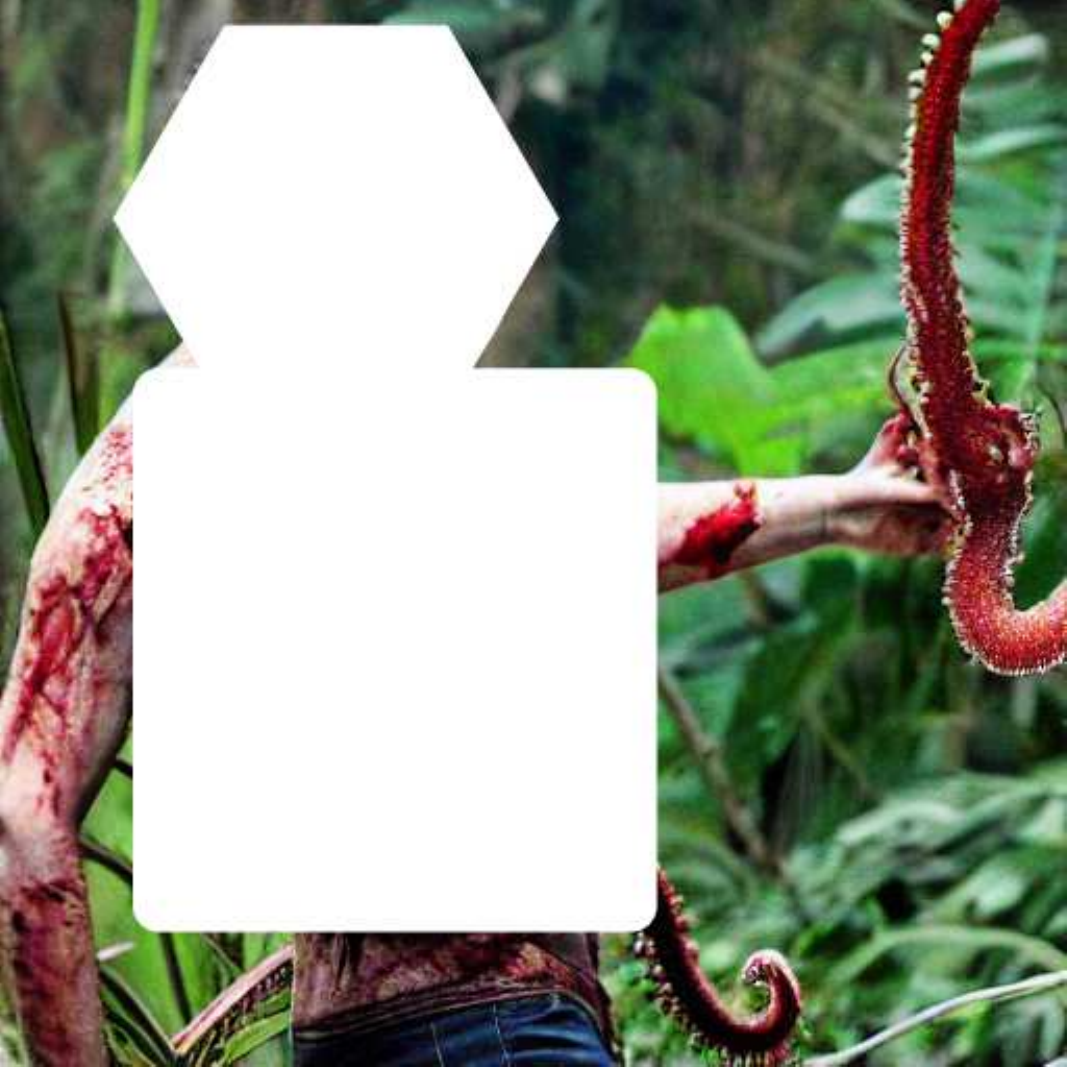}
\caption{\label{warrior}XY (anonymized) fighting tentacles in the jungle, selection of best images, less than 50 runs of LDM, one minute per LDM run with default parameters on a Mac M1.}
\end{figure}}{}

\section{Overview and discussion}\label{discussion}
\subsection{\change{Overview}}
All results in the present paper come from applying {AfterLearnER after a classical ML pipeline}.

{\bf{Depth sensing}} (\cref{midas}): We use the small MiDaS model and optimize \hp{}s on a small dataset. More precisely, 300 images are used for optimizing \hp{}s by distillation of the big MiDaS model so that we get test errors better than the original Small MiDaS, or even (in some cases) the Hybrid MiDaS. These results only require 50 scalars of feedback for adapting a model to different loss functions. In this benchmark, the loss is not differentiable either.

{\bf{Speech resynthesis}} (\cref{spre}): We use the checkpoint previously tuned in~ \cite{resynt} and optimize two \hp{} on 75\% of the test set and check on the remaining test data. {This is} a simple experiment with just two constants, hardcoded in the original code, optimized on a small dataset. {The numbers are not directly comparable to the original work as we use a split of the test set: but we show that a small feedback (40 scalars) on a split of the test set is enough for, on average, an improvement on the rest of the test set.}
Compared to~\cite{reinforceseq}, we do a fast distribution shift and not a whole training.

{\bf{Doom/Arnold}} (\cref{doomxp}): In this reinforcement learning context, we use the same simulator as the one used in the original paper/code. This consistency allows for a direct comparison of our results. We observed enhanced outcomes primarily because we focused on optimizing the actual objective function. 

{\bf{Transcoder}} (\cref{ct}): We use the same training/validation data as in the GitHub repository of~\cite{unsuptran}. We immediately observe an improvement~(\Cref{figcodegen}) only using a few hundred scalars. We do not use any information about the objective function, so we could have an online estimator such as speed, memory consumption or elegance~\textemdash{} which cannot be done with the original supervised learning.

\ifthenelse{\songoku=0}{
{\bf{Interactive GAN}} (\cref{egan}): The results here are close to those in~\cite{evogan}, but obtained faster using discrete optimization \cite{lengler} and tools \cite{mathuringecco}, and we improved diversity by optimizing the initialization as in~\cref{fairalg}. Experimental results show that eight clicks are sufficient for most cases, with three clicks (on average) for our chosen classes.
}{}

{\bf{3D Gan EG3D for AFHQ (cats)}} (\cref{eg3dcase}): With 200 feedbacks, we built a more realistic EG3D-cats, outperforming state-of-the-art in three-dimensional cats\ifthenelse{\songoku=0}{ and three-dimensional face generation}{}.

{\bf{LDM}} (\cref{sd}) : We provide an interactive counterpart of LDM. The fact that it works is quite counter-intuitive: the latent space has dimension $4\cdot 64\cdot 64$. We perform several tests with hair color, image quality, and realism, and results are \change{repeatedly confirmed}. They are in line with user needs: whereas most LDM users (in particular in the new field of prompt engineering for txt2img) do many independent rerolls, we guide these rerolls using previous results. \change{Section \ref{sec:onlinesd} presents results in a very low budget case.}

\subsection{\change{Discussion}}
Before detailing the experimental validation, we take a look at the a priori benefits of retrofitting on the problems described in \cref{overview}, according to the rationale developed in \cref{sec:rationale}.

\Cref{overview} shows, in particular, the different sizes of the feedback used in the following. It can always be provided by humans or computationally expensive solvers. Regarding RLHF (Reinforcement Learning with Human Feedback -- such as EG3D, LDM\ifthenelse{\songoku=0}{, Facial Composites}{}), we consider feedback limited to dozens or hundreds of scalars (far less than most works as discussed in~\cref{sec:rlhf}).%

For \ifthenelse{\songoku=0}{facial composites~(\cref{egan}),}{} speech re-synthesis~(\cref{spre}) and depth sensing~(\cref{midas}), the number of runs is $k=1$ because of the computational cost (for depth and speech) or limited human attention span (for \ifthenelse{\songoku=1}{image generation}{facial composites and latent diffusion models}).  

Our approach requires less data and utilizes feedback at a more aggregated level, compared to most RLHF~\cite{rlhf} and Human In The Loop Reinforcement Learning (HITLRL~\cite{hitlrl}) approaches.   Specifically, our approach needs only one objective function value per parametrized model and dataset. In contrast, classical stochastic-gradient methods require one objective function value and one gradient value per parameterized model and per data point. The process is hence compatible with evaluation by humans, who can study an entire system output, and compare on key points (see \eg{} human feedback \ifthenelse{\songoku=0}{for interactive GAN \cref{egan}}{}), or with end-of-episode (delayed) evaluations by complex simulators (\eg{} VizDoom~\cite{vizdoom} for Doom, in \cref{doomxp}) or non-differentiable computational accuracy (in code translation,~\cref{ct}).

\section{Conclusions}\label{sec:conclusion}

This paper demonstrates the potential of integrating small-scale feedback directly into the optimization process, as in retrofitting, through black-box optimization approaches. Likewise, such a strategy is especially beneficial for optimizing a few scalars using a non-differentiable \change{criterion}, a common challenge in various fields. 
That way, we provide substantial improvements in state-of-the-art Deep Learning~\change{(DL)} applications, \eg{} \change{ use cases } such as Doom and code translation. Furthermore, our methodology enables the generation \ifthenelse{\songoku=0}{of facial composites}{images} with minimal user interaction, requiring as few as three interactions for getting a first approximation.
Similarly, we optimize \hp{}s for depth estimation, language translation, and speech synthesis. Our method requires only dozens to hundreds of~(\eg{} human) evaluations to achieve significant improvements. Additionally, we have enhanced the EG3D-cats model with a streamlined codebase, illustrating our approach's versatility and practicality~(see~\cref{additionalcode} in appendix F). Overall, our findings underscore the value of human-in-the-loop methodologies or simulation-based feedbacks in refining DL solutions across various domains without relying on \change{ thousands} of feedbacks.

 \change{We make the following recommendations for the use of  AfterLearnER}:

\begin{itemize}
	\item Consider problems in which the actual objective is not differentiable and is hence typically approximated for gradient-based training by various proxies, or by ad-hoc reward shaping.
	\item \change{Apply AfterLearER} {\em{after}} the classical gradient-based training, and without any backprop retraining.
	\item Restrict the \hp\ to key parameters:  typically, some parameters at the input or output transformations, or a single layer, as small as possible. 
	 Dimensions in the 1000s is manageable in a black-box manner (though this, of course, depends on the budget), but we prefer as much as possible small families of parameters that intersect all paths from input to output. %
        \item Increasing the number of runs (parameter $k$), or the parallelism $\lambda$, makes the code closer to random search, and less prone to overfitting, as \change{ predicted} by the theoretical analysis of \cref{maths} \change{and confirmed in the experiments}. %
\end{itemize}

We emphasize the concept of small aggregated anytime feedback (\cref{sec:SAAF}). While we did a part of our experiments from automatically computed feedback (\eg{} word error rate based on programs for speech synthesis \textcolor{black}{or the biggest MiDaS model in depth estimation}), AfterLearnER uses an aggregated feedback, which could be provided by humans or expensive simulators, based on a global experience of the system, without the need for fine grain feedback. Also, this shows a clear generalization: compared to systems using millions of fine-grained feedbacks, our system can not so easily ``cheat'' (overfit) using redundancies between train/validation and testing.

\subsection*{Limitations and further work}

Most of the feature construction was performed by DL: what we propose can be added on top of a DL solution, but does not replace it. We outperform Arnold for Doom, but only with Arnold and its reward shaping as an excellent starting point. The situation is similar for speech resynthesis, interactive generation, and depth sensing. In the case of the Transcoder, we manage to optimize a few weights of the checkpoint within a few dozen evaluations. %

Maybe some of our results could be improved by applying AfterLearnER on several layers in turn: in the present paper, we only heuristically guessed which parameters to optimize and did not investigate larger search spaces. This might be particularly true for cases in which we can generate new data with a simulator, as in the case of Doom (\cref{doomxp}).

Improving generalization by searching for flat optima is a challenging topic~\cite{flat1,flat2,flat3}, mostly explored for gradient-based optimization of weights: our context (black-box optimization, validation set, after training) is specific, promising, and deserves exploration. This includes the theory in~\cref{maths}.%

Multi-objective extensions seem to be natural for Arnold/Doom (using the various reward shaping criteria used in the original code) and for code translation, where BLEU and comp. acc. are completely different measures, and for the many criteria of depth sensing: \change{there exist many multi-objective variants of black-box optimizers.}%

The present paper demonstrated the applicability of AfterLearnER, with experiments reproducible with a relatively low budget. \change{Some obvious further works include} the large-scale application on unsolved problems, such as \change{fighting} hallucinations in large language models or optimizing code translation for speed.    %
\ifthenelse{\songoku=0}{
\begin{figure*}
    \begin{subfigure}{.3\textwidth}
      \includegraphics[width=\linewidth]{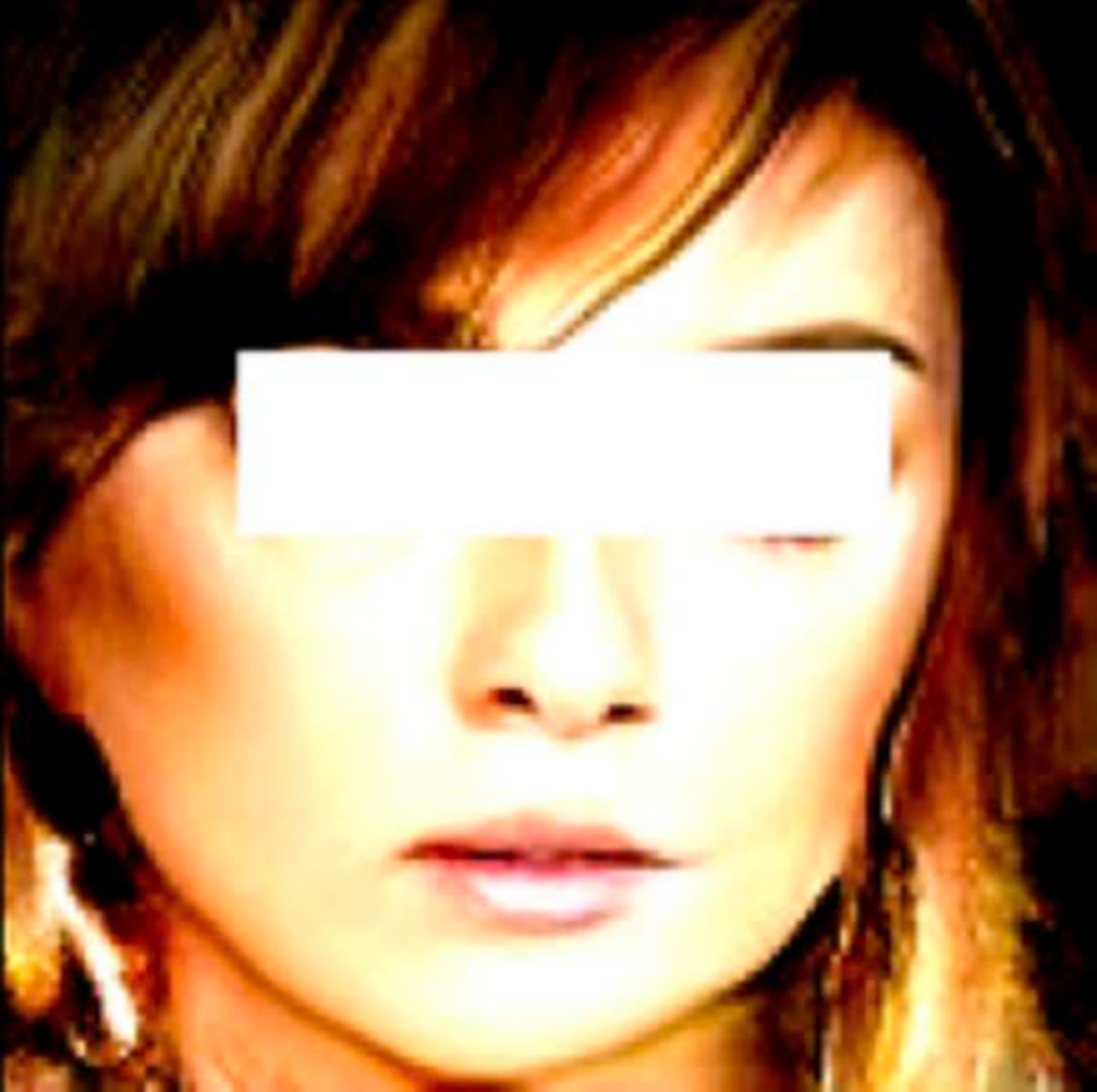}
      \caption{Female Asian}      
    \end{subfigure}
    \begin{subfigure}{.3\textwidth}
      \includegraphics[width=\linewidth]{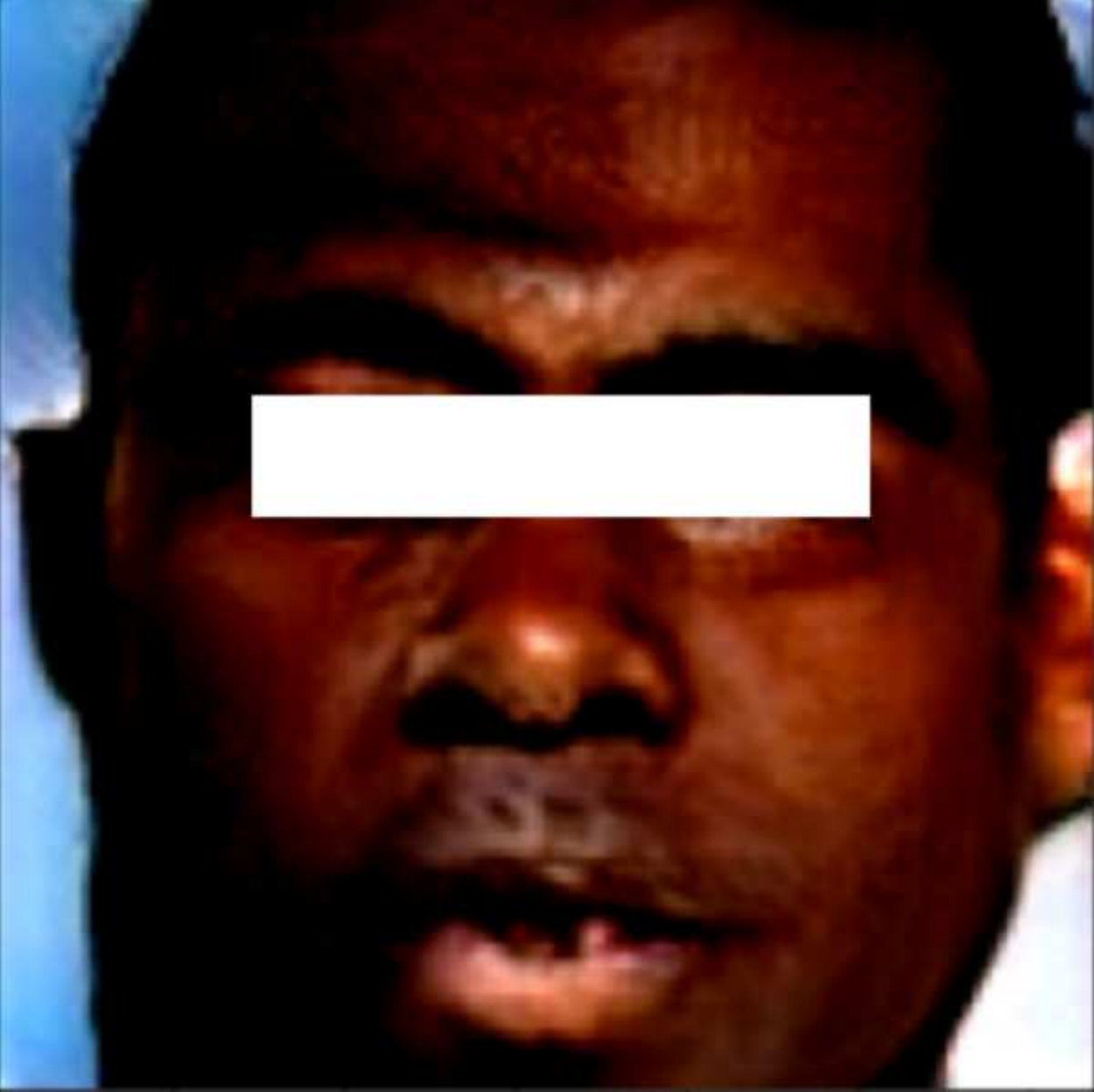}
      \caption{Male Black}
    \end{subfigure}
    \begin{subfigure}{.3\textwidth}
      \includegraphics[width=\linewidth]{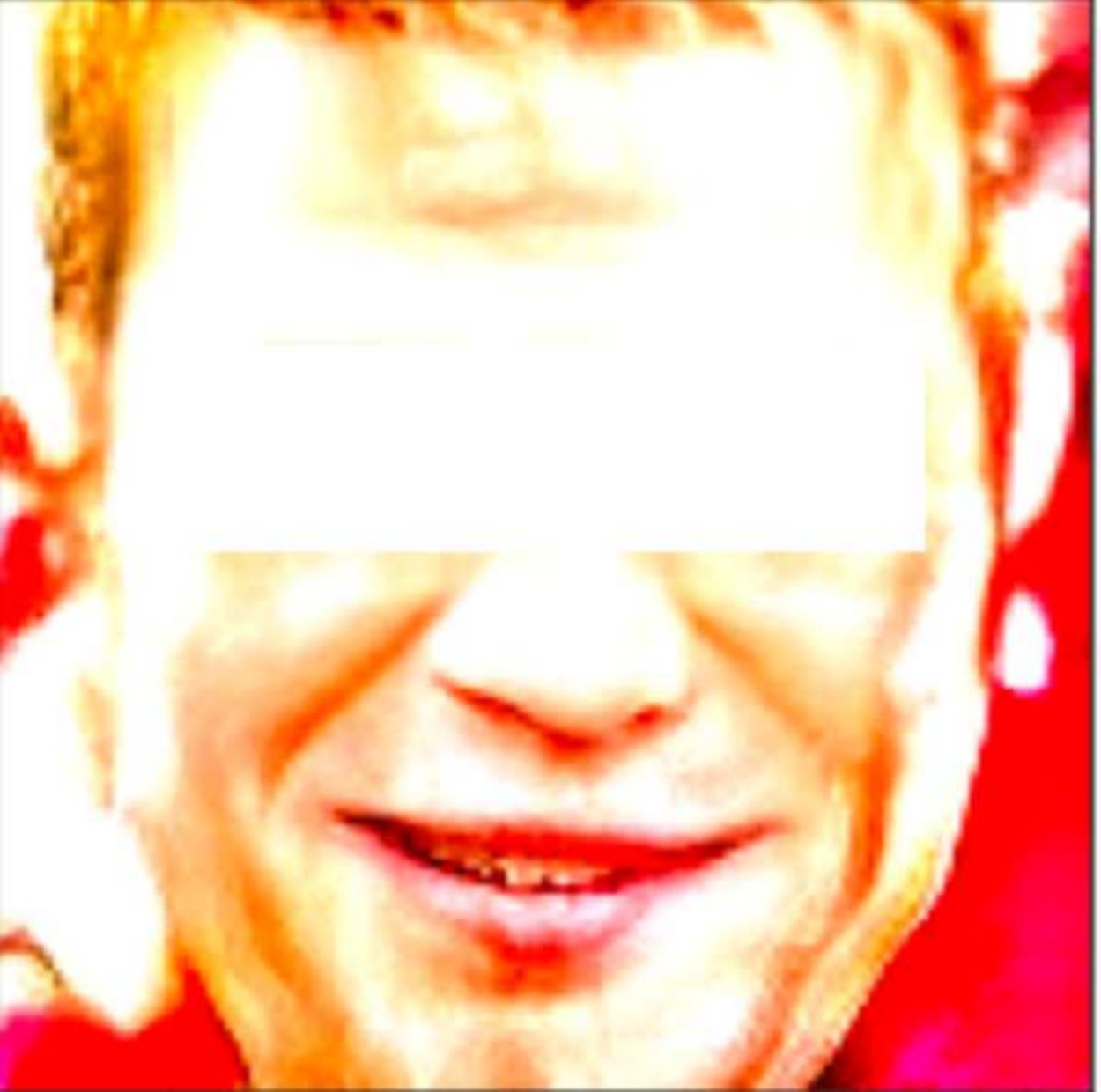}
      \caption{Blonde male}
    \end{subfigure}
    \begin{subfigure}{.3\textwidth}
       \includegraphics[width=\linewidth]{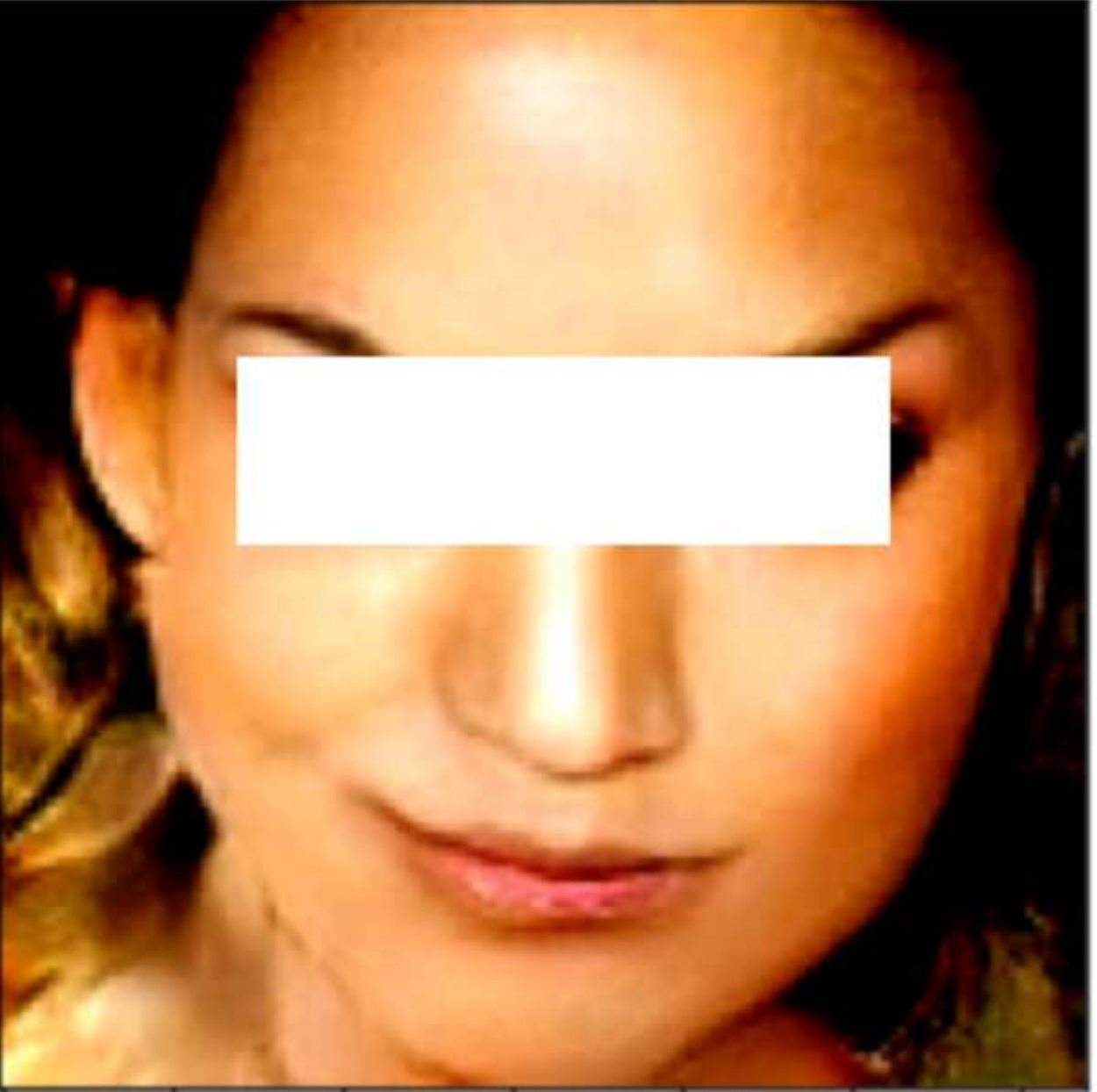}
       \caption{Sarcastic female~(4)}
    \end{subfigure}
    \begin{subfigure}{.3\textwidth}
       \includegraphics[width=\linewidth]{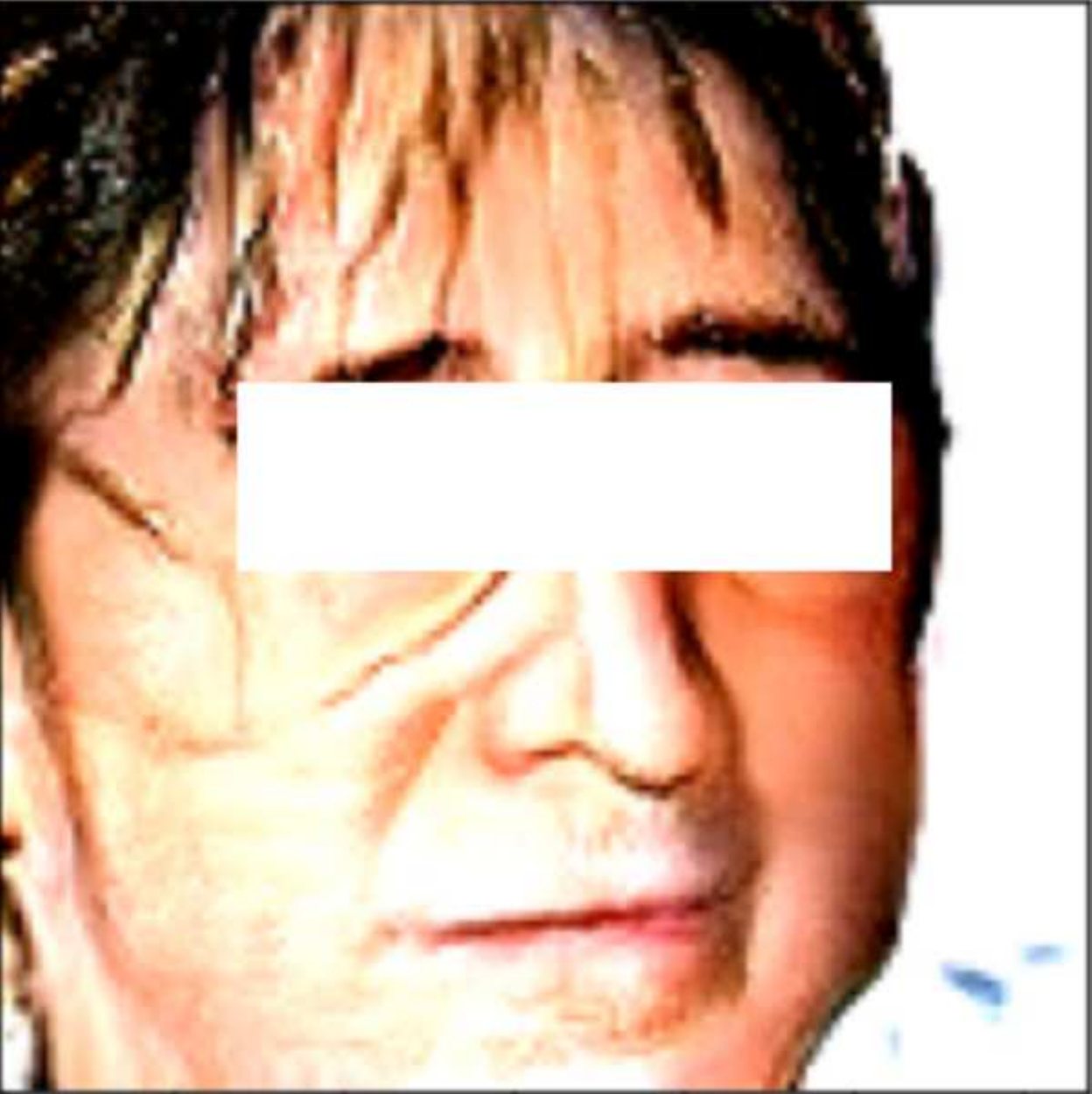}
       \caption{Sad man~(3)}
    \end{subfigure}
    \begin{subfigure}{.3\textwidth}
       \includegraphics[width=\linewidth]{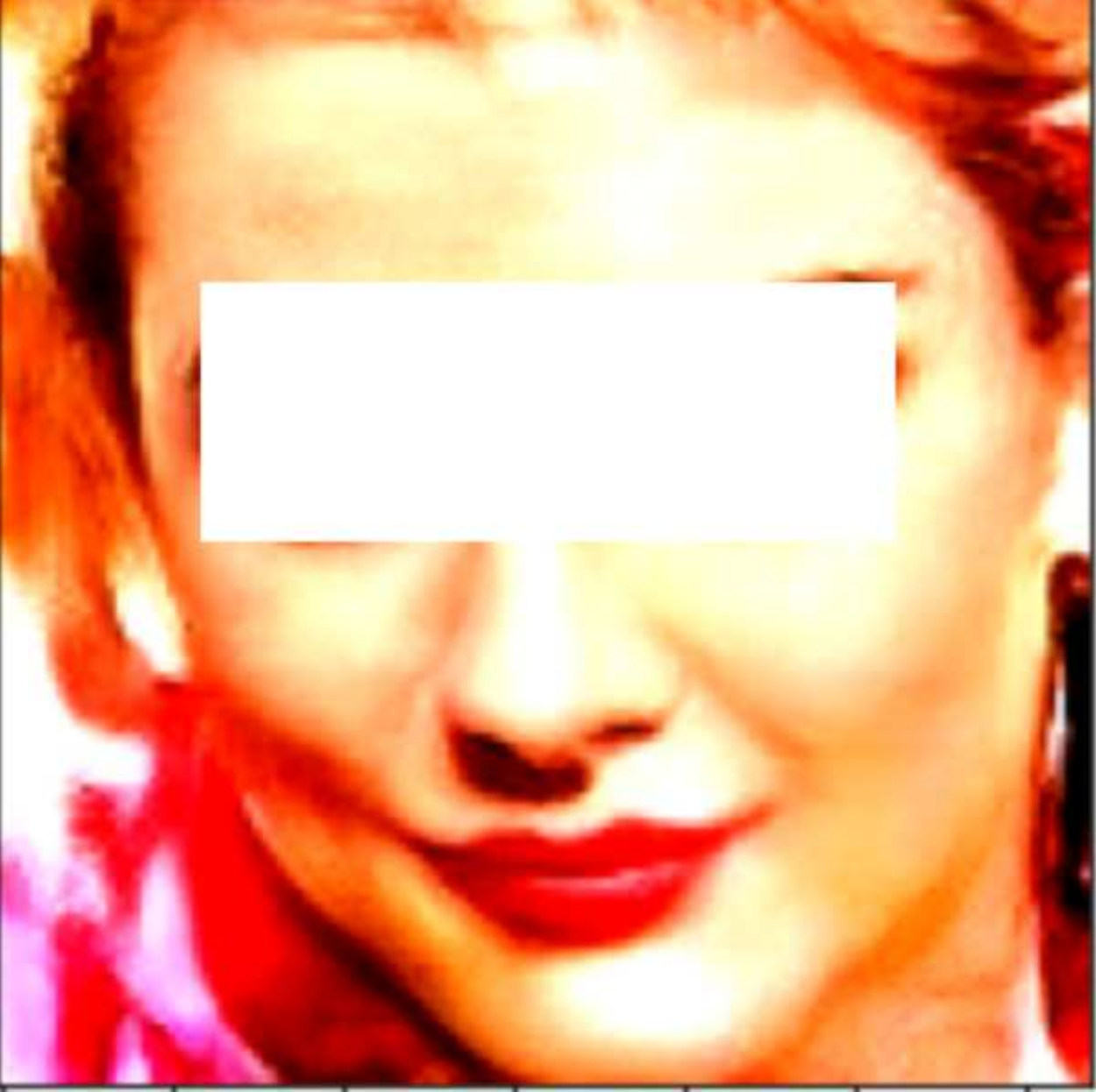}
       \caption{Red haired female~(2)}
    \end{subfigure}
    	\caption{First 3 images (top, and middle left): examples corresponding to female Asian, male black, and blonde male, respectively, \ie{} 3 categories used in our experiment. The 3 other images are test cases with labels not provided in the tuning in~\cref{fairalg}: they show the versatility of the method. The number in parentheses is the number of clicks before obtaining the result. None of the faces presented here corresponds to a real-world person: these faces are obtained by optimization through~\cref{algevolgan}.}\label{examplesfaces}
\end{figure*}
}{}

\section*{Acknowledgements}
The authors thank the reviewers for their insightful and constructive feedback, which has substantially improved the quality of this work. AL acknowledges support from the French National Research Agency (ANR) under Grant No. ANR-23-CPJ1-0099-01.%

\FloatBarrier
\bibliographystyle{ACM-Reference-Format}
\bibliography{otbib,references}


\begin{thebibliography}{120}


\ifx \showCODEN    \undefined \def \showCODEN     #1{\unskip}     \fi
\ifx \showDOI      \undefined \def \showDOI       #1{#1}\fi
\ifx \showISBNx    \undefined \def \showISBNx     #1{\unskip}     \fi
\ifx \showISBNxiii \undefined \def \showISBNxiii  #1{\unskip}     \fi
\ifx \showISSN     \undefined \def \showISSN      #1{\unskip}     \fi
\ifx \showLCCN     \undefined \def \showLCCN      #1{\unskip}     \fi
\ifx \shownote     \undefined \def \shownote      #1{#1}          \fi
\ifx \showarticletitle \undefined \def \showarticletitle #1{#1}   \fi
\ifx \showURL      \undefined \def \showURL       {\relax}        \fi
\providecommand\bibfield[2]{#2}
\providecommand\bibinfo[2]{#2}
\providecommand\natexlab[1]{#1}
\providecommand\showeprint[2][]{arXiv:#2}

\bibitem[Abramson et~al\mbox{.}(2022)]%
        {abra}
\bibfield{author}{\bibinfo{person}{Josh Abramson}, \bibinfo{person}{Arun
  Ahuja}, \bibinfo{person}{Federico Carnevale}, \bibinfo{person}{Petko
  Georgiev}, \bibinfo{person}{Alex Goldin}, \bibinfo{person}{Alden Hung},
  \bibinfo{person}{Jessica Landon}, \bibinfo{person}{Jirka Lhotka},
  \bibinfo{person}{Timothy Lillicrap}, \bibinfo{person}{Alistair Muldal},
  \bibinfo{person}{George Powell}, \bibinfo{person}{Adam Santoro},
  \bibinfo{person}{Guy Scully}, \bibinfo{person}{Sanjana Srivastava},
  \bibinfo{person}{Tamara von Glehn}, \bibinfo{person}{Greg Wayne},
  \bibinfo{person}{Nathaniel Wong}, \bibinfo{person}{Chen Yan}, {and}
  \bibinfo{person}{Rui Zhu}.} \bibinfo{year}{2022}\natexlab{}.
\newblock \bibinfo{title}{Improving Multimodal Interactive Agents with
  Reinforcement Learning from Human Feedback}.
\newblock
\newblock
\showeprint[arxiv]{2211.11602}


\bibitem[Adams et~al\mbox{.}(2022)]%
        {inverserl}
\bibfield{author}{\bibinfo{person}{Stephen Adams}, \bibinfo{person}{Tyler
  Cody}, {and} \bibinfo{person}{Peter~A. Beling}.}
  \bibinfo{year}{2022}\natexlab{}.
\newblock \showarticletitle{A survey of inverse reinforcement learning}.
\newblock \bibinfo{journal}{\emph{Artif. Intell. Rev.}} \bibinfo{volume}{55},
  \bibinfo{number}{6} (\bibinfo{date}{aug} \bibinfo{year}{2022}),
  \bibinfo{pages}{4307--4346}.
\newblock


\bibitem[Arakawa et~al\mbox{.}(2018)]%
        {hitlrl}
\bibfield{author}{\bibinfo{person}{Riku Arakawa}, \bibinfo{person}{Sosuke
  Kobayashi}, \bibinfo{person}{Yuya Unno}, \bibinfo{person}{Yuta Tsuboi}, {and}
  \bibinfo{person}{Shin-ichi Maeda}.} \bibinfo{year}{2018}\natexlab{}.
\newblock \showarticletitle{Dqn-tamer: Human-in-the-loop reinforcement learning
  with intractable feedback}.
\newblock \bibinfo{journal}{\emph{arXiv:1810.11748}} (\bibinfo{year}{2018}).
\newblock


\bibitem[{Artelys}(2015a)]%
        {artelyssqp}
\bibfield{author}{\bibinfo{person}{{SME} {Artelys}}.}
  \bibinfo{year}{2015}\natexlab{a}.
\newblock
\newblock
\urldef\tempurl%
\url{https://www.artelys.com/news/159/16/KNITRO-wins-the-GECCO-2015-Black-Box-Optimization-Competition}
\showURL{%
\tempurl}


\bibitem[{Artelys}(2015b)]%
        {artelysSQP2}
\bibfield{author}{\bibinfo{person}{{S}{M}{E} {Artelys}}.}
  \bibinfo{year}{2015}\natexlab{b}.
\newblock \bibinfo{title}{Artelys {SQP} Wins the {BBCOMP} competition}.
\newblock
  \bibinfo{howpublished}{\url{https://www.ini.rub.de/PEOPLE/glasmtbl/projects/bbcomp/index.html}}.
\newblock


\bibitem[Assunção et~al\mbox{.}(2019)]%
        {denserGPEM2019}
\bibfield{author}{\bibinfo{person}{Filipe Assunção}, \bibinfo{person}{Nuno
  Lourenço}, \bibinfo{person}{Penousal Machado}, {and}
  \bibinfo{person}{Bernardete Ribeiro}.} \bibinfo{year}{2019}\natexlab{}.
\newblock \showarticletitle{{DENSER}: Deep Evolutionary Network Structured
  Representation}.
\newblock \bibinfo{journal}{\emph{Genetic Programming and Evolvable Machines}}
  \bibinfo{volume}{20} (\bibinfo{date}{03} \bibinfo{year}{2019}).
\newblock


\bibitem[Awad et~al\mbox{.}(2020)]%
        {squirrel}
\bibfield{author}{\bibinfo{person}{Noor Awad}, \bibinfo{person}{Gresa Shala},
  \bibinfo{person}{Difan Deng}, \bibinfo{person}{Neeratyoy Mallik},
  \bibinfo{person}{Matthias Feurer}, \bibinfo{person}{Katharina Eggensperger},
  \bibinfo{person}{Andre' Biedenkapp}, \bibinfo{person}{Diederick Vermetten},
  \bibinfo{person}{Hao Wang}, \bibinfo{person}{Carola Doerr},
  \bibinfo{person}{Marius Lindauer}, {and} \bibinfo{person}{Frank Hutter}.}
  \bibinfo{year}{2020}\natexlab{}.
\newblock \bibinfo{title}{Squirrel: A Switching Hyperparameter Optimizer}.
\newblock
\newblock
\showeprint[arxiv]{2012.08180}


\bibitem[B\"ack et~al\mbox{.}(1997)]%
        {HEC97}
\bibfield{editor}{\bibinfo{person}{Th. B\"ack}, \bibinfo{person}{D.B. Fogel},
  {and} \bibinfo{person}{Z. Michalewicz}} (Eds.).
  \bibinfo{year}{1997}\natexlab{}.
\newblock \bibinfo{booktitle}{\emph{Handbook of Evolutionary Computation}}.
\newblock \bibinfo{publisher}{Oxford University Press}.
\newblock


\bibitem[Baevski et~al\mbox{.}(2020)]%
        {wav2vec2}
\bibfield{author}{\bibinfo{person}{Alexei Baevski}, \bibinfo{person}{Yuhao
  Zhou}, \bibinfo{person}{Abdelrahman Mohamed}, {and} \bibinfo{person}{Michael
  Auli}.} \bibinfo{year}{2020}\natexlab{}.
\newblock \showarticletitle{wav2vec 2.0: A framework for self-supervised
  learning of speech representations}.
\newblock \bibinfo{journal}{\emph{Advances in neural information processing
  systems}}  \bibinfo{volume}{33} (\bibinfo{year}{2020}),
  \bibinfo{pages}{12449--12460}.
\newblock


\bibitem[Baldassi et~al\mbox{.}(2020)]%
        {flat2}
\bibfield{author}{\bibinfo{person}{Carlo Baldassi}, \bibinfo{person}{Enrico~M.
  Malatesta}, \bibinfo{person}{Matteo Negri}, {and} \bibinfo{person}{Riccardo
  Zecchina}.} \bibinfo{year}{2020}\natexlab{}.
\newblock \showarticletitle{Wide flat minima and optimal generalization in
  classifying high-dimensional Gaussian mixtures}.
\newblock   \bibinfo{volume}{arXiv:2010.14761} (\bibinfo{year}{2020}).
\newblock


\bibitem[Berthelot et~al\mbox{.}(2017)]%
        {gan}
\bibfield{author}{\bibinfo{person}{David Berthelot}, \bibinfo{person}{Tom
  Schumm}, {and} \bibinfo{person}{Luke Metz}.} \bibinfo{year}{2017}\natexlab{}.
\newblock \showarticletitle{{BEGAN:} Boundary Equilibrium Generative
  Adversarial Networks}.
\newblock \bibinfo{journal}{\emph{CoRR}}  \bibinfo{volume}{abs/1703.10717}
  (\bibinfo{year}{2017}).
\newblock
\showeprint[arxiv]{1703.10717}


\bibitem[Beyer(1998)]%
        {mlis}
\bibfield{author}{\bibinfo{person}{Hans-Georg Beyer}.}
  \bibinfo{year}{1998}\natexlab{}.
\newblock \showarticletitle{Mutate large, but inherit small! On the analysis of
  rescaled mutations in ( $( 1,\lambda)$-ES with noisy fitness data}. In
  \bibinfo{booktitle}{\emph{Parallel Problem Solving from Nature --- PPSN V}}.
  \bibinfo{publisher}{Springer}, \bibinfo{pages}{109--118}.
\newblock


\bibitem[Birkl et~al\mbox{.}(2023)]%
        {birkl2023midas}
\bibfield{author}{\bibinfo{person}{Reiner Birkl}, \bibinfo{person}{Diana Wofk},
  {and} \bibinfo{person}{Matthias M{\"u}ller}.}
  \bibinfo{year}{2023}\natexlab{}.
\newblock \showarticletitle{Midas v3. 1--a model zoo for robust monocular
  relative depth estimation}.
\newblock \bibinfo{journal}{\emph{arXiv:2307.14460}} (\bibinfo{year}{2023}).
\newblock


\bibitem[Bonferroni(1936)]%
        {bonferroni}
\bibfield{author}{\bibinfo{person}{Carlo Bonferroni}.}
  \bibinfo{year}{1936}\natexlab{}.
\newblock \showarticletitle{Teoria statistica delle classi e calcolo delle
  probabilita}.
\newblock \bibinfo{journal}{\emph{Pubblicazioni del R Istituto Superiore di
  Scienze Economiche e Commericiali di Firenze}}  \bibinfo{volume}{8}
  (\bibinfo{year}{1936}), \bibinfo{pages}{3--62}.
\newblock


\bibitem[Bontrager et~al\mbox{.}(2018)]%
        {tog}
\bibfield{author}{\bibinfo{person}{Philip Bontrager}, \bibinfo{person}{Wending
  Lin}, \bibinfo{person}{Julian Togelius}, {and} \bibinfo{person}{Sebastian
  Risi}.} \bibinfo{year}{2018}\natexlab{}.
\newblock \showarticletitle{Deep Interactive Evolution}.
\newblock \bibinfo{journal}{\emph{7th International Conference on Computational
  Intelligence in Music, Sound, Art and Design}} (\bibinfo{year}{2018}).
\newblock


\bibitem[Burkholz(2024)]%
        {small3}
\bibfield{author}{\bibinfo{person}{Rebekka Burkholz}.}
  \bibinfo{year}{2024}\natexlab{}.
\newblock \showarticletitle{Batch normalization is sufficient for universal
  function approximation in {CNN}s}. In \bibinfo{booktitle}{\emph{The 12th
  International Conference on Learning Representations}}.
\newblock


\bibitem[Cauwet and Teytaud(2016)]%
        {cauwetnoise}
\bibfield{author}{\bibinfo{person}{Marie-Liesse Cauwet} {and}
  \bibinfo{person}{Olivier Teytaud}.} \bibinfo{year}{2016}\natexlab{}.
\newblock \showarticletitle{Noisy Optimization: Fast Convergence Rates with
  Comparison-Based Algorithms}. In \bibinfo{booktitle}{\emph{Genetic and
  Evolutionary Computation Conference}}. \bibinfo{pages}{1101--1106}.
\newblock


\bibitem[Chan et~al\mbox{.}(2022)]%
        {eg3d}
\bibfield{author}{\bibinfo{person}{Eric~R. Chan}, \bibinfo{person}{Connor~Z.
  Lin}, \bibinfo{person}{Matthew~A. Chan}, \bibinfo{person}{Koki Nagano},
  \bibinfo{person}{Boxiao Pan}, \bibinfo{person}{Shalini~De Mello},
  \bibinfo{person}{Orazio Gallo}, \bibinfo{person}{Leonidas Guibas},
  \bibinfo{person}{Jonathan Tremblay}, \bibinfo{person}{Sameh Khamis},
  \bibinfo{person}{Tero Karras}, {and} \bibinfo{person}{Gordon Wetzstein}.}
  \bibinfo{year}{2022}\natexlab{}.
\newblock \showarticletitle{Efficient Geometry-aware {3D} Generative
  Adversarial Networks}. In \bibinfo{booktitle}{\emph{CVPR}}.
\newblock


\bibitem[Chen et~al\mbox{.}(2022)]%
        {contrastiveTTA2022}
\bibfield{author}{\bibinfo{person}{Dian Chen}, \bibinfo{person}{Dequan Wang},
  \bibinfo{person}{Trevor Darrell}, {and} \bibinfo{person}{Sayna Ebrahimi}.}
  \bibinfo{year}{2022}\natexlab{}.
\newblock \showarticletitle{Contrastive Test-Time Adaptation}. In
  \bibinfo{booktitle}{\emph{IEEE/CVF Conference on Computer Vision and Pattern
  Recognition}}. \bibinfo{pages}{295--305}.
\newblock


\bibitem[Chiu et~al\mbox{.}(2019)]%
        {Chiu2019}
\bibfield{author}{\bibinfo{person}{Billy Chiu}, \bibinfo{person}{Simon Baker},
  \bibinfo{person}{Martha Palmer}, {and} \bibinfo{person}{Anna Korhonen}.}
  \bibinfo{year}{2019}\natexlab{}.
\newblock \showarticletitle{Enhancing biomedical word embeddings by
  retrofitting to verb clusters}. In \bibinfo{booktitle}{\emph{BioNLP@ACL}}.
\newblock


\bibitem[Chiu et~al\mbox{.}(2015)]%
        {repeatde}
\bibfield{author}{\bibinfo{person}{Shih-Yuan Chiu}, \bibinfo{person}{Ching-Nung
  Lin}, \bibinfo{person}{Jialin Liu}, \bibinfo{person}{Tsang-Cheng Su},
  \bibinfo{person}{Fabien Teytaud}, \bibinfo{person}{Olivier Teytaud}, {and}
  \bibinfo{person}{Shi-Jim Yen}.} \bibinfo{year}{2015}\natexlab{}.
\newblock \showarticletitle{{Differential Evolution for Strongly Noisy
  Optimization: Use 1.01$^n$ Resamplings at Iteration n and Reach the -1/2
  Slope}}. In \bibinfo{booktitle}{\emph{{IEEE Congress on Evolutionary
  Computation}}}.
\newblock


\bibitem[Dang and Lehre(2016)]%
        {danglehre}
\bibfield{author}{\bibinfo{person}{Duc{-}Cuong Dang} {and}
  \bibinfo{person}{Per~Kristian Lehre}.} \bibinfo{year}{2016}\natexlab{}.
\newblock \showarticletitle{Self-adaptation of Mutation Rates in Non-elitist
  Populations}. In \bibinfo{booktitle}{\emph{Parallel Problem Solving from
  Nature - {PPSN} {XIV} - 14th International Conference}}.
  \bibinfo{pages}{803--813}.
\newblock


\bibitem[Dawson(2007)]%
        {dawson:07}
\bibfield{author}{\bibinfo{person}{Richard Dawson}.}
  \bibinfo{year}{2007}\natexlab{}.
\newblock \showarticletitle{Re-engineering cities: a framework for adaptation
  to global change}.
\newblock \bibinfo{journal}{\emph{Philosophical Transactions of the Royal
  Society A: Mathematical, Physical and Engineering Sciences}}
  \bibinfo{volume}{365}, \bibinfo{number}{1861} (\bibinfo{year}{2007}),
  \bibinfo{pages}{3085--3098}.
\newblock


\bibitem[Dixon and Eames(2013)]%
        {dixon:13}
\bibfield{author}{\bibinfo{person}{Tim Dixon} {and} \bibinfo{person}{Malcolm
  Eames}.} \bibinfo{year}{2013}\natexlab{}.
\newblock \showarticletitle{Scaling up: the challenges of urban retrofit}.
\newblock \bibinfo{journal}{\emph{Building research \& information}}
  \bibinfo{volume}{41}, \bibinfo{number}{5} (\bibinfo{year}{2013}),
  \bibinfo{pages}{499--503}.
\newblock


\bibitem[Doerr et~al\mbox{.}(2019)]%
        {lengler}
\bibfield{author}{\bibinfo{person}{Benjamin Doerr}, \bibinfo{person}{Carola
  Doerr}, {and} \bibinfo{person}{Johannes Lengler}.}
  \bibinfo{year}{2019}\natexlab{}.
\newblock \showarticletitle{Self-Adjusting Mutation Rates with Provably Optimal
  Success Rules}. In \bibinfo{booktitle}{\emph{Genetic and Evolutionary
  Computation Conference}}. \bibinfo{pages}{1479--1487}.
\newblock


\bibitem[Douglas(2006)]%
        {douglas:06}
\bibfield{author}{\bibinfo{person}{James Douglas}.}
  \bibinfo{year}{2006}\natexlab{}.
\newblock \bibinfo{booktitle}{\emph{Building adaptation}}.
\newblock \bibinfo{publisher}{Routledge}.
\newblock


\bibitem[Eiben and Smith(2015)]%
        {EibenSmithBook2015}
\bibfield{author}{\bibinfo{person}{A.~E. Eiben} {and} \bibinfo{person}{James~E.
  Smith}.} \bibinfo{year}{2015}\natexlab{}.
\newblock \bibinfo{booktitle}{\emph{Introduction to Evolutionary Computing,
  Second Edition}}.
\newblock \bibinfo{publisher}{Springer}.
\newblock


\bibitem[Einarsson et~al\mbox{.}(2019)]%
        {relengler}
\bibfield{author}{\bibinfo{person}{Hafsteinn Einarsson},
  \bibinfo{person}{Marcelo~Matheus Gauy}, \bibinfo{person}{Johannes Lengler},
  \bibinfo{person}{Florian Meier}, \bibinfo{person}{Asier Mujika},
  \bibinfo{person}{Angelika Steger}, {and} \bibinfo{person}{Felix
  Weissenberger}.} \bibinfo{year}{2019}\natexlab{}.
\newblock \showarticletitle{The linear hidden subset problem for the (1+1)-{EA}
  with scheduled and adaptive mutation rates}.
\newblock \bibinfo{journal}{\emph{Theoretical Computer Science}}
  \bibinfo{volume}{785} (\bibinfo{year}{2019}), \bibinfo{pages}{150--170}.
\newblock


\bibitem[Faruqui et~al\mbox{.}(2015a)]%
        {retrofitting2015}
\bibfield{author}{\bibinfo{person}{Manaal Faruqui}, \bibinfo{person}{Jesse
  Dodge}, \bibinfo{person}{Sujay~Kumar Jauhar}, \bibinfo{person}{Chris Dyer},
  \bibinfo{person}{Eduard Hovy}, {and} \bibinfo{person}{Noah~A. Smith}.}
  \bibinfo{year}{2015}\natexlab{a}.
\newblock \showarticletitle{Retrofitting Word Vectors to Semantic Lexicons}. In
  \bibinfo{booktitle}{\emph{Conference of the North {A}merican Chapter of the
  Association for Computational Linguistics: Human Language Technologies}},
  \bibfield{editor}{\bibinfo{person}{Rada Mihalcea}, \bibinfo{person}{Joyce
  Chai}, {and} \bibinfo{person}{Anoop Sarkar}} (Eds.).
  \bibinfo{pages}{1606--1615}.
\newblock


\bibitem[Faruqui et~al\mbox{.}(2015b)]%
        {faruqui-etal-2015-retrofitting}
\bibfield{author}{\bibinfo{person}{Manaal Faruqui}, \bibinfo{person}{Jesse
  Dodge}, \bibinfo{person}{Sujay~Kumar Jauhar}, \bibinfo{person}{Chris Dyer},
  \bibinfo{person}{Eduard Hovy}, {and} \bibinfo{person}{Noah~A. Smith}.}
  \bibinfo{year}{2015}\natexlab{b}.
\newblock \showarticletitle{Retrofitting Word Vectors to Semantic Lexicons}. In
  \bibinfo{booktitle}{\emph{Conference of the North {A}merican Chapter of the
  Association for Computational Linguistics: Human Language Technologies}},
  \bibfield{editor}{\bibinfo{person}{Rada Mihalcea}, \bibinfo{person}{Joyce
  Chai}, {and} \bibinfo{person}{Anoop Sarkar}} (Eds.).
  \bibinfo{pages}{1606--1615}.
\newblock


\bibitem[Feurer and Hutter(2019)]%
        {HPO-Hutter2019}
\bibfield{author}{\bibinfo{person}{Matthias Feurer} {and}
  \bibinfo{person}{Frank Hutter}.} \bibinfo{year}{2019}\natexlab{}.
\newblock \showarticletitle{Hyperparameter optimization}.
\newblock In \bibinfo{booktitle}{\emph{Automated Machine Learning: Methods,
  Systems, Challenges}}, \bibfield{editor}{\bibinfo{person}{Frank Hutter},
  \bibinfo{person}{Lars Kotthoff}, {and} \bibinfo{person}{Joaquin Vanschoren}}
  (Eds.). \bibinfo{pages}{3--38}.
\newblock


\bibitem[Fournier and Teytaud(2011)]%
        {teytaudfournier}
\bibfield{author}{\bibinfo{person}{Herv{\'{e}} Fournier} {and}
  \bibinfo{person}{Olivier Teytaud}.} \bibinfo{year}{2011}\natexlab{}.
\newblock \showarticletitle{Lower Bounds for Comparison Based Evolution
  Strategies Using VC-dimension and Sign Patterns}.
\newblock \bibinfo{journal}{\emph{Algorithmica}} \bibinfo{volume}{59},
  \bibinfo{number}{3} (\bibinfo{year}{2011}), \bibinfo{pages}{387--408}.
\newblock


\bibitem[Giannou et~al\mbox{.}(2023)]%
        {small2}
\bibfield{author}{\bibinfo{person}{Angeliki Giannou}, \bibinfo{person}{Shashank
  Rajput}, {and} \bibinfo{person}{Dimitris Papailiopoulos}.}
  \bibinfo{year}{2023}\natexlab{}.
\newblock \showarticletitle{The Expressive Power of Tuning Only the
  Normalization Layers}. In \bibinfo{booktitle}{\emph{The 36th Annual
  Conference on Learning Theory}}, \bibfield{editor}{\bibinfo{person}{Gergely
  Neu} {and} \bibinfo{person}{Lorenzo Rosasco}} (Eds.),
  Vol.~\bibinfo{volume}{195}. \bibinfo{pages}{4130--4131}.
\newblock


\bibitem[Glover and Laguna(1997)]%
        {Glover:TabuSearch97}
\bibfield{author}{\bibinfo{person}{Fred Glover} {and} \bibinfo{person}{Manuel
  Laguna}.} \bibinfo{year}{1997}\natexlab{}.
\newblock \bibinfo{booktitle}{\emph{Tabu Search}}.
\newblock \bibinfo{publisher}{Kluwer Academic Publishers}.
\newblock


\bibitem[Goldberg(1989)]%
        {goldberg:gabook89}
\bibfield{author}{\bibinfo{person}{David~E. Goldberg}.}
  \bibinfo{year}{1989}\natexlab{}.
\newblock \bibinfo{booktitle}{\emph{Genetic Algorithms in Search, Optimization
  and Machine Learning}}.
\newblock \bibinfo{publisher}{Addison-Wesley}.
\newblock


\bibitem[Goyal et~al\mbox{.}(2022)]%
        {conjugatePseudoLabelsTTA2022}
\bibfield{author}{\bibinfo{person}{Sachin Goyal}, \bibinfo{person}{Mingjie
  Sun}, \bibinfo{person}{Aditi Raghunathan}, {and} \bibinfo{person}{J.~Zico
  Kolter}.} \bibinfo{year}{2022}\natexlab{}.
\newblock \showarticletitle{Test Time Adaptation via Conjugate Pseudo-labels}.
  In \bibinfo{booktitle}{\emph{Advances in Neural Information Processing
  Systems}}, \bibfield{editor}{\bibinfo{person}{S.~Koyejo},
  \bibinfo{person}{S.~Mohamed}, \bibinfo{person}{A.~Agarwal},
  \bibinfo{person}{D.~Belgrave}, \bibinfo{person}{K.~Cho}, {and}
  \bibinfo{person}{A.~Oh}} (Eds.), Vol.~\bibinfo{volume}{35}.
  \bibinfo{pages}{6204--6218}.
\newblock


\bibitem[Hamda et~al\mbox{.}(2002)]%
        {hamda:02}
\bibfield{author}{\bibinfo{person}{Hatem Hamda},
  \bibinfo{person}{Fran{\c{c}}ois Jouve}, \bibinfo{person}{Evelyne Lutton},
  \bibinfo{person}{Marc Schoenauer}, {and} \bibinfo{person}{Michele Sebag}.}
  \bibinfo{year}{2002}\natexlab{}.
\newblock \showarticletitle{Compact unstructured representations for
  evolutionary design}.
\newblock \bibinfo{journal}{\emph{Applied Intelligence}}  \bibinfo{volume}{16}
  (\bibinfo{year}{2002}), \bibinfo{pages}{139--155}.
\newblock


\bibitem[Hansen and Ostermeier(2003)]%
        {CMA}
\bibfield{author}{\bibinfo{person}{Nikolaus Hansen} {and}
  \bibinfo{person}{Andreas Ostermeier}.} \bibinfo{year}{2003}\natexlab{}.
\newblock \showarticletitle{Completely Derandomized Self-Adaptation in
  Evolution Strategies}.
\newblock \bibinfo{journal}{\emph{Evolutionary Computation}}
  \bibinfo{volume}{11}, \bibinfo{number}{1} (\bibinfo{year}{2003}).
\newblock


\bibitem[He et~al\mbox{.}(2019)]%
        {flat1}
\bibfield{author}{\bibinfo{person}{Haowei He}, \bibinfo{person}{Gao Huang},
  {and} \bibinfo{person}{Yang Yuan}.} \bibinfo{year}{2019}\natexlab{}.
\newblock \showarticletitle{Asymmetric Valleys: Beyond Sharp and Flat Local
  Minima}. In \bibinfo{booktitle}{\emph{Advances in Neural Information
  Processing Systems}}, \bibfield{editor}{\bibinfo{person}{H.~Wallach},
  \bibinfo{person}{H.~Larochelle}, \bibinfo{person}{A.~Beygelzimer},
  \bibinfo{person}{F.~d\textquotesingle Alch\'{e}-Buc},
  \bibinfo{person}{E.~Fox}, {and} \bibinfo{person}{R.~Garnett}} (Eds.),
  Vol.~\bibinfo{volume}{32}.
\newblock


\bibitem[Heidrich-Meisner and Igel(2009)]%
        {igel}
\bibfield{author}{\bibinfo{person}{Verena Heidrich-Meisner} {and}
  \bibinfo{person}{Christian Igel}.} \bibinfo{year}{2009}\natexlab{}.
\newblock \showarticletitle{Hoeffding and Bernstein Races for Selecting
  Policies in Evolutionary Direct Policy Search}. In
  \bibinfo{booktitle}{\emph{Proceedings of the 26th Annual International
  Conference on Machine Learning}} \emph{(\bibinfo{series}{ICML '09})}.
  \bibinfo{publisher}{ACM}, \bibinfo{pages}{401--408}.
\newblock


\bibitem[Hiranandani et~al\mbox{.}(2021)]%
        {nd4}
\bibfield{author}{\bibinfo{person}{Gaurush Hiranandani}, \bibinfo{person}{Hari
  Narasimhan}, \bibinfo{person}{Jatin Mathur}, \bibinfo{person}{Mahdi~Milani
  Fard}, {and} \bibinfo{person}{Sanmi Koyejo}.}
  \bibinfo{year}{2021}\natexlab{}.
\newblock \showarticletitle{Optimizing Blackbox Metrics with Iterative Example
  Weighting}. In \bibinfo{booktitle}{\emph{38th International Conference on
  Machine Learning}}.
\newblock


\bibitem[Hoffer et~al\mbox{.}(2017)]%
        {flat3}
\bibfield{author}{\bibinfo{person}{Elad Hoffer}, \bibinfo{person}{Itay Hubara},
  {and} \bibinfo{person}{Daniel Soudry}.} \bibinfo{year}{2017}\natexlab{}.
\newblock \showarticletitle{Train Longer, Generalize Better: Closing the
  Generalization Gap in Large Batch Training of Neural Networks}. In
  \bibinfo{booktitle}{\emph{31st International Conference on Neural Information
  Processing Systems}}. \bibinfo{pages}{1729–1739}.
\newblock


\bibitem[Holland(1975)]%
        {holland}
\bibfield{author}{\bibinfo{person}{John~H. Holland}.}
  \bibinfo{year}{1975}\natexlab{}.
\newblock \bibinfo{booktitle}{\emph{Adaptation in Natural and Artificial
  Systems}}.
\newblock \bibinfo{publisher}{University of Michigan Press}.
\newblock


\bibitem[Houlsby et~al\mbox{.}(2019)]%
        {houlsby2019parameter}
\bibfield{author}{\bibinfo{person}{Neil Houlsby}, \bibinfo{person}{Andrei
  Giurgiu}, \bibinfo{person}{Stanislaw Jastrzebski}, \bibinfo{person}{Bruna
  Morrone}, \bibinfo{person}{Quentin De~Laroussilhe}, \bibinfo{person}{Andrea
  Gesmundo}, \bibinfo{person}{Mona Attariyan}, {and} \bibinfo{person}{Sylvain
  Gelly}.} \bibinfo{year}{2019}\natexlab{}.
\newblock \showarticletitle{Parameter-efficient transfer learning for NLP}. In
  \bibinfo{booktitle}{\emph{International Conference on Machine Learning}}.
  \bibinfo{pages}{2790--2799}.
\newblock


\bibitem[Hu et~al\mbox{.}(2021)]%
        {hu2021lora}
\bibfield{author}{\bibinfo{person}{Edward~J Hu}, \bibinfo{person}{Yelong Shen},
  \bibinfo{person}{Phillip Wallis}, \bibinfo{person}{Zeyuan Allen-Zhu},
  \bibinfo{person}{Yuanzhi Li}, \bibinfo{person}{Shean Wang},
  \bibinfo{person}{Lu Wang}, {and} \bibinfo{person}{Weizhu Chen}.}
  \bibinfo{year}{2021}\natexlab{}.
\newblock \showarticletitle{Lora: Low-rank adaptation of large language
  models}.
\newblock \bibinfo{journal}{\emph{arXiv:2106.09685}} (\bibinfo{year}{2021}).
\newblock


\bibitem[Huang et~al\mbox{.}(2021)]%
        {nd3}
\bibfield{author}{\bibinfo{person}{Chen Huang}, \bibinfo{person}{Shuangfei
  Zhai}, \bibinfo{person}{Pengsheng Guo}, {and} \bibinfo{person}{Josh
  Susskind}.} \bibinfo{year}{2021}\natexlab{}.
\newblock \showarticletitle{MetricOpt: Learning To Optimize Black-Box
  Evaluation Metrics}. In \bibinfo{booktitle}{\emph{IEEE/CVF Conference on
  Computer Vision and Pattern Recognition}}. \bibinfo{pages}{174--183}.
\newblock


\bibitem[Hwang et~al\mbox{.}(2021)]%
        {deathmap2}
\bibfield{author}{\bibinfo{person}{Seung-Jun Hwang}, \bibinfo{person}{Sung-Jun
  Park}, \bibinfo{person}{Gyu-Min Kim}, {and} \bibinfo{person}{Joong-Hwan
  Baek}.} \bibinfo{year}{2021}\natexlab{}.
\newblock \showarticletitle{Unsupervised Monocular Depth Estimation for
  Colonoscope System Using Feedback Network}.
\newblock \bibinfo{journal}{\emph{Sensors}} \bibinfo{volume}{21},
  \bibinfo{number}{8} (\bibinfo{year}{2021}).
\newblock


\bibitem[Jain et~al\mbox{.}(2015)]%
        {rlhfbot2}
\bibfield{author}{\bibinfo{person}{Ashesh Jain}, \bibinfo{person}{Shikhar
  Sharma}, \bibinfo{person}{Thorsten Joachims}, {and} \bibinfo{person}{Ashutosh
  Saxena}.} \bibinfo{year}{2015}\natexlab{}.
\newblock \showarticletitle{Learning preferences for manipulation tasks from
  online coactive feedback}.
\newblock \bibinfo{journal}{\emph{Int. J. Robotics Res.}} \bibinfo{volume}{34},
  \bibinfo{number}{10} (\bibinfo{year}{2015}), \bibinfo{pages}{1296--1313}.
\newblock


\bibitem[Jiang et~al\mbox{.}(2020)]%
        {nd2}
\bibfield{author}{\bibinfo{person}{Qijia Jiang}, \bibinfo{person}{Olaoluwa
  Adigun}, \bibinfo{person}{Harikrishna Narasimhan},
  \bibinfo{person}{Mahdi~Milani Fard}, {and} \bibinfo{person}{Maya Gupta}.}
  \bibinfo{year}{2020}\natexlab{}.
\newblock \showarticletitle{Optimizing Black-box Metrics with Adaptive
  Surrogates}. In \bibinfo{booktitle}{\emph{37th International Conference on
  Machine Learning}}, \bibfield{editor}{\bibinfo{person}{Hal~Daumé III} {and}
  \bibinfo{person}{Aarti Singh}} (Eds.), Vol.~\bibinfo{volume}{119}.
  \bibinfo{pages}{4784--4793}.
\newblock


\bibitem[Jiao et~al\mbox{.}(2023)]%
        {bugct}
\bibfield{author}{\bibinfo{person}{Mingsheng Jiao}, \bibinfo{person}{Tingrui
  Yu}, \bibinfo{person}{Xuan Li}, \bibinfo{person}{Guanjie Qiu},
  \bibinfo{person}{Xiaodong Gu}, {and} \bibinfo{person}{Beijun Shen}.}
  \bibinfo{year}{2023}\natexlab{}.
\newblock \bibinfo{title}{On the Evaluation of Neural Code Translation:
  Taxonomy and Benchmark}.
\newblock
\newblock
\showeprint[arxiv]{2308.08961}


\bibitem[Kandasamy et~al\mbox{.}(2018)]%
        {BO4NAS2018}
\bibfield{author}{\bibinfo{person}{Kirthevasan Kandasamy},
  \bibinfo{person}{Willie Neiswanger}, \bibinfo{person}{Jeff Schneider},
  \bibinfo{person}{Barnabás Póczos}, {and} \bibinfo{person}{Eric~P. Xing}.}
  \bibinfo{year}{2018}\natexlab{}.
\newblock \showarticletitle{Neural architecture search with Bayesian
  optimisation and optimal transport}. In \bibinfo{booktitle}{\emph{32nd
  International Conference on Neural Information Processing Systems}},
  \bibfield{editor}{\bibinfo{person}{Samy Bengio}, \bibinfo{person}{Hanna~M.
  Wallach}, \bibinfo{person}{Hugo Larochelle}, \bibinfo{person}{Kristen
  Grauman}, {and} \bibinfo{person}{Nicolò Cesa-Bianchi}} (Eds.).
  \bibinfo{pages}{2020--2029}.
\newblock


\bibitem[Karras et~al\mbox{.}(2018)]%
        {pgan}
\bibfield{author}{\bibinfo{person}{Tero Karras}, \bibinfo{person}{Timo Aila},
  \bibinfo{person}{Samuli Laine}, {and} \bibinfo{person}{Jaakko Lehtinen}.}
  \bibinfo{year}{2018}\natexlab{}.
\newblock \showarticletitle{Progressive Growing of {GAN}s for Improved Quality,
  Stability, and Variation}. In \bibinfo{booktitle}{\emph{Proceedings of
  ICLR}}.
\newblock


\bibitem[Kempka et~al\mbox{.}(2016)]%
        {vizdoom}
\bibfield{author}{\bibinfo{person}{Michał Kempka}, \bibinfo{person}{Marek
  Wydmuch}, \bibinfo{person}{Grzegorz Runc}, \bibinfo{person}{Jakub Toczek},
  {and} \bibinfo{person}{Wojciech Jaśkowski}.}
  \bibinfo{year}{2016}\natexlab{}.
\newblock \showarticletitle{{ViZDoom: A Doom-based AI Research Platform for
  Visual Reinforcement Learning}}.
\newblock \bibinfo{journal}{\emph{arXiv:1605.02097}} (\bibinfo{year}{2016}).
\newblock


\bibitem[Kennedy and Eberhart(1995)]%
        {pso}
\bibfield{author}{\bibinfo{person}{James Kennedy} {and}
  \bibinfo{person}{Russell~C. Eberhart}.} \bibinfo{year}{1995}\natexlab{}.
\newblock \showarticletitle{Particle swarm optimization}. In
  \bibinfo{booktitle}{\emph{IEEE International Conference on Neural Networks}}.
  \bibinfo{pages}{1942--1948}.
\newblock


\bibitem[Khalidov et~al\mbox{.}(2019)]%
        {vasilfoga}
\bibfield{author}{\bibinfo{person}{Vasil Khalidov}, \bibinfo{person}{Maxime
  Oquab}, \bibinfo{person}{J{\'{e}}r{\'{e}}my Rapin}, {and}
  \bibinfo{person}{Olivier Teytaud}.} \bibinfo{year}{2019}\natexlab{}.
\newblock \showarticletitle{Consistent population control: generate plenty of
  points, but with a bit of resampling}. In \bibinfo{booktitle}{\emph{15th
  {ACM/SIGEVO} Conference on Foundations of Genetic Algorithms}}.
  \bibinfo{pages}{116--123}.
\newblock


\bibitem[Kharitonov et~al\mbox{.}(2022)]%
        {resynt}
\bibfield{author}{\bibinfo{person}{Eugene Kharitonov}, \bibinfo{person}{Jade
  Copet}, \bibinfo{person}{Kushal Lakhotia}, \bibinfo{person}{Tu~Anh Nguyen},
  \bibinfo{person}{Paden Tomasello}, \bibinfo{person}{Ann Lee},
  \bibinfo{person}{Ali Elkahky}, \bibinfo{person}{Wei-Ning Hsu},
  \bibinfo{person}{Abdelrahman Mohamed}, \bibinfo{person}{Emmanuel Dupoux},
  {and} \bibinfo{person}{Yossi Adi}.} \bibinfo{year}{2022}\natexlab{}.
\newblock \showarticletitle{textless-lib: a Library for Textless Spoken
  Language Processing}.
\newblock \bibinfo{journal}{\emph{arXiv:2202.07359}} (\bibinfo{year}{2022}).
\newblock


\bibitem[Kim et~al\mbox{.}(2023)]%
        {kim2023}
\bibfield{author}{\bibinfo{person}{Changyeon Kim}, \bibinfo{person}{Jongjin
  Park}, \bibinfo{person}{Jinwoo Shin}, \bibinfo{person}{Honglak Lee},
  \bibinfo{person}{Pieter Abbeel}, {and} \bibinfo{person}{Kimin Lee}.}
  \bibinfo{year}{2023}\natexlab{}.
\newblock \showarticletitle{Preference Transformer: Modeling Human Preferences
  using Transformers for {RL}}. In \bibinfo{booktitle}{\emph{The 11th
  International Conference on Learning Representations}}.
\newblock


\bibitem[Kirkpatrick et~al\mbox{.}(1983)]%
        {sa1983}
\bibfield{author}{\bibinfo{person}{S. Kirkpatrick}, \bibinfo{person}{C.~D.
  Gelatt}, {and} \bibinfo{person}{M.~P. Vecchi}.}
  \bibinfo{year}{1983}\natexlab{}.
\newblock \showarticletitle{Optimization by Simulated Annealing}.
\newblock  \bibinfo{volume}{220}, \bibinfo{number}{4598}
  (\bibinfo{year}{1983}), \bibinfo{pages}{671--680}.
\newblock


\bibitem[Lample and Chaplot(2017)]%
        {doomai}
\bibfield{author}{\bibinfo{person}{Guillaume Lample} {and}
  \bibinfo{person}{Devendra~Singh Chaplot}.} \bibinfo{year}{2017}\natexlab{}.
\newblock \showarticletitle{Playing FPS Games with Deep Reinforcement
  Learning}. In \bibinfo{booktitle}{\emph{Thirty-First AAAI Conference on
  Artificial Intelligence}}. \bibinfo{pages}{2140–2146}.
\newblock


\bibitem[Lee et~al\mbox{.}(2023a)]%
        {small6}
\bibfield{author}{\bibinfo{person}{J. Lee}, \bibinfo{person}{D. Das},
  \bibinfo{person}{J. Choo}, {and} \bibinfo{person}{S. Choi}.}
  \bibinfo{year}{2023}\natexlab{a}.
\newblock \showarticletitle{Towards Open-Set Test-Time Adaptation Utilizing the
  Wisdom of Crowds in Entropy Minimization}. In
  \bibinfo{booktitle}{\emph{IEEE/CVF International Conference on Computer
  Vision}}. \bibinfo{pages}{16334--16334}.
\newblock


\bibitem[Lee et~al\mbox{.}(2023b)]%
        {lee2023aligning}
\bibfield{author}{\bibinfo{person}{Kimin Lee}, \bibinfo{person}{Hao Liu},
  \bibinfo{person}{Moonkyung Ryu}, \bibinfo{person}{Olivia Watkins},
  \bibinfo{person}{Yuqing Du}, \bibinfo{person}{Craig Boutilier},
  \bibinfo{person}{Pieter Abbeel}, \bibinfo{person}{Mohammad Ghavamzadeh},
  {and} \bibinfo{person}{Shixiang~Shane Gu}.} \bibinfo{year}{2023}\natexlab{b}.
\newblock \bibinfo{title}{Aligning Text-to-Image Models using Human Feedback}.
\newblock
\newblock
\showeprint[arxiv]{2302.12192}


\bibitem[Li et~al\mbox{.}(2019)]%
        {frozen}
\bibfield{author}{\bibinfo{person}{Zhengqi Li}, \bibinfo{person}{Tali Dekel},
  \bibinfo{person}{Forrester Cole}, \bibinfo{person}{Richard Tucker},
  \bibinfo{person}{Noah Snavely}, \bibinfo{person}{Ce Liu}, {and}
  \bibinfo{person}{William~T. Freeman}.} \bibinfo{year}{2019}\natexlab{}.
\newblock \showarticletitle{Learning the Depths of Moving People by Watching
  Frozen People}. In \bibinfo{booktitle}{\emph{Proceedings of the IEEE
  Conference on Computer Vision and Pattern Recognition (CVPR)}}.
\newblock


\bibitem[Lialin et~al\mbox{.}(2023)]%
        {fineTuningSurvey2023}
\bibfield{author}{\bibinfo{person}{Vladislav Lialin}, \bibinfo{person}{Vijeta
  Deshpande}, {and} \bibinfo{person}{Anna Rumshisky}.}
  \bibinfo{year}{2023}\natexlab{}.
\newblock \showarticletitle{Scaling down to scale up: A guide to
  parameter-efficient fine-tuning}.
\newblock \bibinfo{journal}{\emph{arXiv:2303.15647}} (\bibinfo{year}{2023}).
\newblock


\bibitem[Liu et~al\mbox{.}(2020)]%
        {versatile}
\bibfield{author}{\bibinfo{person}{Jialin Liu}, \bibinfo{person}{Antoine
  Moreau}, \bibinfo{person}{Mike Preuss}, \bibinfo{person}{Jeremy Rapin},
  \bibinfo{person}{Baptiste Roziere}, \bibinfo{person}{Fabien Teytaud}, {and}
  \bibinfo{person}{Olivier Teytaud}.} \bibinfo{year}{2020}\natexlab{}.
\newblock \showarticletitle{Versatile Black-Box Optimization}. In
  \bibinfo{booktitle}{\emph{Genetic and Evolutionary Computation Conference}}.
  \bibinfo{pages}{620--628}.
\newblock


\bibitem[Liu et~al\mbox{.}(2024)]%
        {liu2024dora}
\bibfield{author}{\bibinfo{person}{Shih-Yang Liu}, \bibinfo{person}{Chien-Yi
  Wang}, \bibinfo{person}{Hongxu Yin}, \bibinfo{person}{Pavlo Molchanov},
  \bibinfo{person}{Yu-Chiang~Frank Wang}, \bibinfo{person}{Kwang-Ting Cheng},
  {and} \bibinfo{person}{Min-Hung Chen}.} \bibinfo{year}{2024}\natexlab{}.
\newblock \showarticletitle{Dora: Weight-decomposed low-rank adaptation}.
\newblock \bibinfo{journal}{\emph{arXiv:2402.09353}} (\bibinfo{year}{2024}).
\newblock


\bibitem[Lu et~al\mbox{.}(2022)]%
        {small1}
\bibfield{author}{\bibinfo{person}{Kevin Lu}, \bibinfo{person}{Aditya Grover},
  \bibinfo{person}{Pieter Abbeel}, {and} \bibinfo{person}{Igor Mordatch}.}
  \bibinfo{year}{2022}\natexlab{}.
\newblock \showarticletitle{Frozen Pretrained Transformers as Universal
  Computation Engines}.
\newblock \bibinfo{journal}{\emph{{AAAI} Conference on Artificial
  Intelligence}} \bibinfo{volume}{36}, \bibinfo{number}{7}
  (\bibinfo{date}{Jun.} \bibinfo{year}{2022}), \bibinfo{pages}{7628--7636}.
\newblock


\bibitem[Ma and Karaman(2018)]%
        {deathmap3}
\bibfield{author}{\bibinfo{person}{Fangchang Ma} {and} \bibinfo{person}{Sertac
  Karaman}.} \bibinfo{year}{2018}\natexlab{}.
\newblock \showarticletitle{Sparse-to-dense: Depth prediction from sparse depth
  samples and a single image}. In \bibinfo{booktitle}{\emph{IEEE international
  conference on robotics and automation}}. \bibinfo{pages}{4796--4803}.
\newblock


\bibitem[Meunier et~al\mbox{.}(2022a)]%
        {NGOpt}
\bibfield{author}{\bibinfo{person}{Laurent Meunier},
  \bibinfo{person}{Herilalaina Rakotoarison}, \bibinfo{person}{Pak{-}Kan Wong},
  \bibinfo{person}{Baptiste Rozi{\`{e}}re}, \bibinfo{person}{J{\'{e}}r{\'{e}}my
  Rapin}, \bibinfo{person}{Olivier Teytaud}, \bibinfo{person}{Antoine Moreau},
  {and} \bibinfo{person}{Carola Doerr}.} \bibinfo{year}{2022}\natexlab{a}.
\newblock \showarticletitle{Black-Box Optimization Revisited: Improving
  Algorithm Selection Wizards Through Massive Benchmarking}.
\newblock \bibinfo{journal}{\emph{{IEEE} Trans. Evol. Comput.}}
  \bibinfo{volume}{26}, \bibinfo{number}{3} (\bibinfo{year}{2022}),
  \bibinfo{pages}{490--500}.
\newblock


\bibitem[Meunier et~al\mbox{.}(2022b)]%
        {iclrbb}
\bibfield{author}{\bibinfo{person}{Laurent Meunier},
  \bibinfo{person}{Herilalaina Rakotoarison}, \bibinfo{person}{Pak~Kan Wong},
  \bibinfo{person}{Baptiste Roziere}, \bibinfo{person}{Jérémy Rapin},
  \bibinfo{person}{Olivier Teytaud}, \bibinfo{person}{Antoine Moreau}, {and}
  \bibinfo{person}{Carola Doerr}.} \bibinfo{year}{2022}\natexlab{b}.
\newblock \showarticletitle{Black-Box Optimization Revisited: Improving
  Algorithm Selection Wizards Through Massive Benchmarking}.
\newblock \bibinfo{journal}{\emph{IEEE Transactions on Evolutionary
  Computation}} \bibinfo{volume}{26}, \bibinfo{number}{3}
  (\bibinfo{year}{2022}), \bibinfo{pages}{490--500}.
\newblock


\bibitem[Mirza and Osindero(2014)]%
        {goodcgan}
\bibfield{author}{\bibinfo{person}{Mehdi Mirza} {and} \bibinfo{person}{Simon
  Osindero}.} \bibinfo{year}{2014}\natexlab{}.
\newblock \showarticletitle{Conditional generative adversarial nets}.
\newblock \bibinfo{journal}{\emph{arXiv preprint arXiv:1411.1784}}
  (\bibinfo{year}{2014}).
\newblock


\bibitem[Munos(2014)]%
        {munos_bandits_2014-1}
\bibfield{author}{\bibinfo{person}{R{\'e}mi Munos}.}
  \bibinfo{year}{2014}\natexlab{}.
\newblock \bibinfo{booktitle}{\emph{From {{Bandits}} to {{Monte}}-{{Carlo Tree
  Search}}: {{The Optimistic Principle Applied}} to {{Optimization}} and
  {{Planning}}}}.
\newblock \bibinfo{type}{{T}echnical {R}eport}.
\newblock


\bibitem[Niu et~al\mbox{.}(2022)]%
        {efficientTTA2022}
\bibfield{author}{\bibinfo{person}{Shuaicheng Niu}, \bibinfo{person}{Jiaxiang
  Wu}, \bibinfo{person}{Yifan Zhang}, \bibinfo{person}{Yaofo Chen},
  \bibinfo{person}{Shijian Zheng}, \bibinfo{person}{Peilin Zhao}, {and}
  \bibinfo{person}{Mingkui Tan}.} \bibinfo{year}{2022}\natexlab{}.
\newblock \showarticletitle{Efficient Test-Time Model Adaptation without
  Forgetting}. In \bibinfo{booktitle}{\emph{39th International Conference on
  Machine Learning}}, \bibfield{editor}{\bibinfo{person}{Kamalika Chaudhuri},
  \bibinfo{person}{Stefanie Jegelka}, \bibinfo{person}{Le~Song},
  \bibinfo{person}{Csaba Szepesvari}, \bibinfo{person}{Gang Niu}, {and}
  \bibinfo{person}{Sivan Sabato}} (Eds.), Vol.~\bibinfo{volume}{162}.
  \bibinfo{pages}{16888--16905}.
\newblock


\bibitem[Niu et~al\mbox{.}(2023)]%
        {stableTTA2023}
\bibfield{author}{\bibinfo{person}{Shuaicheng Niu}, \bibinfo{person}{Jiaxiang
  Wu}, \bibinfo{person}{Yifan Zhang}, \bibinfo{person}{Zhiquan Wen},
  \bibinfo{person}{Yaofo Chen}, \bibinfo{person}{Peilin Zhao}, {and}
  \bibinfo{person}{Mingkui Tan}.} \bibinfo{year}{2023}\natexlab{}.
\newblock \showarticletitle{Towards Stable Test-time Adaptation in Dynamic Wild
  World}. In \bibinfo{booktitle}{\emph{The 11th International Conference on
  Learning Representations}}.
\newblock


\bibitem[Oquab et~al\mbox{.}(2023)]%
        {oquab2023dinov2}
\bibfield{author}{\bibinfo{person}{Maxime Oquab}, \bibinfo{person}{Timoth{\'e}e
  Darcet}, \bibinfo{person}{Th{\'e}o Moutakanni}, \bibinfo{person}{Huy Vo},
  \bibinfo{person}{Marc Szafraniec}, \bibinfo{person}{Vasil Khalidov},
  \bibinfo{person}{Pierre Fernandez}, \bibinfo{person}{Daniel Haziza},
  \bibinfo{person}{Francisco Massa}, \bibinfo{person}{Alaaeldin El-Nouby},
  {et~al\mbox{.}}} \bibinfo{year}{2023}\natexlab{}.
\newblock \showarticletitle{Dinov2: Learning robust visual features without
  supervision}.
\newblock \bibinfo{journal}{\emph{arXiv:2304.07193}} (\bibinfo{year}{2023}).
\newblock


\bibitem[Ouyang et~al\mbox{.}(2022)]%
        {ouyang2022training}
\bibfield{author}{\bibinfo{person}{Long Ouyang}, \bibinfo{person}{Jeffrey Wu},
  \bibinfo{person}{Xu Jiang}, \bibinfo{person}{Diogo Almeida},
  \bibinfo{person}{Carroll Wainwright}, \bibinfo{person}{Pamela Mishkin},
  \bibinfo{person}{Chong Zhang}, \bibinfo{person}{Sandhini Agarwal},
  \bibinfo{person}{Katarina Slama}, \bibinfo{person}{Alex Ray},
  {et~al\mbox{.}}} \bibinfo{year}{2022}\natexlab{}.
\newblock \showarticletitle{Training language models to follow instructions
  with human feedback}.
\newblock \bibinfo{journal}{\emph{Advances in Neural Information Processing
  Systems}}  \bibinfo{volume}{35} (\bibinfo{year}{2022}),
  \bibinfo{pages}{27730--27744}.
\newblock


\bibitem[Papineni et~al\mbox{.}(2002)]%
        {bleu}
\bibfield{author}{\bibinfo{person}{Kishore Papineni}, \bibinfo{person}{Salim
  Roukos}, \bibinfo{person}{Todd Ward}, {and} \bibinfo{person}{Wei-Jing Zhu}.}
  \bibinfo{year}{2002}\natexlab{}.
\newblock \showarticletitle{{BLEU}: a method for automatic evaluation of
  machine translation}. In \bibinfo{booktitle}{\emph{40th annual meeting on
  association for computational linguistics}}. \bibinfo{pages}{311--318}.
\newblock


\bibitem[Paul et~al\mbox{.}(2022)]%
        {deathmap}
\bibfield{author}{\bibinfo{person}{Sandip Paul}, \bibinfo{person}{Bhuvan
  Jhamb}, \bibinfo{person}{Deepak Mishra}, {and} \bibinfo{person}{M.~Senthil
  Kumar}.} \bibinfo{year}{2022}\natexlab{}.
\newblock \showarticletitle{Edge loss functions for deep-learning depth-map}.
\newblock \bibinfo{journal}{\emph{Machine Learning with Applications}}
  \bibinfo{volume}{7} (\bibinfo{year}{2022}), \bibinfo{pages}{100218}.
\newblock


\bibitem[Pedregosa et~al\mbox{.}(2011)]%
        {sklearn}
\bibfield{author}{\bibinfo{person}{Fabian Pedregosa},
  \bibinfo{person}{Ga{{\"e}}l Varoquaux}, \bibinfo{person}{Alexandre Gramfort},
  \bibinfo{person}{Vincent Michel}, \bibinfo{person}{Bertrand Thirion},
  \bibinfo{person}{Olivier Grisel}, \bibinfo{person}{Mathieu Blondel},
  \bibinfo{person}{Peter Prettenhofer}, \bibinfo{person}{Ron Weiss},
  \bibinfo{person}{Vincent Dubourg}, \bibinfo{person}{Jake Vanderplas},
  \bibinfo{person}{Alexandre Passos}, \bibinfo{person}{David Cournapeau},
  \bibinfo{person}{Matthieu Brucher}, \bibinfo{person}{Matthieu Perrot}, {and}
  \bibinfo{person}{{{\'E}}douard Duchesnay}.} \bibinfo{year}{2011}\natexlab{}.
\newblock \showarticletitle{Scikit-learn: Machine Learning in {P}ython}.
\newblock \bibinfo{journal}{\emph{Journal of Machine Learning Research}}
  \bibinfo{volume}{12} (\bibinfo{year}{2011}), \bibinfo{pages}{2825--2830}.
\newblock


\bibitem[Pehlivanoglu(2012)]%
        {voronoi2}
\bibfield{author}{\bibinfo{person}{Y.~Volkan Pehlivanoglu}.}
  \bibinfo{year}{2012}\natexlab{}.
\newblock \showarticletitle{A new vibrational genetic algorithm enhanced with a
  Voronoi diagram for path planning of autonomous {UAV}}.
\newblock \bibinfo{journal}{\emph{Aerospace Science and Technology}}
  \bibinfo{volume}{16}, \bibinfo{number}{1} (\bibinfo{year}{2012}),
  \bibinfo{pages}{47--55}.
\newblock


\bibitem[Polyak et~al\mbox{.}(2021)]%
        {resynt2}
\bibfield{author}{\bibinfo{person}{Adam Polyak}, \bibinfo{person}{Yossi Adi},
  \bibinfo{person}{Jade Copet}, \bibinfo{person}{Eugene Kharitonov},
  \bibinfo{person}{Kushal Lakhotia}, \bibinfo{person}{Wei{-}Ning Hsu},
  \bibinfo{person}{Abdelrahman Mohamed}, {and} \bibinfo{person}{Emmanuel
  Dupoux}.} \bibinfo{year}{2021}\natexlab{}.
\newblock \showarticletitle{Speech Resynthesis from Discrete Disentangled
  Self-Supervised Representations}.
\newblock \bibinfo{journal}{\emph{arXiv:2104.00355}} (\bibinfo{year}{2021}).
\newblock


\bibitem[Powell(1964)]%
        {powell}
\bibfield{author}{\bibinfo{person}{Michael~J.D. Powell}.}
  \bibinfo{year}{1964}\natexlab{}.
\newblock \showarticletitle{An efficient method for finding the minimum of a
  function of several variables without calculating derivatives}.
\newblock \bibinfo{journal}{\emph{Comput. J.}} \bibinfo{volume}{7},
  \bibinfo{number}{2} (\bibinfo{year}{1964}), \bibinfo{pages}{155--162}.
\newblock


\bibitem[Powell(1994)]%
        {cobyla}
\bibfield{author}{\bibinfo{person}{Michael~J.D. Powell}.}
  \bibinfo{year}{1994}\natexlab{}.
\newblock \bibinfo{booktitle}{\emph{A Direct Search Optimization Method That
  Models the Objective and Constraint Functions by Linear Interpolation}}.
\newblock \bibinfo{publisher}{Springer Netherlands}, \bibinfo{pages}{51--67}.
\newblock
\showISBNx{978-94-015-8330-5}


\bibitem[Prenger et~al\mbox{.}(2019)]%
        {waveglow}
\bibfield{author}{\bibinfo{person}{Ryan Prenger}, \bibinfo{person}{Rafael
  Valle}, {and} \bibinfo{person}{Bryan Catanzaro}.}
  \bibinfo{year}{2019}\natexlab{}.
\newblock \showarticletitle{Waveglow: A flow-based generative network for
  speech synthesis}. In \bibinfo{booktitle}{\emph{IEEE International Conference
  on Acoustics, Speech and Signal Processing}}. \bibinfo{pages}{3617--3621}.
\newblock


\bibitem[Ranftl et~al\mbox{.}(2022)]%
        {midas}
\bibfield{author}{\bibinfo{person}{Ren\'{e} Ranftl}, \bibinfo{person}{Katrin
  Lasinger}, \bibinfo{person}{David Hafner}, \bibinfo{person}{Konrad
  Schindler}, {and} \bibinfo{person}{Vladlen Koltun}.}
  \bibinfo{year}{2022}\natexlab{}.
\newblock \showarticletitle{Towards Robust Monocular Depth Estimation: Mixing
  Datasets for Zero-Shot Cross-Dataset Transfer}.
\newblock \bibinfo{journal}{\emph{IEEE Transactions on Pattern Analysis and
  Machine Intelligence}} \bibinfo{volume}{44}, \bibinfo{number}{3}
  (\bibinfo{year}{2022}).
\newblock


\bibitem[Ranzato et~al\mbox{.}(2016)]%
        {reinforceseq}
\bibfield{author}{\bibinfo{person}{Marc'Aurelio Ranzato},
  \bibinfo{person}{Sumit Chopra}, \bibinfo{person}{Michael Auli}, {and}
  \bibinfo{person}{Wojciech Zaremba}.} \bibinfo{year}{2016}\natexlab{}.
\newblock \showarticletitle{Sequence Level Training with Recurrent Neural
  Networks}. In \bibinfo{booktitle}{\emph{4th International Conference on
  Learning Representations}}, \bibfield{editor}{\bibinfo{person}{Yoshua Bengio}
  {and} \bibinfo{person}{Yann LeCun}} (Eds.).
\newblock


\bibitem[Rapin and Teytaud(2018)]%
        {nevergrad}
\bibfield{author}{\bibinfo{person}{Jeremy Rapin} {and} \bibinfo{person}{Olivier
  Teytaud}.} \bibinfo{year}{2018}\natexlab{}.
\newblock \bibinfo{title}{{Nevergrad - A gradient-free optimization platform}}.
\newblock
  \bibinfo{howpublished}{\url{https://GitHub.com/FacebookResearch/Nevergrad}}.
\newblock


\bibitem[Rechenberg(1973)]%
        {rechenberg73}
\bibfield{author}{\bibinfo{person}{Ingo Rechenberg}.}
  \bibinfo{year}{1973}\natexlab{}.
\newblock \bibinfo{booktitle}{\emph{Evolutionstrategie: Optimierung Technischer
  Systeme nach Prinzipien des Biologischen Evolution}}.
\newblock \bibinfo{publisher}{Fromman-Holzboog Verlag}.
\newblock


\bibitem[Riviere(2019)]%
        {pytorchganzoo}
\bibfield{author}{\bibinfo{person}{M. Riviere}.}
  \bibinfo{year}{2019}\natexlab{}.
\newblock \bibinfo{title}{{Pytorch GAN Zoo}}.
\newblock
  \bibinfo{howpublished}{\url{https://GitHub.com/FacebookResearch/pytorch_GAN_zoo}}.
\newblock


\bibitem[Rolinek et~al\mbox{.}(2020)]%
        {nd1}
\bibfield{author}{\bibinfo{person}{Michal Rolinek}, \bibinfo{person}{Vit
  Musil}, \bibinfo{person}{Anselm Paulus}, \bibinfo{person}{Marin Vlastelica},
  \bibinfo{person}{Claudio Michaelis}, {and} \bibinfo{person}{Georg Martius}.}
  \bibinfo{year}{2020}\natexlab{}.
\newblock \showarticletitle{Optimizing Rank-Based Metrics With Blackbox
  Differentiation}. In \bibinfo{booktitle}{\emph{IEEE/CVF Conference on
  Computer Vision and Pattern Recognition}}.
\newblock


\bibitem[Rombach et~al\mbox{.}(2022)]%
        {stablediffusion}
\bibfield{author}{\bibinfo{person}{Robin Rombach}, \bibinfo{person}{Andreas
  Blattmann}, \bibinfo{person}{Dominik Lorenz}, \bibinfo{person}{Patrick
  Esser}, {and} \bibinfo{person}{Bj\"orn Ommer}.}
  \bibinfo{year}{2022}\natexlab{}.
\newblock \showarticletitle{High-Resolution Image Synthesis With Latent
  Diffusion Models}. In \bibinfo{booktitle}{\emph{IEEE/CVF Conference on
  Computer Vision and Pattern Recognition}}. \bibinfo{pages}{10684--10695}.
\newblock


\bibitem[Ros and Hansen(2008)]%
        {diagcma}
\bibfield{author}{\bibinfo{person}{Raymond Ros} {and} \bibinfo{person}{Nikolaus
  Hansen}.} \bibinfo{year}{2008}\natexlab{}.
\newblock \showarticletitle{A Simple Modification in {CMA}-{ES} Achieving
  Linear Time and Space Complexity}. In \bibinfo{booktitle}{\emph{Parallel
  Problem Solving from Nature -- PPSN X}}. \bibinfo{publisher}{Springer Berlin
  Heidelberg}, \bibinfo{pages}{296--305}.
\newblock


\bibitem[Rozado(2023)]%
        {RightWingGPT23}
\bibfield{author}{\bibinfo{person}{David Rozado}.}
  \bibinfo{year}{2023}\natexlab{}.
\newblock \bibinfo{title}{{RightWingGPT} -- An {AI} Manifesting the Opposite
  Political Biases of {ChatGPT}}.
\newblock
\newblock
\urldef\tempurl%
\url{https://davidrozado.substack.com/p/rightwinggpt}
\showURL{%
\tempurl}
\newblock
\shownote{Last accessed on August 2024}.


\bibitem[Roziere et~al\mbox{.}(2020)]%
        {unsuptran}
\bibfield{author}{\bibinfo{person}{Baptiste Roziere},
  \bibinfo{person}{Marie-Anne Lachaux}, \bibinfo{person}{Lowik Chanussot},
  {and} \bibinfo{person}{Guillaume Lample}.} \bibinfo{year}{2020}\natexlab{}.
\newblock \showarticletitle{Unsupervised translation of programming languages}.
\newblock \bibinfo{journal}{\emph{Advances in Neural Information Processing
  Systems}}  \bibinfo{volume}{33} (\bibinfo{year}{2020}),
  \bibinfo{pages}{20601--20611}.
\newblock


\bibitem[Rozi{\`{e}}re et~al\mbox{.}(2020)]%
        {evogan}
\bibfield{author}{\bibinfo{person}{Baptiste Rozi{\`{e}}re},
  \bibinfo{person}{Fabien Teytaud}, \bibinfo{person}{Vlad Hosu},
  \bibinfo{person}{Hanhe Lin}, \bibinfo{person}{J{\'{e}}r{\'{e}}my Rapin},
  \bibinfo{person}{Mariia Zameshina}, {and} \bibinfo{person}{Olivier Teytaud}.}
  \bibinfo{year}{2020}\natexlab{}.
\newblock \showarticletitle{EvolGAN: Evolutionary Generative Adversarial
  Networks}. In \bibinfo{booktitle}{\emph{15th Asian Conference on Computer
  Vision}} \emph{(\bibinfo{series}{Lecture Notes in Computer Science},
  Vol.~\bibinfo{volume}{12625})}, \bibfield{editor}{\bibinfo{person}{Hiroshi
  Ishikawa}, \bibinfo{person}{Cheng{-}Lin Liu}, \bibinfo{person}{Tom{\'{a}}s
  Pajdla}, {and} \bibinfo{person}{Jianbo Shi}} (Eds.).
  \bibinfo{pages}{679--694}.
\newblock


\bibitem[Salimans et~al\mbox{.}(2017)]%
        {salimans2017evolution}
\bibfield{author}{\bibinfo{person}{Tim Salimans}, \bibinfo{person}{Jonathan
  Ho}, \bibinfo{person}{Xi Chen}, \bibinfo{person}{Szymon Sidor}, {and}
  \bibinfo{person}{Ilya Sutskever}.} \bibinfo{year}{2017}\natexlab{}.
\newblock \showarticletitle{Evolution strategies as a scalable alternative to
  reinforcement learning}.
\newblock \bibinfo{journal}{\emph{arXiv:1703.03864}} (\bibinfo{year}{2017}).
\newblock


\bibitem[Schwefel(1995)]%
        {Schwefel81}
\bibfield{author}{\bibinfo{person}{H.-P. Schwefel}.}
  \bibinfo{year}{1995}\natexlab{}.
\newblock \bibinfo{booktitle}{\emph{Numerical Optimization of Computer Models}
  (\bibinfo{edition}{2nd} ed.)}.
\newblock \bibinfo{publisher}{John Wiley \& Sons}.
\newblock


\bibitem[Seo and Moon(2002)]%
        {voronoi1}
\bibfield{author}{\bibinfo{person}{Dong{-}il Seo} {and}
  \bibinfo{person}{Byung~Ro Moon}.} \bibinfo{year}{2002}\natexlab{}.
\newblock \showarticletitle{Voronoi Quantizied Crossover For Traveling Salesman
  Problem}. In \bibinfo{booktitle}{\emph{{GECCO} 2002: Proceedings of the
  Genetic and Evolutionary Computation Conference, New York, USA, 9-13 July
  2002}}, \bibfield{editor}{\bibinfo{person}{William~B. Langdon},
  \bibinfo{person}{Erick Cant{\'{u}}{-}Paz}, \bibinfo{person}{Keith~E.
  Mathias}, \bibinfo{person}{Rajkumar Roy}, \bibinfo{person}{David Davis},
  \bibinfo{person}{Riccardo Poli}, \bibinfo{person}{Karthik Balakrishnan},
  \bibinfo{person}{Vasant~G. Honavar}, \bibinfo{person}{G{\"{u}}nter Rudolph},
  \bibinfo{person}{Joachim Wegener}, \bibinfo{person}{Larry Bull},
  \bibinfo{person}{Mitchell~A. Potter}, \bibinfo{person}{Alan~C. Schultz},
  \bibinfo{person}{Julian~F. Miller}, \bibinfo{person}{Edmund~K. Burke}, {and}
  \bibinfo{person}{Natasa Jonoska}} (Eds.). \bibinfo{publisher}{Morgan
  Kaufmann}, \bibinfo{pages}{544--552}.
\newblock


\bibitem[Seo et~al\mbox{.}(2005)]%
        {voronoi3}
\bibfield{author}{\bibinfo{person}{Jeong-Yeon Seo}, \bibinfo{person}{Sang-Min
  Park}, \bibinfo{person}{Seoung~Soo Lee}, {and} \bibinfo{person}{Deok-Soo
  Kim}.} \bibinfo{year}{2005}\natexlab{}.
\newblock \showarticletitle{Regrouping Service Sites: A Genetic Approach Using
  a Voronoi Diagram}. In \bibinfo{booktitle}{\emph{Computational Science and
  Its Applications -- ICCSA 2005}}, \bibfield{editor}{\bibinfo{person}{Osvaldo
  Gervasi}, \bibinfo{person}{Marina~L. Gavrilova}, \bibinfo{person}{Vipin
  Kumar}, \bibinfo{person}{Antonio Lagan{\'a}}, \bibinfo{person}{Heow~Pueh
  Lee}, \bibinfo{person}{Youngsong Mun}, \bibinfo{person}{David Taniar}, {and}
  \bibinfo{person}{Chih Jeng~Kenneth Tan}} (Eds.). \bibinfo{publisher}{Springer
  Berlin Heidelberg}, \bibinfo{address}{Berlin, Heidelberg},
  \bibinfo{pages}{652--661}.
\newblock
\showISBNx{978-3-540-32309-9}


\bibitem[Skalse et~al\mbox{.}(2023)]%
        {rh}
\bibfield{author}{\bibinfo{person}{Joar Skalse}, \bibinfo{person}{Nikolaus
  H.~R. Howe}, \bibinfo{person}{Dmitrii Krasheninnikov}, {and}
  \bibinfo{person}{David Krueger}.} \bibinfo{year}{2023}\natexlab{}.
\newblock \showarticletitle{Defining and characterizing reward hacking}. In
  \bibinfo{booktitle}{\emph{36th International Conference on Neural Information
  Processing Systems}}.
\newblock


\bibitem[Soemers et~al\mbox{.}(2023)]%
        {dennis}
\bibfield{author}{\bibinfo{person}{Dennis J. N.~J. Soemers},
  \bibinfo{person}{Vegard Mella}, \bibinfo{person}{Eric Piette},
  \bibinfo{person}{Matthew Stephenson}, \bibinfo{person}{Cameron Browne}, {and}
  \bibinfo{person}{Olivier Teytaud}.} \bibinfo{year}{2023}\natexlab{}.
\newblock \showarticletitle{Towards a General Transfer Approach for
  Policy-Value Networks}.
\newblock \bibinfo{journal}{\emph{Transactions on Machine Learning Research}}
  (\bibinfo{year}{2023}).
\newblock
\showISSN{2835-8856}


\bibitem[Speer et~al\mbox{.}(2017)]%
        {speer2017conceptnet}
\bibfield{author}{\bibinfo{person}{Robyn Speer}, \bibinfo{person}{Joshua Chin},
  {and} \bibinfo{person}{Catherine Havasi}.} \bibinfo{year}{2017}\natexlab{}.
\newblock \showarticletitle{Conceptnet 5.5: An open multilingual graph of
  general knowledge}. In \bibinfo{booktitle}{\emph{AAAI Conference on
  Artificial Intelligence}}, Vol.~\bibinfo{volume}{31}.
\newblock


\bibitem[Storn and Price(1997)]%
        {de}
\bibfield{author}{\bibinfo{person}{Rainer Storn} {and} \bibinfo{person}{Kenneth
  Price}.} \bibinfo{year}{1997}\natexlab{}.
\newblock \showarticletitle{Differential Evolution - A Simple and Efficient
  Heuristic for Global Optimization over Continuous Spaces}.
\newblock \bibinfo{journal}{\emph{J. of Global Optimization}}
  \bibinfo{volume}{11}, \bibinfo{number}{4} (\bibinfo{date}{Dec.}
  \bibinfo{year}{1997}), \bibinfo{pages}{341--359}.
\newblock


\bibitem[St{\"{u}}tzle and Ruiz(2018)]%
        {StutzleILS2018}
\bibfield{author}{\bibinfo{person}{Thomas St{\"{u}}tzle} {and}
  \bibinfo{person}{Rub{\'{e}}n Ruiz}.} \bibinfo{year}{2018}\natexlab{}.
\newblock \showarticletitle{Iterated Local Search}.
\newblock In \bibinfo{booktitle}{\emph{Handbook of Heuristics}},
  \bibfield{editor}{\bibinfo{person}{Rafael Mart{\'{\i}}},
  \bibinfo{person}{Panos~M. Pardalos}, {and} \bibinfo{person}{Mauricio G.~C.
  Resende}} (Eds.). \bibinfo{publisher}{Springer}, \bibinfo{pages}{579--605}.
\newblock


\bibitem[Sun et~al\mbox{.}(2022)]%
        {nd5}
\bibfield{author}{\bibinfo{person}{Tianxiang Sun}, \bibinfo{person}{Yunfan
  Shao}, \bibinfo{person}{Hong Qian}, \bibinfo{person}{Xuanjing Huang}, {and}
  \bibinfo{person}{Xipeng Qiu}.} \bibinfo{year}{2022}\natexlab{}.
\newblock \showarticletitle{Black-Box Tuning for Language-Model-as-a-Service}.
  In \bibinfo{booktitle}{\emph{39th International Conference on Machine
  Learning}}, \bibfield{editor}{\bibinfo{person}{Kamalika Chaudhuri},
  \bibinfo{person}{Stefanie Jegelka}, \bibinfo{person}{Le~Song},
  \bibinfo{person}{Csaba Szepesvari}, \bibinfo{person}{Gang Niu}, {and}
  \bibinfo{person}{Sivan Sabato}} (Eds.), Vol.~\bibinfo{volume}{162}.
  \bibinfo{pages}{20841--20855}.
\newblock


\bibitem[Sun et~al\mbox{.}(2020)]%
        {TTTseminalICML2020}
\bibfield{author}{\bibinfo{person}{Yu Sun}, \bibinfo{person}{Xiaolong Wang},
  \bibinfo{person}{Zhuang Liu}, \bibinfo{person}{John Miller},
  \bibinfo{person}{Alexei Efros}, {and} \bibinfo{person}{Moritz Hardt}.}
  \bibinfo{year}{2020}\natexlab{}.
\newblock \showarticletitle{Test-Time Training with Self-Supervision for
  Generalization under Distribution Shifts}. In \bibinfo{booktitle}{\emph{37th
  International Conference on Machine Learning}},
  \bibfield{editor}{\bibinfo{person}{Hal~Daumé III} {and}
  \bibinfo{person}{Aarti Singh}} (Eds.), Vol.~\bibinfo{volume}{119}.
  \bibinfo{pages}{9229--9248}.
\newblock


\bibitem[Susano~Pinto et~al\mbox{.}(2023)]%
        {nd6}
\bibfield{author}{\bibinfo{person}{Andr\'{e} Susano~Pinto},
  \bibinfo{person}{Alexander Kolesnikov}, \bibinfo{person}{Yuge Shi},
  \bibinfo{person}{Lucas Beyer}, {and} \bibinfo{person}{Xiaohua Zhai}.}
  \bibinfo{year}{2023}\natexlab{}.
\newblock \showarticletitle{Tuning Computer Vision Models With Task Rewards}.
  In \bibinfo{booktitle}{\emph{40th International Conference on Machine
  Learning}}, \bibfield{editor}{\bibinfo{person}{Andreas Krause},
  \bibinfo{person}{Emma Brunskill}, \bibinfo{person}{Kyunghyun Cho},
  \bibinfo{person}{Barbara Engelhardt}, \bibinfo{person}{Sivan Sabato}, {and}
  \bibinfo{person}{Jonathan Scarlett}} (Eds.), Vol.~\bibinfo{volume}{202}.
  \bibinfo{pages}{33229--33239}.
\newblock


\bibitem[Ventos et~al\mbox{.}(2017)]%
        {seedbridge}
\bibfield{author}{\bibinfo{person}{Veronique Ventos}, \bibinfo{person}{Yves
  Costel}, \bibinfo{person}{Olivier Teytaud}, {and} \bibinfo{person}{Solène
  Thépaut~Ventos}.} \bibinfo{year}{2017}\natexlab{}.
\newblock \showarticletitle{Boosting a Bridge Artificial Intelligence}. In
  \bibinfo{booktitle}{\emph{29th International Conference on Tools with
  Artificial Intelligence}}. \bibinfo{pages}{1280--1287}.
\newblock


\bibitem[Videau et~al\mbox{.}(2023)]%
        {mathuringecco}
\bibfield{author}{\bibinfo{person}{Mathurin Videau}, \bibinfo{person}{Nickolai
  Knizev}, \bibinfo{person}{Alessandro Leite}, \bibinfo{person}{Marc
  Schoenauer}, {and} \bibinfo{person}{Olivier Teytaud}.}
  \bibinfo{year}{2023}\natexlab{}.
\newblock \showarticletitle{Interactive Latent Diffusion Model}. In
  \bibinfo{booktitle}{\emph{Genetic and Evolutionary Computation Conference}}.
  \bibinfo{pages}{586--596}.
\newblock


\bibitem[Wang et~al\mbox{.}(2021)]%
        {TENT2021}
\bibfield{author}{\bibinfo{person}{Dequan Wang}, \bibinfo{person}{Evan
  Shelhamer}, \bibinfo{person}{Shaoteng Liu}, \bibinfo{person}{Bruno
  Olshausen}, {and} \bibinfo{person}{Trevor Darrell}.}
  \bibinfo{year}{2021}\natexlab{}.
\newblock \showarticletitle{Tent: Fully Test-Time Adaptation by Entropy
  Minimization}. In \bibinfo{booktitle}{\emph{International Conference on
  Learning Representations}}.
\newblock


\bibitem[Wang et~al\mbox{.}(2009)]%
        {wam}
\bibfield{author}{\bibinfo{person}{Yizao Wang}, \bibinfo{person}{Jean-Yves
  Audibert}, {and} \bibinfo{person}{R\'{e}mi Munos}.}
  \bibinfo{year}{2009}\natexlab{}.
\newblock \showarticletitle{Algorithms for Infinitely Many-Armed Bandits}.
\newblock In \bibinfo{booktitle}{\emph{Advances in Neural Information
  Processing Systems 21}}, \bibfield{editor}{\bibinfo{person}{D.~Koller},
  \bibinfo{person}{D.~Schuurmans}, \bibinfo{person}{Y.~Bengio}, {and}
  \bibinfo{person}{L.~Bottou}} (Eds.). \bibinfo{publisher}{Curran Associates,
  Inc.}, \bibinfo{pages}{1729--1736}.
\newblock


\bibitem[Wang et~al\mbox{.}(2017)]%
        {tacotron}
\bibfield{author}{\bibinfo{person}{Yuxuan Wang}, \bibinfo{person}{R.~J.
  Skerry-Ryan}, \bibinfo{person}{Daisy Stanton}, \bibinfo{person}{Yonghui Wu},
  \bibinfo{person}{Ron~J. Weiss}, \bibinfo{person}{Navdeep Jaitly},
  \bibinfo{person}{Zongheng Yang}, \bibinfo{person}{Ying Xiao},
  \bibinfo{person}{Zhifeng Chen}, \bibinfo{person}{Samy Bengio},
  \bibinfo{person}{Quoc~V. Le}, \bibinfo{person}{Yannis Agiomyrgiannakis},
  \bibinfo{person}{Rob Clark}, {and} \bibinfo{person}{Rif~A. Saurous}.}
  \bibinfo{year}{2017}\natexlab{}.
\newblock In \bibinfo{booktitle}{\emph{INTERSPEECH}},
  \bibfield{editor}{\bibinfo{person}{Francisco Lacerda}} (Ed.).
  \bibinfo{pages}{4006--4010}.
\newblock


\bibitem[Weise et~al\mbox{.}({[n.\,d.]})]%
        {betandrun}
\bibfield{author}{\bibinfo{person}{Thomas Weise}, \bibinfo{person}{Zijun Wu},
  {and} \bibinfo{person}{Markus Wagner}.} \bibinfo{year}{[n.\,d.]}\natexlab{}.
\newblock \showarticletitle{An Improved Generic Bet-and-Run Strategy for
  Speeding Up Stochastic Local Search}.
\newblock \bibinfo{journal}{\emph{arXiv:1806.08984}}
  (\bibinfo{year}{[n.\,d.]}).
\newblock


\bibitem[Williams(1992)]%
        {Williams_92_SimpleStatisticalGradientFollowing}
\bibfield{author}{\bibinfo{person}{Ronald~J. Williams}.}
  \bibinfo{year}{1992}\natexlab{}.
\newblock \showarticletitle{Simple {{Statistical Gradient}}-{{Following
  Algorithms}} for {{Connectionist Reinforcement Learning}}}. In
  \bibinfo{booktitle}{\emph{Machine {{Learning}}}}. \bibinfo{pages}{229--256}.
\newblock


\bibitem[Xian et~al\mbox{.}(2018)]%
        {Monocular_relative_depth_perception_with_web_stereo_data_supervision}
\bibfield{author}{\bibinfo{person}{Ke Xian}, \bibinfo{person}{Chunhua Shen},
  \bibinfo{person}{Zhiguo Cao}, \bibinfo{person}{Hao Lu}, \bibinfo{person}{Yang
  Xiao}, \bibinfo{person}{Ruibo Li}, {and} \bibinfo{person}{Zhenbo Luo}.}
  \bibinfo{year}{2018}\natexlab{}.
\newblock \showarticletitle{Monocular Relative Depth Perception with Web Stereo
  Data Supervision}. In \bibinfo{booktitle}{\emph{IEEE/CVF Conference on
  Computer Vision and Pattern Recognition}}. \bibinfo{pages}{311--320}.
\newblock


\bibitem[Xu et~al\mbox{.}(2008)]%
        {SATzilla}
\bibfield{author}{\bibinfo{person}{Lin Xu}, \bibinfo{person}{Frank Hutter},
  \bibinfo{person}{Holger~H. Hoos}, {and} \bibinfo{person}{Kevin
  Leyton-Brown}.} \bibinfo{year}{2008}\natexlab{}.
\newblock \showarticletitle{{SAT}zilla: Portfolio-based Algorithm Selection for
  {SAT}}.
\newblock \bibinfo{journal}{\emph{J. Artif. Int. Res.}} \bibinfo{volume}{32},
  \bibinfo{number}{1} (\bibinfo{date}{June} \bibinfo{year}{2008}),
  \bibinfo{pages}{565--606}.
\newblock


\bibitem[Yin et~al\mbox{.}(2020)]%
        {depthcriteria}
\bibfield{author}{\bibinfo{person}{Wei Yin}, \bibinfo{person}{Xinlong Wang},
  \bibinfo{person}{Chunhua Shen}, \bibinfo{person}{Yifan Liu},
  \bibinfo{person}{Zhi Tian}, \bibinfo{person}{Songcen Xu},
  \bibinfo{person}{Changming Sun}, {and} \bibinfo{person}{Dou Renyin}.}
  \bibinfo{year}{2020}\natexlab{}.
\newblock \showarticletitle{DiverseDepth: Affine-invariant Depth Prediction
  Using Diverse Data}.
\newblock \bibinfo{journal}{\emph{arXiv:2002.00569}} (\bibinfo{year}{2020}).
\newblock


\bibitem[Zameshina et~al\mbox{.}(2022)]%
        {fairgan}
\bibfield{author}{\bibinfo{person}{M. Zameshina}, \bibinfo{person}{O. Teytaud},
  \bibinfo{person}{Fabien Teytaud}, \bibinfo{person}{Vlad Hosu},
  \bibinfo{person}{Nathanael Carraz}, \bibinfo{person}{Laurent Najman}, {and}
  \bibinfo{person}{Markus Wagner}.} \bibinfo{year}{2022}\natexlab{}.
\newblock \showarticletitle{Fairness in Generative Modeling: Do It
  Unsupervised!}. In \bibinfo{booktitle}{\emph{Genetic and Evolutionary
  Computation Conference Companion}}. \bibinfo{pages}{320--323}.
\newblock


\bibitem[Zhu et~al\mbox{.}(2023)]%
        {rlhf}
\bibfield{author}{\bibinfo{person}{Banghua Zhu}, \bibinfo{person}{Michael
  Jordan}, {and} \bibinfo{person}{Jiantao Jiao}.}
  \bibinfo{year}{2023}\natexlab{}.
\newblock \showarticletitle{Principled Reinforcement Learning with Human
  Feedback from Pairwise or K-wise Comparisons}. In
  \bibinfo{booktitle}{\emph{40th International Conference on Machine
  Learning}}, \bibfield{editor}{\bibinfo{person}{Andreas Krause},
  \bibinfo{person}{Emma Brunskill}, \bibinfo{person}{Kyunghyun Cho},
  \bibinfo{person}{Barbara Engelhardt}, \bibinfo{person}{Sivan Sabato}, {and}
  \bibinfo{person}{Jonathan Scarlett}} (Eds.), Vol.~\bibinfo{volume}{202}.
  \bibinfo{pages}{43037--43067}.
\newblock


\bibitem[Zhuang et~al\mbox{.}(2021)]%
        {transferLearningSurvey2021}
\bibfield{author}{\bibinfo{person}{Fuzhen Zhuang}, \bibinfo{person}{Zhiyuan
  Qi}, \bibinfo{person}{Keyu Duan}, \bibinfo{person}{Dongbo Xi},
  \bibinfo{person}{Yongchun Zhu}, \bibinfo{person}{Hengshu Zhu},
  \bibinfo{person}{Hui Xiong}, {and} \bibinfo{person}{Qing He}.}
  \bibinfo{year}{2021}\natexlab{}.
\newblock \showarticletitle{A Comprehensive Survey on Transfer Learning}.
\newblock \bibinfo{journal}{\emph{Proc. IEEE}} \bibinfo{volume}{109},
  \bibinfo{number}{1} (\bibinfo{year}{2021}), \bibinfo{pages}{43--76}.
\newblock


\bibitem[Zilberstein(1996)]%
        {anytime}
\bibfield{author}{\bibinfo{person}{Shlomo Zilberstein}.}
  \bibinfo{year}{1996}\natexlab{}.
\newblock \showarticletitle{Using Anytime Algorithms in Intelligent Systems}.
\newblock \bibinfo{journal}{\emph{AI Magazine}} \bibinfo{volume}{17},
  \bibinfo{number}{3} (\bibinfo{year}{1996}), \bibinfo{pages}{73}.
\newblock


\end{thebibliography}

\FloatBarrier

\appendix

\section{Full results for the AfterLearnER / MiDaS experiment}\label{appmidas}
We observe here (i) good performance even with a low budget (total budget $<300$ is enough for positive results) (ii) good transfer, in particular when training on Th3. \change{Note that AfterLearnER succeeds in 25 cases out of 27 independent runs, which happens with p-value $2.8e-6$ under the null hypothesis that the success rate is at most $0.5$.}

\begin{figure}
Testing on Th1, training with Th1, Th2, Th3 respectively:\\
\includegraphics[width=.3\textwidth]{{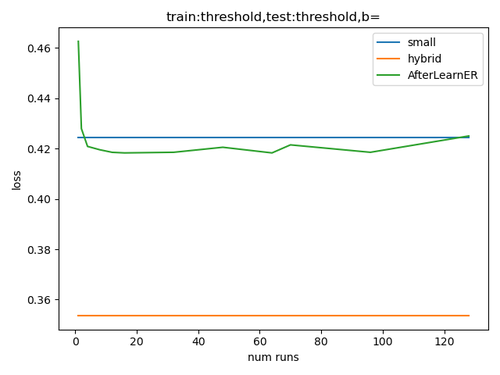}}
\includegraphics[width=.3\textwidth]{{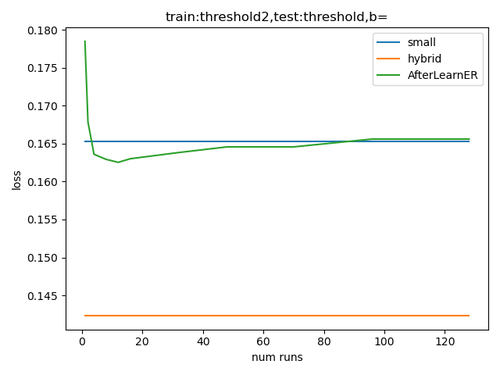}}
\includegraphics[width=.3\textwidth]{{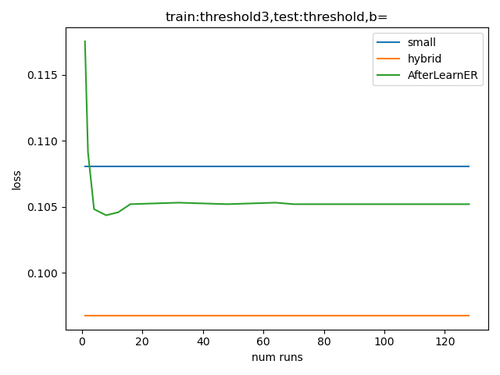}}\\

Testing on Th2, training with Th1, Th2, Th3 respectively:\\
\includegraphics[width=.3\textwidth]{{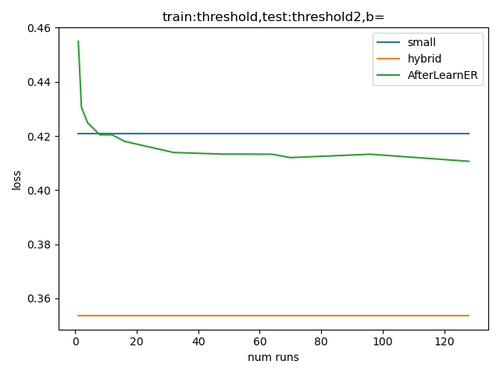}}
\includegraphics[width=.3\textwidth]{{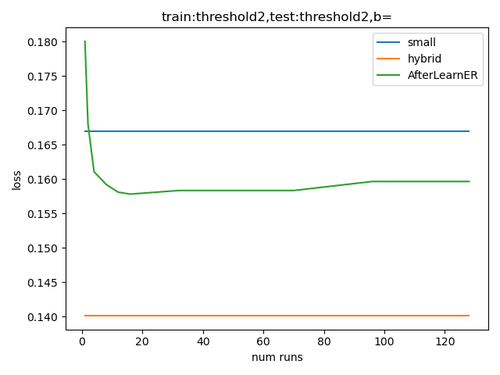}}
\includegraphics[width=.3\textwidth]{{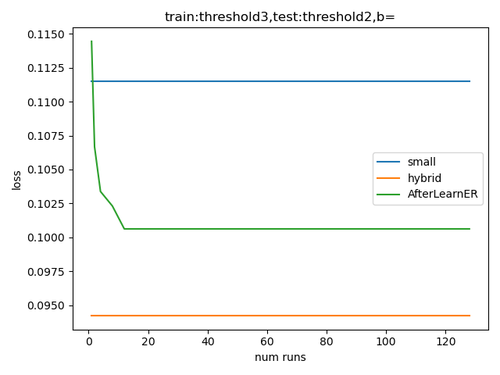}}\\

Testing on Th3, training with Th1, Th2, Th3 respectively:\\
\includegraphics[width=.3\textwidth]{{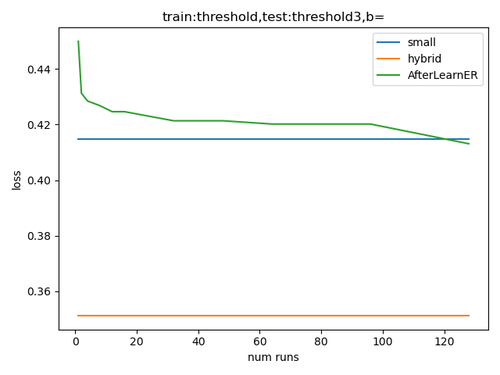}}
\includegraphics[width=.3\textwidth]{{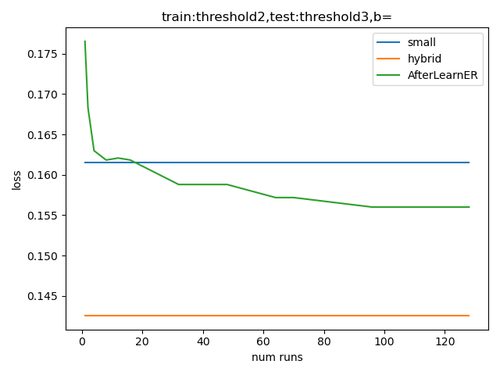}}
\includegraphics[width=.3\textwidth]{{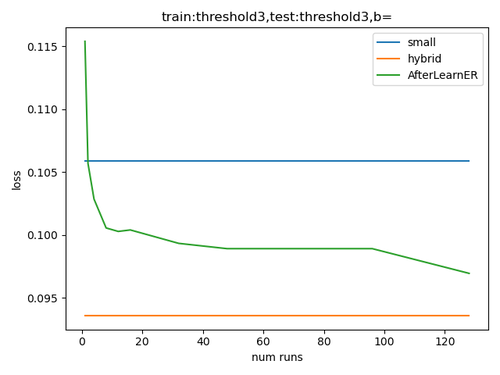}}\\
\caption{AfterLearnER applied to MiDaS with budget $B=\{1\}$: with this budget 1, we are actually running random search and already get positive results. We note that with a training on Th3, we get positive results for all tests with $b\times k$ (total number of scalar evaluations provided to AfterLearner) greater than 20: so 20 evaluations is enough for a good transfer to all criteria Th1, Th2, Th3. \label{b1}}
\end{figure}

\begin{figure}
Testing on Th1, training with Th1, Th2, Th3 respectively:\\
\includegraphics[width=.3\textwidth]{{newmidas/thresholdthresholdb_50.png}}
\includegraphics[width=.3\textwidth]{{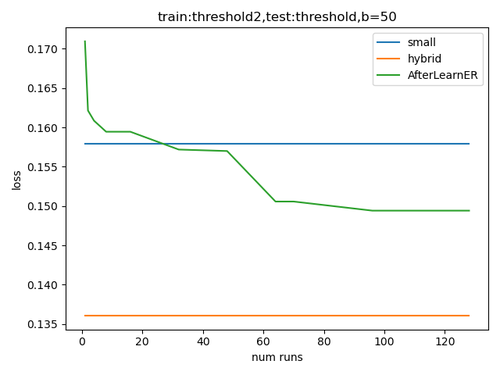}}
\includegraphics[width=.3\textwidth]{{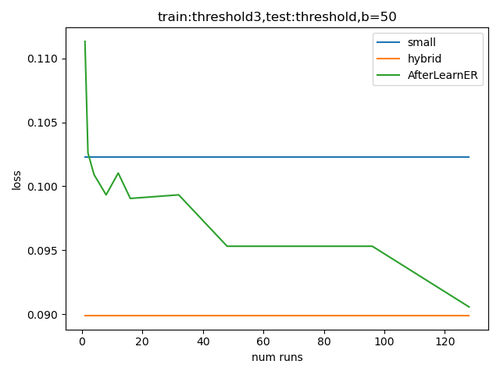}}\\

Testing on Th2, training with Th1, Th2, Th3 respectively:\\
\includegraphics[width=.3\textwidth]{{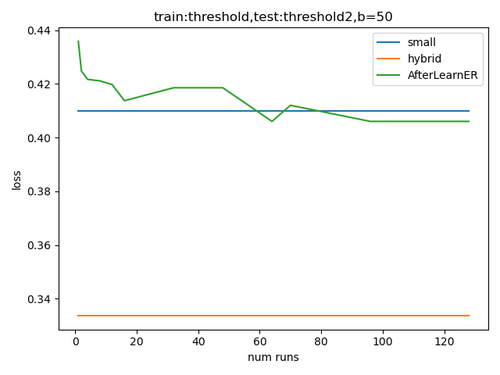}}
\includegraphics[width=.3\textwidth]{{newmidas/threshold2threshold2b_50.png}}
\includegraphics[width=.3\textwidth]{{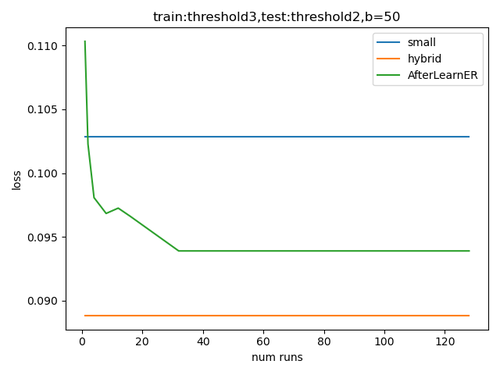}}\\

Testing on Th3, training with Th1, Th2, Th3 respectively:\\
\includegraphics[width=.3\textwidth]{{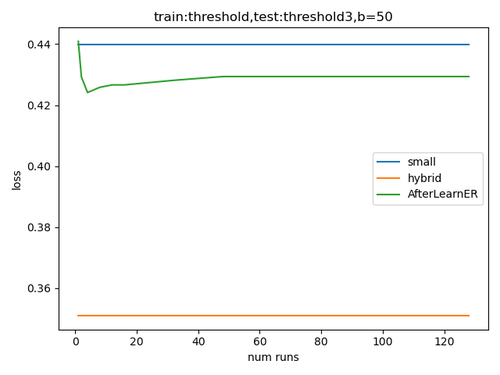}}
\includegraphics[width=.3\textwidth]{{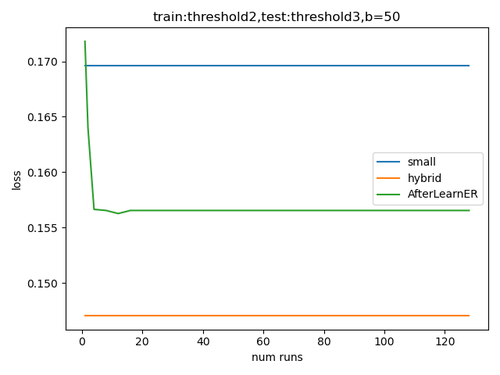}}
\includegraphics[width=.3\textwidth]{{newmidas/threshold3threshold3b_50.png}}\\
\caption{AfterLearnER applied to MiDaS with budget $B=\{50\}$. Compared to $B=\{1\}$, the number of scalar evaluations needed for beating the baseline is greater, but remains in the hundreds. Th3 remains the best training criterion for good transfer to all 3 criteria Th1,Th2,Th3.\label{b50}}
\end{figure}

\begin{figure}
Testing on Th1, training with Th1, Th2, Th3 respectively:\\
\includegraphics[width=.3\textwidth]{{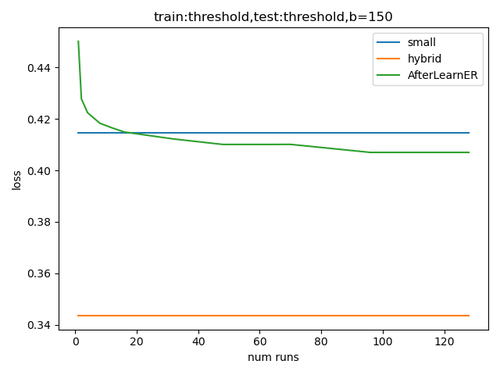}}
\includegraphics[width=.3\textwidth]{{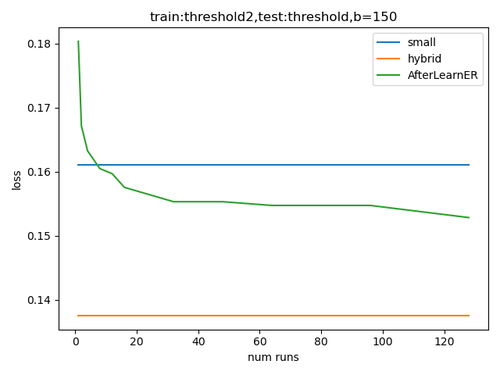}}
\includegraphics[width=.3\textwidth]{{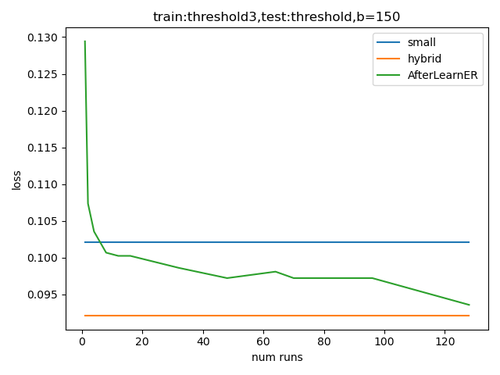}}\\

Testing on Th2, training with Th1, Th2, Th3 respectively:\\
\includegraphics[width=.3\textwidth]{{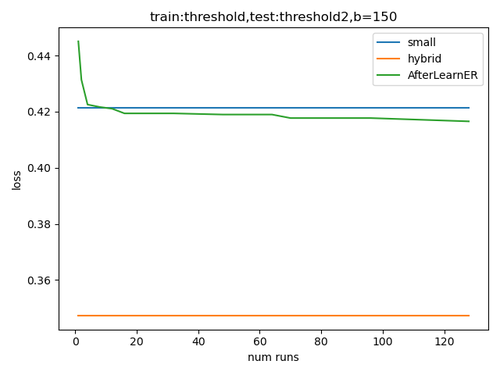}}
\includegraphics[width=.3\textwidth]{{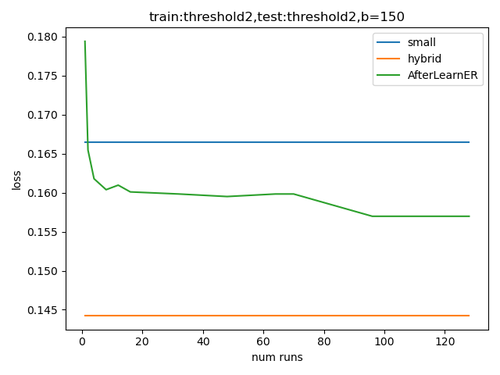}}
\includegraphics[width=.3\textwidth]{{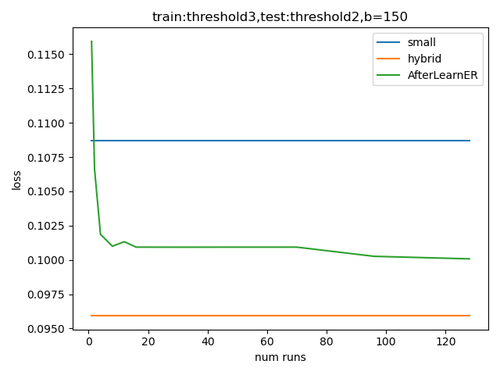}}\\

Testing on Th3, training with Th1, Th2, Th3 respectively:\\
\includegraphics[width=.3\textwidth]{{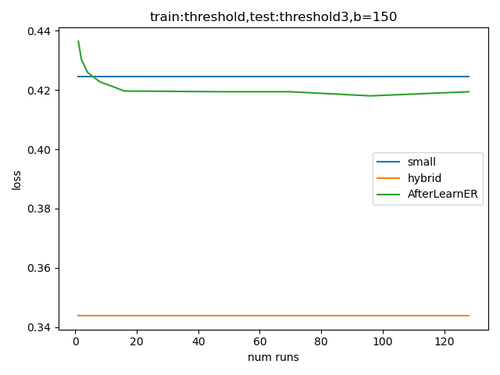}}
\includegraphics[width=.3\textwidth]{{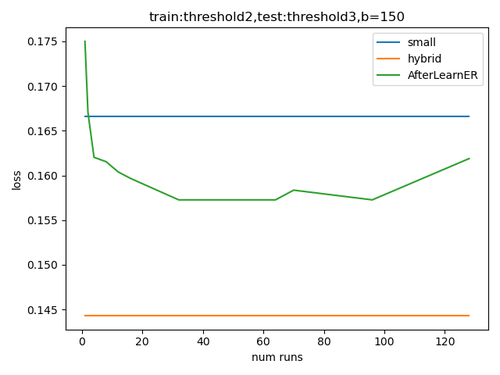}}
\includegraphics[width=.3\textwidth]{{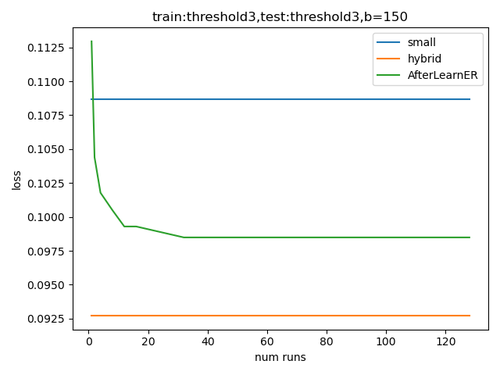}}\\
\caption{AfterLearnER applied to MiDaS with budget $B=\{150\}$. Compared to $B=\{1\}$, the number of scalar evaluations needed for beating the baseline is greater, but remains in the hundreds. Th3 remains the best training criterion for good transfer to all 3 criteria Th1,Th2,Th3.\label{b150}}
\end{figure}

For each budget, there are three loss functions Th1, Th2 and Th3 which can be used in training or in test, hence 9 curves.
For 3 of them, train and test use the same loss function: the 6 others are about transfer.
The non-transfer cases are also presented in the main text.
Fig. \ref{b1} considers a budget $b=1$. Fig. \ref{b50} considers a budget $b=50$. Fig. \ref{b150} considers a budget 150.
In each figure, the x-axis is the number of runs $k$. \change{We observe an excellent transfer with $b$ small, which suggests that the good mathematical properties in terms of overfitting (Eq. \ref{eqov}) also indicate a robust transfer.}

\FloatBarrier
\section{Ask and Tell}
\label{askandtell}
\begin{algorithm}[!h]
   \centering
    \begin{algorithmic}
        \Require{objective function $f$, number of workers $P\geq 1$, domain $D$, budget $b$}
        \State{$o.initialize\left(D,b,P\right)$} \Comment{Initialize $o$ given $P$, domain $D$, and budget $b$}
        \Procedure{Worker-task}{$x$}
           \State{Compute $f\left(x\right)$}
           \State{ $o.tell\left(x,f\left(x\right)\right)$} \Comment{Inform the method $o$ of the performance obtained by the \HP{} $x$, no gradient needed}
        \EndProcedure
        \For{$i \in \{1,\dots,b\}$}
           \State{Wait until there are less than $P$ computations in progress in the FIFO}
           \State{Request $x\leftarrow o.ask\left(\right)$ ($x\in D$)} \Comment{Choice of \hp{} to be tested}
           \State{Put $x$ in the FIFO of workers tasks}
        \EndFor
        \State{Wait until the FIFO is empty}
        \State{\Return{$o.recommend\left(\right)$}} \Comment{Final returned value}
    \end{algorithmic}
	\caption{\label{bbo}General organization of a parallel asynchronous black-box optimization method $o$. There is a first-in-first-out (FIFO) pool of pairs $\left(x,f\left(x\right)\right)$ to be computed by the workers. The sequential case corresponds to $P=1$. 
 }
\end{algorithm}
\FloatBarrier
\section{Black-box optimization wizards and NGOpt}\label{sec:ngopt}
\FloatBarrier
After becoming famous in SAT solving\cite{SATzilla}, wizards become important in black-box optimization as well\cite{betandrun,versatile,squirrel,NGOpt}.

Optimization wizards are based on selecting, from characteristics of the domain (dimension, type of variables) and of the optimization run (budget, parallelism), an optimization algorithm or a bet-and-run~\cite{betandrun} or chaining of several optimization algorithms. NGOpt, an open-source wizard for black-box optimization, is available in \cite{nevergrad} and as a PyPi package.

For example, in a continuous setting, with moderate dimension, a large budget, and a sequential optimization context (no parallelism), 
NGOpt\cite{NGOpt} chooses a chaining of CMA~\cite{CMA} fastened with a metamodel and followed by Sequential Quadratic Programming\cite{artelysSQP2}.

An example of code involving NGOpt is provided in~\cref{repro}.

\FloatBarrier
\section{Reproducibility} \label{repro}
\FloatBarrier
All black-box optimization algorithms are available in~\cite{nevergrad}. Nevergrad can be installed by cloning \url{https://github.com/facebookresearch/nevergrad} or by ``pip install nevergrad''.

A minimal example of optimization with some optimization methods mentioned above is in~\cref{nevex} (without link with deep learning).
\Cref{pyt} presents an application for the optimization of a single layer of a PyTorch model.
\begin{algorithm*}
\caption{\label{nevex}Sample code: Minimal example of black-box optimization.}
   \centering
\begin{lstlisting}[style=pythonstyle]

import nevergrad as ng
import numpy as np

def loss(x):
    return np.sum((x - 1.7) ** 2)

# 3 optimization methods in dimension 10, continuous.
optim_ngopt = ng.optimizers.registry["NGOpt"](10, 500)
optim_de = ng.optimizers.registry["DE"](10, 500)
optim_lengler= ng.optimizers.registry["DiscreteLenglerOnePlusOne"](10, 500)

# 1 optimization method for dimension 10, integer values between 1 and 4, 
# with 30 parallel evaluations.
array = ng.p.Array(init=np.array([2] * 10), lower=1, upper=4).set_integer_casting()
optim_opo = ng.optimizers.registry["DiscreteOnePlusOne"](array, 500)

# an example with minimize
print(optim_ngopt.minimize(lambda x: loss(x)).value)

# an example with ask and tell
for k in range(50):
    x = optim_de.ask()
    y = loss(x.value)
    optim_de.tell(x, y)

# an example with minimize in a discrete context
print(optim_opo.minimize(loss).value)

\end{lstlisting}
\end{algorithm*}

The detailed code to improve EG3D is in~\cref{additionalcode}.

A minimal code for running Retrofit on a Pytorch model (all parameters) is~\cref{minimalalg}.

\begin{algorithm*}
\begin{lstlisting}[style=pythonstyle]

  # Our retrofit method:
  # less overfitting, more privacy, less poisoning, no fair use issue.
  def retrofit(model, epochs,optim_name="NGOpt",cross_entropy=False):
    with torch.no_grad():
      if cross_entropy:
        criterion = nn.CrossEntropyLoss()
      else:
        criterion = binary_accuracy
      d = 0
      # In this version we work on all parameters.
      for name, param in model.named_parameters():
          d += len(param.data.detach().cpu().numpy().flatten())
          #print(d)
      optim = ng.optimizers.registry[optim_name](d,epochs)
      for epoch in range(epochs):
          total_loss = 0
          num = 0
          w = optim.ask()
          set_weights(model,w.value)
          for x, y in train_loader:
              y_scores = model(x)
              if cross_entropy:
                loss = criterion(y_scores, y)
              else:
                loss = criterion(y_scores[:,1], y)
              #loss.backward() not needed here!
              #optimizer.step()
              total_loss += loss.item()
              num += len(y)
          optim.tell(w,total_loss/num)
          #if epoch %
          #    print(epoch, total_loss / num)
      set_weights(model,optim.recommend().value)
    
\end{lstlisting}
\caption{\label{minimalalg}Minimal example of code modifying a Pytorch model for cross-entropy or directly for the classification error. The key point is not minimizing the cross-entropy error (which can be done much faster with gradient-based optimization) but the direct optimization of the success rate.
}
\end{algorithm*}

\section{Computational cost}\label{sec:compcost}
\FloatBarrier
\begin{table}\centering
\caption{\label{cc}Computational cost. *Here, the cost is in terms of data collection. **We get positive results even with a very low budget, though greater budgets lead to better results (\Cref{figcodegen}).
The online overhead exists only when we modify, online, the latent variables of a GAN for each new generation of images; however, it is typically negligible compared to the generation of images itself.}

\begin{tabular}{|c|c|c|}
\hline
Application & \multicolumn{2}{|c|}{computational cost} \\
\hline
            &  Offline comp. cost & Online overhead \\
\hline
\hline
Depth sensing & & \\
(Fig. \ref{xpmidas}, $k=10$, $B=50$) & 12h, 10 GPUs   & zero \\
\hline
Speech resynthesis (Tab. \ref{multirunsresynt}, 1st line) & 1 GPU, 2 hours & zero \\
\hline
EG3D (\cref{eg3dcase})&  200 labelled examples* & negligible \\
\hline
\ifthenelse{\songoku=0}{Evol. Interactive GAN (\cref{egan}) }{} & \ifthenelse{\songoku=0}{selecting 5 batches*}{} & \ifthenelse{\songoku=0}{zero}{}\\
 & \ifthenelse{\songoku=0}{out of 30}{} & \\
 \hline
LDM (offline, Section \ref{sd})  & 200 images & negligible \\
\hline
LDM (online, Section \ref{sec:onlinesd})  & negligible & negligible \\
\hline
Doom (Tab. \ref{arnoldtab}) & 48 hours, $k$ GPU & zero \\
\hline
Code translation (\cref{ct}) & 0 to 54 hours**, & zero \\
      & 0 to thousands of   & \\
  &   labelled translations & \\ 
\hline
\end{tabular}
\end{table}

\Cref{cc} presents typical computational costs of AfterLearnER for a typical entire run, providing significant improvements compared to the initial DL tool.

\section{Code: AfterLearnER online in the case of EG3D for cat generation}
\change{
\Cref{additionalcode} shows the integration of our work in the case of cats generated by EG3D.
}%

\begin{algorithm*}
   \centering
\begin{lstlisting}[style=pythonstyle]
          # Code for modifying a latent vector z, using a list of the 
          # bad random seeds over the 200 first seeds.
          if "afhqcats512-128" in network_pkl:
                Y = [1] * 200  # Seeds 15, 84, 47 are bad for cats.
                bad_latent_z = [15,84,47,30,33,78,4,50,49,
		    56,11,17, 36,85,59,16,63,81,25,39, 87,45,41,58,60,68,46,38]
                bad_latent_z += [130,113,146,107,143,
		      175,158,192,121, 110,181,186,170,120,163,193,155,
		      144,148,191,141, 105,187,183,174,
                    129,117,100,157,166,194,198,138, 156,168,114,127,103,109]
            if "ffhqrebalanced512-64" in network_pkl:
                Y = [1] * 200
                bad_latent_z = [27,14,88,42,37,25,73, 53,40,17,45,92,
                    50,3,84,29,12,31,98,72,51,46,57,16,13,33,
                    43,69,60,74, 77,75,41,89,39,87,68,67,56,
                    35,26,9,23,7,38,20,63,55, 28,36,8,6,64]
                bad_latent_z += [132,145,194,187,119,169, 137,176,185, 147,111,139,171,
                    101,161,116,115,155,175,118,112,144,172,
                    198,124,181, 129,117,189,108,174,138,121,143,
                    156,130,146,179,140,153,136,168,131,180,110, 109,199,100]
            for p in bad_latent_z:
                Y[p] = 0
            # Random generation
            X = [list(np.random.RandomState(i).randn(1, G.z_dim)[0])
                 for i in range(len(Y))]
            # Learning with Scikit-Learn.
            from sklearn import tree
            clf = tree.DecisionTreeClassifier()
            print(len(X), len(Y), len(X[0]))
            clf = clf.fit(X, Y)
            # Optimization with Nevergrad: modify bad cats.
            z = np.random.RandomState(seed).randn(1, G.z_dim)
            import nevergrad as ng
            epsilon = 0.01
            def loss(x):
                assert len(x) == G.z_dim
                return clf.predict_proba(
                   [list(z[0] + epsilon * x)])[0][0]
            nevergrad_optimizer = ng.optimizers.NGOpt(G.z_dim, 10000)
            for nevergrad_iteration in range(1, 10000):
                x = nevergrad_optimizer.ask()
                l = loss(x.value)
                if l < 1e-5:
                    break
                else:
                    nevergrad_optimizer.tell(x, l)
            print(f"Success in {nevergrad_iteration} iterations")
            z[0] += epsilon * nevergrad_optimizer.recommend().value
\end{lstlisting}
   \caption{The short code at the top of EG3D-cats for getting AfterLearnER on top of it. There is no significant computational overhead, we learn directly from $z$: we do not generate anything.}\label{additionalcode}
\end{algorithm*}

\section{{Tuning a Pytorch model: code}}\label{apcode}
\change{
\Cref{repro} presents a minimal example of black-box optimization in the general case, in the case of the platform Nevergrad used in all our experiments. In~\cref{pyt}, we present a few lines of code for optimizing a single layer of a Pytorch model. In \cref{minimalalg}, we provide a function for optimizing all weights of a Pytorch model.
The long URL for a small-scale experiment is \url{https://colab.research.google.com/drive/10GujGQp_59SX60kLoGQ5TOKF9Na1ybRM} (short URL: \url{https://tinyurl.com/teloretrofit}): we refer to the reading guide at the top of this link for more details. Additional experiments on residual networks for classifying images are available at \url{https://colab.research.google.com/drive/1tlmBE55mo-NatTlGd-q3nM0mMak4DN92} (short URL \url{https://tinyurl.com/retrofitresnet}) and for a small scale \url{https://colab.research.google.com/drive/108hV_zNjAqP21C9JIwbIre63UhvX-gGu} (short URL \url{https://tinyurl.com/retrofitresnetsmall}).}
\begin{algorithm}
\begin{lstlisting}[style=pythonstyle]
    import nevergrad as ng
    # Here you need to load your Python model.
    Todo.
    # Get a good initial value:
    with torch.no_grad():
       w0 = my_model.output.weight.data.detach().cpu().numpy
    # w0.shape can then be used for initializing an array ng.p.Array(shape=w0.shape):
    my_optim = ng.optimizers.DiscreteLenglerOnePlusOne(
       ng.p.Array(shape=w0.shape), budget=500)
    # w0 can then be used in optim.suggest().
    my_optim.suggest(w0)
    # Modify the value so that it becomes w:
    for i in range(my_optim.budget):
        w = my_optim.ask()
        with torch.no_grad():
            my_model.output.weight.data.copy_(torch.tensor(w.value))
        loss = Todo
        #After computing the loss loss (e.g. loss=-accuracy):
	   my_optim.tell(w, loss)
    # The optimization is over:
    # we put the recommended w 
    # in the pytorch model.
    w = my_optim.recommend()
    with torch.no_grad():
        my_model.output.weight.data.copy_(torch.tensor(w.value))
    # You might save your model here.
    Todo.
\end{lstlisting}
\caption{\label{pyt}\change{Python code for tuning a single layer of a Pytorch model (we refer to~\cref{minimalalg} for a more general case). In this example, we focus on the last layer.}}
\end{algorithm}

\end{document}